\documentclass[runningheads]{llncs}
\usepackage{graphicx}

\usepackage{tikz}
\usepackage{comment}
\usepackage{amsmath,amssymb} 
\usepackage{color}
\usepackage{graphicx}
\usepackage{amsmath}
\usepackage{amssymb}
\usepackage{booktabs}
\usepackage{soul}
\usepackage{multirow}
\usepackage{textcomp}  
\usepackage{gensymb}
\usepackage{lipsum} 
\usepackage{pifont}
\newcommand{\cmark}{\ding{51}}%
\usepackage{enumitem}
\usepackage{colortbl}
\usepackage{rotating}

\usepackage{xcolor}
\usepackage{subcaption}
\usepackage[pagebackref,breaklinks,colorlinks,
citecolor=blue,linkcolor=blue,urlcolor=blue]{hyperref}
\usepackage{wrapfig}

\usepackage{lipsum}
\usepackage{blindtext}
\newcommand{\rot}[1]{\rotatebox[origin=c]{50}{#1}}
\usepackage[capitalize]{cleveref}
\crefname{section}{Sec.}{Secs.}
\Crefname{section}{Section}{Sections}
\Crefname{table}{Table}{Tables}
\crefname{table}{Tab.}{Tabs.}

\colorlet{lightgray}{gray!20}

\newcommand\numberthis{\addtocounter{equation}{1}\tag{\theequation}}

\usepackage{xspace}
\makeatletter
\DeclareRobustCommand\onedot{\futurelet\@let@token\@onedot}
\def\@onedot{\ifx\@let@token.\else.\null\fi\xspace}
\def\eg{\emph{e.g}\onedot} 
\def\ie{\emph{i.e}\onedot} 
 
\def\etc{\emph{etc}\onedot} \def\vs{\emph{vs}\onedot}

\def\etal{\emph{et al}\onedot}
\makeatother

\usepackage[accsupp]{axessibility}  

\usepackage[width=122mm,left=12mm,paperwidth=146mm,height=193mm,top=12mm,paperheight=217mm]{geometry}

\begin{document}
\pagestyle{headings}
\mainmatter
\def\ECCVSubNumber{4589}  

\title{V2X-ViT: Vehicle-to-Everything Cooperative Perception with Vision Transformer} 

\titlerunning{V2X-ViT: V2X Perception with Vision Transformer}
%
\author{Runsheng Xu\inst{1}\thanks{Equal contribution. $^\dagger$ Corresponding author: \texttt{jiaqima@ucla.edu}} \and
Hao Xiang\inst{1}$^\star$\and
Zhengzhong Tu\inst{2}$^\star$\and
Xin Xia\inst{1} \and \\
Ming-Hsuan Yang\inst{3,4} \and
Jiaqi Ma\inst{1}$^\dagger$
}
\authorrunning{R. Xu et al.}
%
\institute{University of California, Los Angeles \and
University of Texas at Austin \and
Google Research \and
University of California, Merced
}

\maketitle

\begin{abstract}
In this paper, we investigate the application of Vehicle-to-Everything (V2X) communication to improve the perception performance of autonomous vehicles. We present a robust cooperative perception framework with V2X communication using a novel vision Transformer. Specifically, we build a holistic attention model, namely V2X-ViT, to effectively fuse information across on-road agents (i.e., vehicles and infrastructure). V2X-ViT consists of alternating layers of heterogeneous multi-agent self-attention and multi-scale window self-attention, which captures inter-agent interaction and per-agent spatial relationships. These key modules are designed in a unified Transformer architecture to handle common V2X challenges, including asynchronous information sharing, pose errors, and heterogeneity of V2X components. To validate our approach, we create a large-scale V2X perception dataset using CARLA and OpenCDA. Extensive experimental results demonstrate that V2X-ViT sets new state-of-the-art performance for 3D object detection and achieves robust performance even under harsh, noisy environments. The code is available at \url{https://github.com/DerrickXuNu/v2x-vit}.
\end{abstract}

\section{Introduction}
\label{sec:intro}
Perceiving the complex driving environment precisely is crucial to the safety of autonomous vehicles~(AVs). With recent advancements of deep learning, the robustness of single-vehicle perception systems has demonstrated significant improvement in several tasks such as semantic segmentation~\cite{treml2016speeding,el2019rgb}, depth estimation~\cite{zhou2021r,9826399}, and object detection and tracking~\cite{lang2019pointpillars,liang2018deep,zhao2022track,fan2021deep}. Despite recent advancements, challenges remain. Single-agent perception system tends to suffer from occlusion and sparse sensor observation at a far distance, which can potentially cause catastrophic consequences~\cite{zhang2021safe}. The cause of such a problem is that an individual vehicle can only perceive the environment from a single perspective with limited sight-of-view. To address these issues, recent studies~\cite{wang2020v2vnet,chen2019cooper,xu2021opv2v,chen2019f,yuan2022keypoints,lei2022latency} leverage the advantages of multiple viewpoints of the same scene by investigating Vehicle-to-Vehicle~(V2V) collaboration, where visual information (\eg, detection outputs, raw sensory information, intermediate deep learning features, details see~\cref{sec:related}) from multiple nearby AVs are shared for a complete and accurate understanding of the environment.

\begin{figure}[!t]
\centering
\subfloat[Snapshot of Simulation]{%
  \includegraphics[height=0.22\columnwidth,width=0.4\columnwidth]{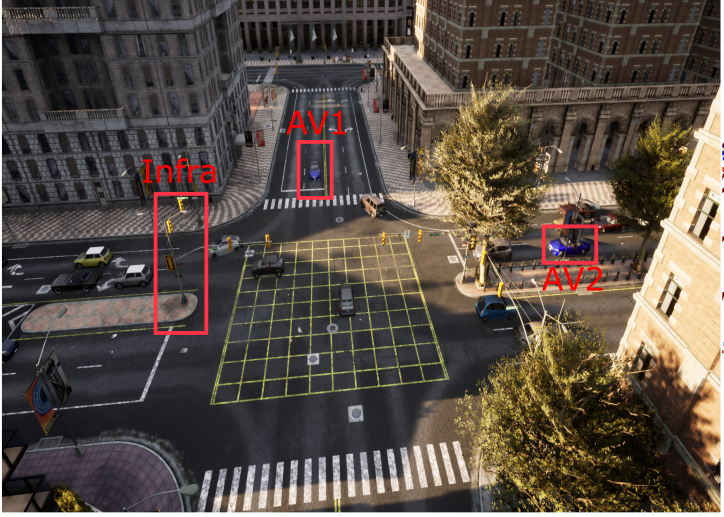}%
}
\hfil
\subfloat[Aggregated LiDAR point cloud]{%
  \includegraphics[height=0.22\columnwidth,width=0.4\columnwidth]{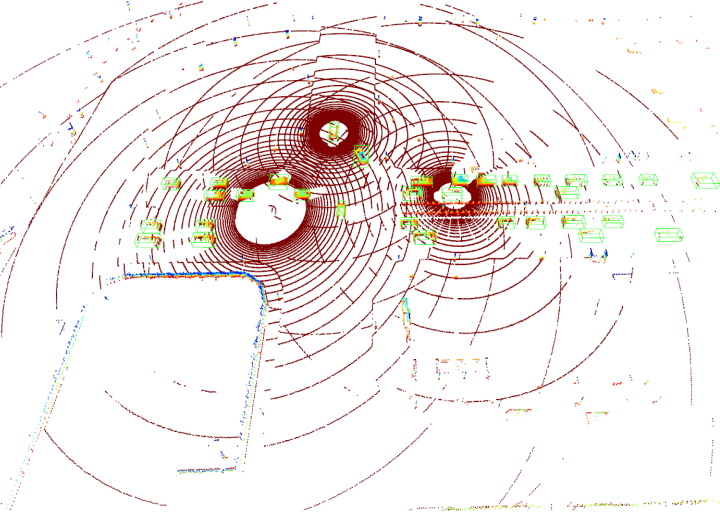}%
}
\caption{\textbf{A data sample from the proposed V2XSet}. (a) A simulated scenario in CARLA where two AVs and infrastructure are located at different sides of a busy intersection. (b) The aggregated LiDAR point clouds of these three agents.}
\label{fig:data_sample}
\vspace{-4mm}
\end{figure}

Although V2V technologies have the prospect to revolutionize the mobility industry, it ignores a critical collaborator -- roadside infrastructure. The presence of AVs is usually unpredictable, whereas the infrastructure can always provide supports once installed in key scenes such as intersections and crosswalks. Moreover, infrastructure equipped with sensors on an elevated position has a broader sight-of-view and potentially less occlusion.
Despite these advantages, including infrastructure to deploy a robust V2X perception system is non-trivial. Unlike V2V collaboration, where all agents are homogeneous, V2X systems often involve a heterogeneous graph formed by infrastructure and AVs. The configuration discrepancies between infrastructure and vehicle sensors, such as types, noise levels, installation height, and even sensor attributes and modality, make the design of a V2X perception system challenging.
Moreover, the GPS localization noises and the asynchronous sensor measurements of AVs and infrastructure can introduce inaccurate coordinate transformation and lagged sensing information.
Failing to properly handle these challenges will make the system vulnerable.

\begin{figure*}[!t]
\centering
\includegraphics[width=0.90\linewidth]{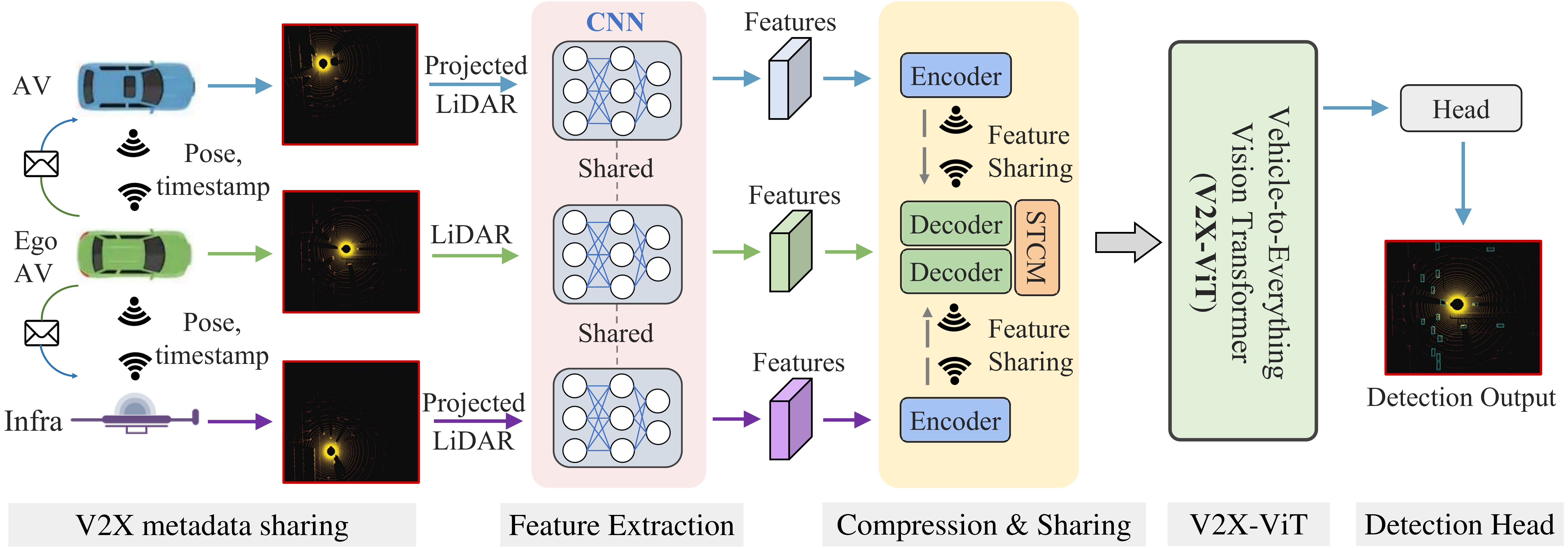}
\caption{\textbf{Overview of our proposed V2X perception system.} It consists of five sequential steps: V2X metadata sharing, feature extraction, compression \& sharing, V2X-ViT, and the detection head. The details of each individual component are illustrated in \cref{ssec:overall}.}
\label{fig:Architecture}
\vspace{-3mm}
\end{figure*}

In this paper, we introduce a unified fusion framework, namely V2X Vision Transformer or {\textbf{V2X-ViT}}, for V2X perception, that can jointly handle these challenges. \cref{fig:Architecture} illustrates the entire system. AVs and infrastructure capture, encode, compress, and send intermediate visual features with each other, while the ego vehicle~(\ie, receiver) employs V2X-Transformer to perform information fusion for object detection. We propose two novel attention modules to accommodate V2X challenges: 1) a customized heterogeneous multi-agent self-attention module that explicitly considers agent types (vehicles and infrastructure) and their connections when performing attentive fusion; 2) a multi-scale window attention module that can handle localization errors by using multi-resolution windows in parallel. These two modules will adaptively iteratively fuse visual features to capture inter-agent interaction and per-agent spatial relationship, correcting the feature misalignment caused by localization error and time delay.  Moreover, we also integrate a delay-aware positional encoding to handle the time delay uncertainty further. Notably, all these modules are incorporated in a single transformer that learns to address these challenges end-to-end.

To evaluate our approach, we collect a new large-scale open dataset,  namely V2XSet, that explicitly considers real-world noises during V2X communication using the high-fidelity simulator CARLA~\cite{Dosovitskiy17}, and a cooperative driving automation simulation tool OpenCDA. \cref{fig:data_sample} shows a data sample in the collected dataset. Experiments show that our proposed V2X-ViT significantly advances the performance on V2X LiDAR-based 3D object detection, achieving a 21.2\% gain of AP compared to single-agent baseline and performing favorably against leading intermediate fusion methods by at least 7.3\%.
Our contributions are:
\begin{itemize}[leftmargin=*]
\itemsep0em
    \item We present the first unified transformer architecture~(V2X-ViT) for V2X perception, which can capture the heterogeneity nature of V2X systems with strong robustness against various noises. Moreover, the proposed model achieves state-of-the-art performance on the challenging cooperative detection task.
    \item We propose a novel heterogeneous multi-agent attention module (HMSA) tailored for adaptive information fusion between heterogeneous agents.
    \item We present a new multi-scale window attention module (MSwin) that simultaneously captures local and global spatial feature interactions in parallel.
    \item We construct V2XSet, a new large-scale open simulation dataset for V2X perception, which explicitly accounts for imperfect real-world conditions.
\end{itemize}

\section{Related work}
\label{sec:related}
\noindent\textbf{V2X perception.} Cooperative perception studies how to efficiently fuse visual cues from neighboring agents. Based on its message sharing strategy, it can be divided into 3 categories: 1) early fusion~\cite{chen2019cooper} where raw data is shared and gathered to form a holistic view, 2) intermediate fusion~\cite{wang2020v2vnet,xu2021opv2v,vadivelu2020learning,chen2019f} where intermediate neural features are extracted based on each agent's observation and then transmitted, and 3) late fusion~\cite{rauch2012car2x,rawashdeh2018collaborative} where detection outputs~(\eg, 3D bounding box position, confidence score) are circulated. As early fusion usually requires large transmission bandwidth and late fusion fails to provide valuable scenario context~\cite{wang2020v2vnet},  intermediate fusion has attracted increasing attention because of its good balance between accuracy and transmission bandwidth. Several intermediate fusion methods have been proposed for V2V perception recently. OPV2V~\cite{xu2021opv2v} implements a single-head self-attention module to fuse features, while F-Cooper employs \textit{maxout}~\cite{goodfellow2013maxout} fusion operation. V2VNet~\cite{wang2020v2vnet} proposes a spatial-aware message passing mechanism to jointly reason detection and prediction. To attenuate outlier messages, \cite{vadivelu2020learning} regresses vehicles' localization errors with consistent pose constraints. DiscoNet~\cite{li2021learning} leverages knowledge distillation to enhance training by constraining the corresponding features to the ones from the network for early fusion. However, intermediate fusion for V2X is still in its infancy. Most V2X methods explored late fusion strategies to aggregate information from infrastructure and vehicles. For example, a late fusion two-level Kalman filter is proposed by~\cite{mo2021method} for roadside infrastructure failure conditions. Xiangmo~\etal~\cite{zhao2017cooperative} propose fusing the lane mark detection from infrastructure and vehicle sensors, leveraging Dempster-Shafer theory to model the uncertainty. 

\noindent\textbf{LiDAR-based 3D object detection.} Numerous methods have been explored to extract features from raw points, voxels, bird-eye-view~(BEV) images, and their mixtures. PointRCNN~\cite{shi2019pointrcnn} proposes a two-stage strategy based on raw point clouds, which learns rough estimation in the first stage and then refines it with semantic attributes. The authors of~\cite{zhou2018voxelnet,yan2018second} propose to split the space into voxels and produce features per voxel. Despite having high accuracy, their inference speed and memory consumption are difficult to optimize due to reliance on 3D convolutions. To avoid computationally expensive 3D convolutions,~\cite{lang2019pointpillars,yang2018pixor} propose an efficient BEV representation. To satisfy both computational and flexible receptive field requirements, ~\cite{shi2020pv,zhong2021vin,yang2019std} combine voxel-based and point-based approaches to detect 3D objects. 

\noindent\textbf{Transformers in vision.}
The Transformer~\cite{vaswani2017attention} is first proposed for machine translation~\cite{vaswani2017attention}, where multi-head self-attention and feed-forward layers are stacked to capture long-range interactions between words. Dosovitskiy \etal ~\cite{dosovitskiy2020image} present a Vision Transformer (ViT) for image recognition by regarding image patches as visual words and directly applying self-attention. The full self-attention in ViT~\cite{vaswani2017attention,dosovitskiy2020image,fan2022svt}, despite having global interaction, suffers from heavy computational complexity and does not scale to long-range sequences or high-resolution images. To ameliorate this issue, numerous methods have introduced locality into self-attention, such as Swin~\cite{liu2021swin}, CSwin~\cite{dong2021cswin}, Twins~\cite{chu2021twins}, window~\cite{wang2021uformer,tu2022maxim}, and sparse attention~\cite{vaswani2021scaling,tu2022maxvit,xu2022cobevt}. A hierarchical architecture is usually adopted to progressively increase the receptive fields for capturing longer dependencies.

While these vision transformers have proven efficient in modeling homogeneous structured data, their efficacy to represent heterogeneous graphs has been less studied. One of the developments related to our work is the heterogeneous graph transformer (HGT)~\cite{hu2020heterogeneous}. HGT was originally designed for web-scale Open Academic Graph where the nodes are text and attributes. Inspired by HGT, we build a customized heterogeneous multi-head self-attention module tailored for graph attribute-aware multi-agent 3D visual feature fusion, which is able to capture the heterogeneity of V2X systems.

\section{Methodology}
\label{sec:methodology}

In this paper, we consider V2X perception as a heterogeneous multi-agent perception system, where different types of agents (\ie, smart infrastructure and AVs) perceive the surrounding environment and communicate with each other. To simulate real-world scenarios, we assume that all the agents have imperfect localization and time delay exists during feature transmission.
Given this, our goal is to develop a robust fusion system to enhance the vehicle's perception capability and handle these aforementioned challenges in a unified end-to-end fashion. The overall architecture of our framework is illustrated in \cref{fig:Architecture}, which includes five major components: 1) metadata sharing, 2) feature extraction, 3) compression and sharing, 4) V2X vision Transformer, and 5) a detection head. 

\subsection{Main architecture design}
\label{ssec:overall}

\noindent\textbf{V2X metadata sharing.}
During the early stage of collaboration, every agent $i\in\{1\dots N\}$ within the communication networks shares metadata such as poses, extrinsics, and agent type $c_i\in\{I, V\}$~(meaning infrastructure or vehicle) with each other. We select one of the connected AVs as the ego vehicle ($e$) to construct a V2X graph around it where the nodes are either AVs or infrastructure and the edges represent directional V2X communication channels.  
To be more specific, we assume the transmission of metadata is well-synchronized, which means each agent $i$ can receive ego pose $x_{e}^{t_i}$ at the time $t_i$. Upon receiving the pose of the ego vehicle, all the other connected agents nearby will project their own LiDAR point clouds to the ego-vehicle's coordinate frame before feature extraction. 

\noindent\textbf{Feature extraction.} We leverage the anchor-based PointPillar method~\cite{lang2019pointpillars} to extract visual features from point clouds because of its low inference latency and optimized memory usage~\cite{xu2021opv2v}. The raw point clouds will be converted to a stacked pillar tensor, then scattered to a 2D pseudo-image and fed to the PointPillar backbone. The backbone extracts informative feature maps $\mathbf{F}_i^{t_i}\in \mathbb{R}^{H \times W \times C}$, denoting agent $i$'s feature at time $t_i$ with  height $H$, width $W$, and channels $C$.

\noindent\textbf{Compression and sharing.} 
To reduce the required transmission bandwidth, we utilize a series of $1\times1$ convolutions to progressively compress the feature maps along the channel dimension. The compressed features with the size $(H,W,C')$ (where $C'\ll C$) are then transmitted to the ego vehicle ($e$), on which the features are projected back to $(H,W,C)$ using $1 \times 1$ convolutions.

There exists an inevitable time gap between the time when the LiDAR data is captured by connected agents and when the extracted features are received by the ego vehicle (details see appendix). Thus, features collected from surrounding agents are often temporally misaligned with the features captured on the ego vehicle.
To correct this delay-induced global spatial misalignment, we need to transform (\ie, rotate and translate) the received features to the current ego-vehicle's pose. Thus, we leverage a spatial-temporal correction module~(STCM), which employs a differential transformation and sampling operator $\mathbf{\Gamma_{\xi}}$ to spatially warp the feature maps~\cite{jaderberg2015spatial}. An ROI mask is also calculated to prevent the network from paying attention to the padded zeros caused by the spatial warp.

\noindent\textbf{V2X-ViT.}
The intermediate features $\mathbf{H}_i=\mathbf{\Gamma_{\xi}}\left(\mathbf{F}_i^{t_i}\right)\in \mathbb{R}^{H \times W \times C}$ aggregated from connected agents are fed  into the major component of our framework \ie, V2X-ViT to conduct an iterative inter-agent and intra-agent feature fusion using self-attention mechanisms. We maintain the feature maps in the same level of high resolution throughout the entire Transformer as we have observed that the absence of high-definition features greatly harms the objection detection performance. The details of our proposed V2X-ViT will be unfolded in \cref{ssec:v2ctransformer}.

\noindent\textbf{Detection head.} After receiving the final fused feature maps, we apply two $1\times1$ convolution layers for box regression and classification. The regression output is $(x, y, z, w, l, h, \theta)$, denoting the position, size, and yaw angle of the predefined anchor boxes, respectively. The classification output is the confidence score of being an object or background for each anchor box. We use the smooth $\ell_1$ loss for regression and a focal loss~\cite{lin2017focal} for classification.

\begin{figure*}[!t]
\centering
\includegraphics[width=0.90\linewidth ]{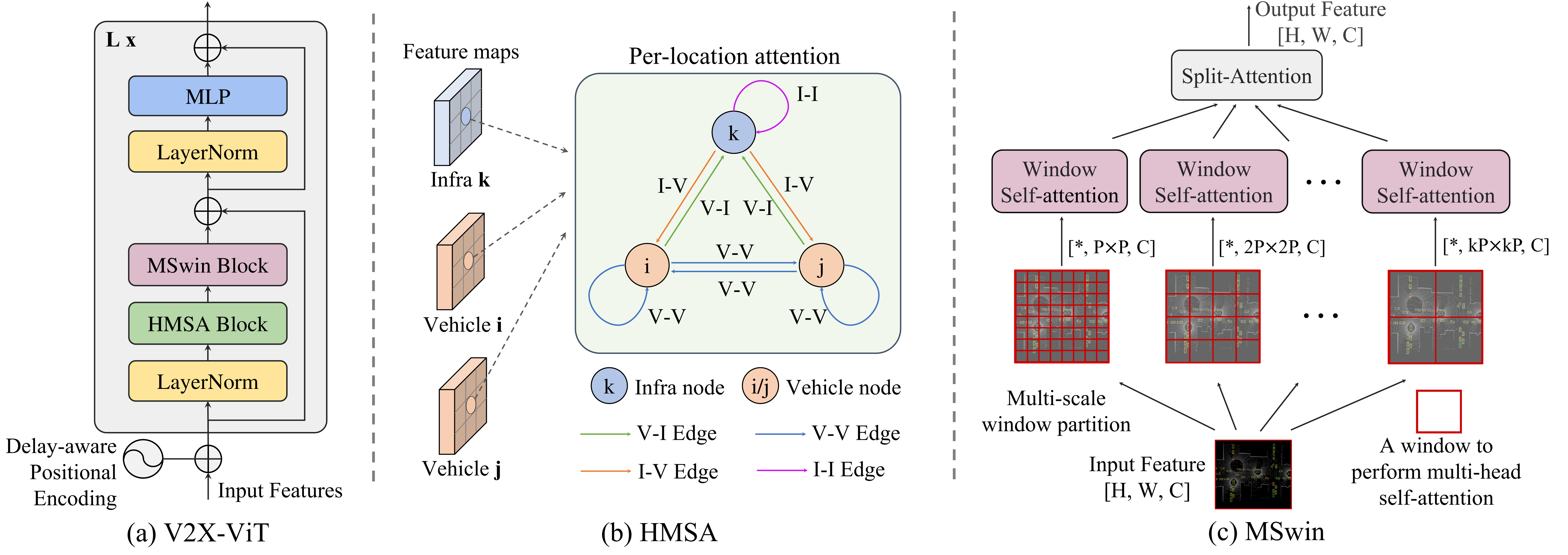}
\caption{\textbf{V2X-ViT architecture.} (a) The architecture of our proposed V2X-ViT model. (b) Heterogeneous multi-agent self-attention (HMSA) presented in \cref{sssec:hmasa}. (c) Multi-scale window attention module (MSwin) illustrated in \cref{sssec:mswin}.}
\label{fig:V2X-ViT-architecture}
\vspace{-4mm}
\end{figure*}

\subsection{V2X-Vision Transformer}
\label{ssec:v2ctransformer}
Our goal is to design a customized vision Transformer that can jointly handle the common V2X challenges.
Firstly, to effectively capture the heterogeneous graph representation between infrastructure and AVs, we build a heterogeneous multi-agent self-attention module that learns different relationships based on node and edge types. Moreover, we propose a novel spatial attention module, namely multi-scale window attention (MSwin), that captures long-range interactions at various scales. MSwin uses multiple window sizes to aggregate spatial information, which greatly improves the detection robustness against localization errors. Lastly, these two attention modules are integrated into a single V2X-ViT block in a factorized manner (illustrated in \cref{fig:V2X-ViT-architecture}a), enabling us to maintain high-resolution features throughout the entire process. We stack a series of V2X-ViT blocks to iteratively learn inter-agent interaction and per-agent spatial attention, leading to a robust aggregated feature representation for detection.

\subsubsection{Heterogeneous multi-agent self-attention}
\label{sssec:hmasa}
The sensor measurements captured by infrastructure and AVs possibly have distinct characteristics. The infrastructure's LiDAR is often installed at a higher position with less occlusion and different view angles. In addition, the sensors may have different levels of sensor noise due to maintenance frequency, hardware quality \etc. To encode this heterogeneity, we build a novel heterogeneous multi-agent self-attention~({HMSA}) where we attach types to both nodes and edges in the directed graph. To simplify the graph structure, we assume the sensor setups among the same category of agents are identical. 
As shown in \cref{fig:V2X-ViT-architecture}b, we have two types of nodes and four types of edges, \ie, node type $c_{i}\in\{I,V\}$ and edge type $\phi\left(e_{ij}\right)\in\{V\!-\!V,V\!-\!I,I\!-\!V,I\!-\!I\}$. Note that unlike traditional attention where the node features are treated as a vector, we only reason the interaction of features \textit{in the same spatial position} from different agents to preserve spatial cues.
Formally, HSMA is expressed as:
\begin{equation}
\mathbf{H}_i=\underset{\forall j\in N\left(i\right)}{\mathsf{Dense}_{c_i}}\left(\mathbf{ATT}\left(i,j\right)\cdot\mathbf{MSG}\left(i,j\right) \right)
\label{eq:hmsa}
\end{equation}
which contains 3 operators: a linear aggregator $\mathsf{Dense}_{c_i}$, attention weights estimator $\mathbf{ATT}$, and message aggregator $\mathbf{MSG}$. The $\mathsf{Dense}$ is a set of linear projectors indexed by the node type $c_i$, aggregating multi-head information. $\mathbf{ATT}$ calculates the importance weights between pairs of nodes conditioned on the associated node and edge types:
\vspace{-3mm}
\begin{align*}
    {\mathbf{ATT}}\left(i,j\right)&=\underset{\forall j\in N\left(i \right)}{\mathsf{softmax}}\left(\underset{m\in[1,h]}{\|}\text{head}^m_{\mathrm{ATT}}\left(i,j\right) \right)\numberthis \label{1}\\
    \text{head}^m_{\mathrm{ATT}}\left(i,j\right)&=\left(\mathbf{K}^m\left(j\right)\mathbf{W}_{\phi\left(e_{ij}\right)}^{m,\mathrm{ATT}}\mathbf{Q}^m\left(i\right)^T \right) \frac{1}{\sqrt{C}}\numberthis \label{2}\\
    \mathbf{K}^m\left(j\right)&=\mathsf{Dense}^{m}_{c_j}\left({\mathbf{H}_j}\right) \numberthis \label{3}\\
    \mathbf{Q}^m\left(i\right)&=\mathsf{Dense}^{m}_{c_i}\left({\mathbf{H}_i}\right) \numberthis \label{4}
\end{align*}
where $\|$ denotes concatenation, $m$ is the current head number and $h$ is the total number of heads. Notice that $\mathsf{Dense}$ here is indexed by both node type $c_{i/j}$, and head number $m$. The linear layers in $\textbf{K}$ and $\textbf{Q}$ have distinct parameters. To incorporate the semantic meaning of edges, we calculate the dot product between Query and Key vectors weighted by a matrix $\mathbf{W}_{\phi\left(e_{ij}\right)}^{m, \mathrm{ATT}}\in\mathbb{R}^{C\times C}$. Similarly, when parsing messages from the neighboring agent, we embed infrastructure and vehicle's features separately via $\mathsf{Dense}_{c_j}^m$. A matrix $\mathbf{W}_{\phi\left(e_{ij}\right)}^{m,\mathrm{MSG}}$ is used to project the features based on the edge type between source node and target node:
\vspace{-2mm}
\begin{align*}
    \mathbf{MSG}\left(i,j\right)&=\underset{m\in[1,h]}{\|}\text{head}^m_{\mathrm{MSG}}\left(i ,j\right)\numberthis \label{5}\\
    \text{head}^m_{\mathrm{MSG}}\left(i,j\right)&=\mathsf{Dense}^m_{c_j}\left(\mathbf{H}_j\right)\mathbf{W}_{\phi\left(e_{ij}\right)}^{m,\mathrm{MSG}} .\numberthis \label{6}
\end{align*}

\subsubsection{Multi-scale window attention}
\label{sssec:mswin}
We present a new type of attention mechanism tailored for efficient long-range spatial interaction on high-resolution detection, called multi-scale window attention (MSwin). It uses a pyramid of windows, each of which caps a different attention range, as illustrated in \cref{fig:V2X-ViT-architecture}c. The usage of variable window sizes can greatly improve the detection robustness of V2X-ViT against localization errors (see ablation study in \cref{fig:ablation}b). Attention performed within larger windows can capture long-range visual cues to compensate for large localization errors, whereas smaller window branches perform attention at finer scales to preserve local context. Afterward, the split-attention module~\cite{zhang2020resnest} is used to adaptively fuse information coming from multiple branches, empowering MSwin to handle a range of pose errors. Note that MSwin is applied on each agent independently without considering any inter-agent fusion; therefore we omit the agent subscript in this subsection for simplicity.

Formally, let $\mathbf{H}\in\mathbb{R}^{H\times W\times C}$ be an input feature map of a single agent. In branch $j$ out of $k$ parallel branches, $\mathbf{H}$ is partitioned using window size $P_j\! \times\! P_j$, into a tensor of shape $(\frac{H}{P_j}\times\frac{W}{P_j},P_j\times P_j,C)$, which represents a $\frac{H}{P_j}\times\frac{W}{P_j}$ grid of non-overlapping patches each with size $P_j\times P_j$. We use $h_j$ number of heads to improve the attention power at $j$-th branch. More detailed formulation can be found in Appendix.
Following~\cite{liu2021swin,hu2019local}, we also consider an additional relative positional encoding $\mathbf{B}$ that acts as a bias term added to the attention map. As the relative position along each axis lies in the range $[-P_j+1,P_j-1]$,  we take $\mathbf{B}$ from a parameterized matrix $\hat{\mathbf{B}}\in\mathbb{R}^{(2P_j-1)\times(2P_j-1)}$.

To attain per-agent multi-range spatial relationship, each branch partitions input tensor $\mathbf{H}$ with different window sizes \ie $\{P_j\}_{j=1}^k=\{P,2P,...,kP\}$. 
We progressively decrease the number of heads when using a larger window size to save memory usage. Finally, we fuse the features from all the branches by a Split-Attention module~\cite{zhang2020resnest}, yielding the output feature $\mathbf{Y}$.
The complexity of the proposed MSwin is \textit{linear} to image size $HW$, while enjoying long-range multi-scale receptive fields and adaptively fuses both local and (sub)-global visual hints in parallel. Notably, unlike Swin Transformer~\cite{liu2021swin}, our multi-scale window approach requires no masking, padding, or cyclic-shifting, making it more efficient in implementations while having larger-scale spatial interactions.

\vspace{-3mm}
\subsubsection{Delay-aware positional encoding}
Although the global misalignment is captured by the spatial warping matrix $\Gamma_{\xi}$, another type of local misalignment, arising from object motions during the delay-induced time lag, also needs to be considered.
To encode this temporal information, we leverage an adaptive delay-aware positional encoding (DPE), composed of a linear projection and a learnable embedding. We initialize it with sinusoid functions conditioned on time delay $\Delta t_i$ and channel $c\in[1,C]$: 
\begin{equation}
    \mathbf{p}_c\left(\Delta t_{i}\right)=\begin{cases} 
      \sin\left(\Delta t_{i}/10000^{\frac{2c}{C}}\right),&c=2k\\
      \cos\left(\Delta t_{i}/10000^{\frac{2c}{C}}\right),&c=2k+1\\
   \end{cases}
\end{equation}

A linear projection $f:\mathbb{R}^C\rightarrow\mathbb{R}^C$ will further warp the learnable embedding so it can generalize better for unseen time delay~\cite{hu2020heterogeneous}. We add this projected embedding to each agents' feature $\mathbf{H}_i$ before feeding into the Transformer so that the features are temporally aligned beforehand. 
\begin{align*}
    \text{DPE}\left(\Delta t_{i}\right)&=f\left(\mathbf{p}\left(\Delta t_{i}\right)\right)\numberthis \label{9}\\
    \mathbf{H}_i&= \mathbf{H}_i+\text{DPE}\left(\Delta t_{i}\right)\numberthis \label{10}
\end{align*}
\section{Experiments}
\subsection{V2XSet: An open dataset for V2X cooperative perception}
\label{sec:dataset}
To the best of our knowledge, no fully public V2X perception dataset exists suitable for investigating common V2X challenges such as localization error and transmission time delay.
DAIR-V2X~\cite{DAIR-V2X2021} is a large-scale real-world V2I dataset without V2V cooperation. V2X-Sim~\cite{Li_2021_RAL} is an open V2X simulated dataset but does not simulate noisy settings and only contains a single road type. OPV2V~\cite{xu2021opv2v} contains more road types but are restricted to V2V cooperation.
To this end, we collect a new large-scale dataset for V2X perception that explicitly considers these real-world noises during V2X collaboration using CARLA~\cite{Dosovitskiy17} and OpenCDA~\cite{xu2021opencda} together. 
In total, there are 11,447 frames in our dataset~(33,081 samples if we count frames per agent in the same scene), and the train/validation/test splits are 6,694/1,920/2,833, respectively. Compared with existing datasets, V2XSet incorporates both V2X cooperation and realistic noise simulation. Please refer to the supplementary material for more details.

\subsection{Experimental setup}
Th evaluation range in $x$ and $y$ direction are $[-140, 140]$~m and $[-40, 40]$~m respectively.  
We assess models under two settings: 1)~\textit{Perfect Setting}, where the pose is accurate, and everything is synchronized across agents; 2)~\textit{Noisy Setting}, where pose error and time delay are both considered. In the \textit{Noisy Setting},  the positional and heading noises of the transmitter are drawn from a Gaussian distribution with a default standard deviation of 0.2~m and 0.2\textdegree~respectively, following the real-world noise levels~\cite{xia2021advancing,li2020toward,RT3000}. The time delay is set to 100~ms for all the evaluated models to have a fair comparison of their robustness against asynchronous message propagation.

\noindent\textbf{Evaluation metrics.} 
The detection performance is measured with Average Precisions (AP) at Intersection-over-Union (IoU) thresholds of 0.5 and 0.7. In this work, we focus on LiDAR-based vehicle detection. Vehicles hit by at least one LiDAR point from any connected agent will be included as evaluation targets.

\noindent\textbf{Implementation details.}
During training, a random AV is selected as the ego vehicle, while during testing, we evaluate on a fixed ego vehicle for all the compared models.  The communication range of each agent is set as 70~m based on ~\cite{kenney2011dedicated}, whereas all the agents out of this broadcasting radius of ego vehicle is ignored. For the PointPillar backbone, we set the voxel resolution to 0.4~m for both height and width. The default compression rate is 32 for all intermediate fusion methods. Our V2X-ViT has 3 encoder layers with 3 window sizes in MSwin: 4, 8, and 16. We first train the model under the \textit{Perfect Setting}, then fine-tune it for \textit{Noisy Setting}. We adopt Adam optimizer~\cite{kingma2014adam} with an initial learning rate of $10^{-3}$ and steadily decay it every 10 epochs using a factor of 0.1. All models are trained on Tesla V100.

\noindent\textbf{Compared methods.} We consider \textit{No Fusion} as our baseline, which only uses ego-vehicle's LiDAR point clouds.
We also compare with \textit{Late Fusion}, which gathers all detected outputs from agents and applies Non-maximum suppression to produce the final results, and \textit{Early Fusion}, which directly aggregates raw LiDAR point clouds from nearby agents. For \textit{intermediate fusion} strategy, we evaluate four state-of-the-art approaches: OPV2V~\cite{xu2021opv2v}, F-Cooper~\cite{chen2019f} V2VNet~\cite{wang2020v2vnet}, and DiscoNet~\cite{li2021learning}.
For a fair comparison, all the models use PointPillar as the backbone, and every compared V2V methods also receive infrastructure data, but they do not distinguish between infrastructure and vehicles.

\begin{figure*}[!t]
\centering
\includegraphics[width=0.98\textwidth]{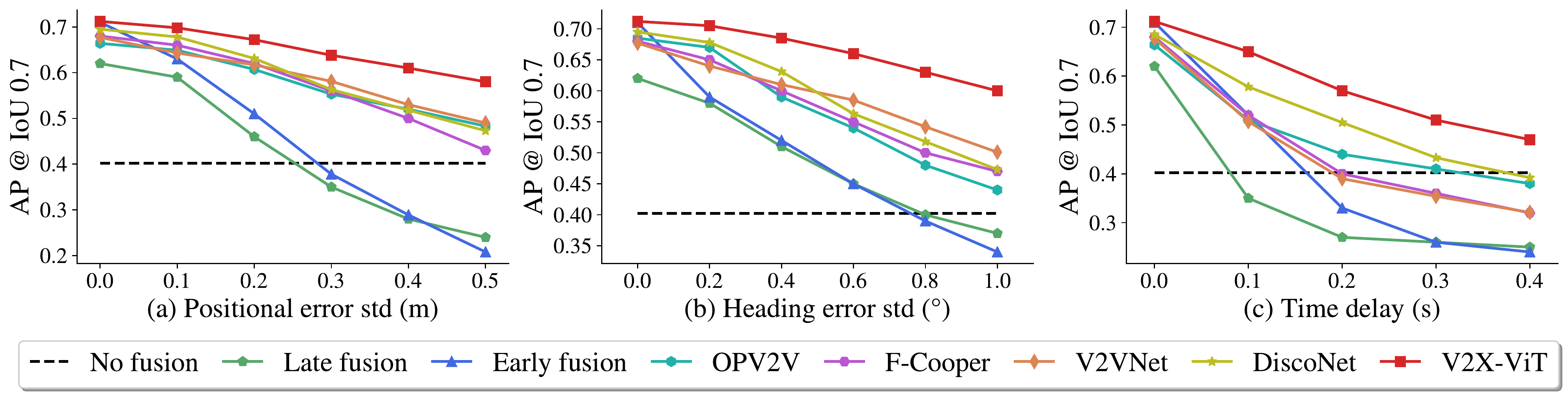}
\vspace{-2mm}
\caption{\textbf{Robustness assessment} on positional and heading errors.}
\label{fig:pose_error}
\vspace{-4mm}
\end{figure*}

\begin{table}[!t]
\centering
\scriptsize 
\setlength{\tabcolsep}{8pt}
\renewcommand{\arraystretch}{1.0}
\caption{\textbf{3D detection performance comparison on V2XSet.} We show Average Precision (AP) at IoU={0.5, 0.7} on \textit{Perfect} and \textit{Noisy} settings, respectively.}
\label{table:benchmark}
\begin{tabular}{lcccc}
\cellcolor{lightgray} & \multicolumn{2}{c}{\cellcolor{lightgray} Perfect}&\multicolumn{2}{c}{\cellcolor{lightgray} Noisy}\\ 
\cline{2-3}\cline{4-5}
{\cellcolor{lightgray} Models}
& \cellcolor{lightgray} AP0.5& \cellcolor{lightgray} AP0.7& \cellcolor{lightgray} AP0.5& \cellcolor{lightgray} AP0.7\\ \toprule
No Fusion           & 0.606          & 0.402          & 0.606          & 0.402          \\ 
Late Fusion          & 0.727          & 0.620          & 0.549          & 0.307          \\
Early Fusion         & 0.819          & 0.710          & 0.720          & 0.384          \\\midrule
F-Cooper~\cite{chen2019f}             & 0.840          & 0.680          & 0.715          & 0.469          \\
OPV2V~\cite{xu2021opv2v}                & 0.807          & 0.664          & 0.709          & 0.487          \\
V2VNet~\cite{wang2020v2vnet}                & 0.845          & 0.677          & 0.791        & 0.493          \\
DiscoNet~\cite{li2021learning}& 0.844 & 0.695 & 0.798& 0.541 \\ 
\midrule
V2X-ViT (Ours)          & \textbf{0.882} & \textbf{0.712} & \textbf{0.836} & \textbf{0.614} \\ 
\bottomrule
\end{tabular}
\vspace{-3mm}
\end{table}

\subsection{Quantitative evaluation}
\noindent\textbf{Main performance comparison.}
\cref{table:benchmark} shows the performance comparisons on both \textit{Perfect} and \textit{Noisy Setting}. 
Under the \textit{Perfect Setting}, all the cooperative methods significantly outperform \textit{No Fusion} baseline. Our proposed V2X-ViT outperforms SOTA intermediate fusion methods by 3.8\%/1.7\% for AP@0.5/0.7. It
is even higher than the ideal \textit{Early fusion} by 0.2\% AP@0.7, which receives complete raw information. Under noisy setting, when localization error and time delay are considered, the performance of \textit{Early Fusion} and \textit{Late Fusion} drastically drop to 38.4\% and 30.7\% in AP@0.7, even worse than single-agent baseline \textit{No Fusion}. Although OPV2V~\cite{xu2021opv2v}, F-Cooper~\cite{chen2019f} V2VNet~\cite{wang2020v2vnet}, and DiscoNet~\cite{li2021learning} are still higher than \textit{No fusion}, their performance decrease by 17.7\%, 21.1\%, 18.4\% and 15.4\% in AP@0.7, respectively. In contrast, V2X-ViT performs favorably against the \textit{No fusion} method by a large margin, \ie 23\% and 21.2\% higher in AP@0.5 and AP@0.7. Moreover, when compared to the \textit{Perfect Setting}, V2X-ViT only drops by less than 5\% and 10\% in AP@0.5 and AP@0.7 under \textit{Noisy Setting}, demonstrating its robustness against normal V2X noises. The real-time performance of V2X-ViT is also shown in~\cref{tbl:time}. The inference time of V2X-ViT is 57 ms, and by using only 1 encoder layer, V2X-ViT${}_S$ can still beat DiscoNet while reaching only 28 ms inference time, which achieves real-time performance.

\noindent\textbf{Sensitivity to localization error.}
To assess the models' sensitivity to pose error, we sample noises from Gaussian distribution with standard deviation $\sigma_{xyz}\in[0, 0.5]$~m, $\sigma_\mathrm{heading}\in[0\degree, 1.0\degree$].
As Fig.~\ref{fig:pose_error} depicts, when the positional and heading errors stay within a normal range~(\ie, $\sigma_{xyz} \leq 0.2m, \sigma_{heading} \leq 0.4\degree$~\cite{RT3000,li2020toward,xia2021advancing}), the performance of V2X-ViT only drops by less than 3\%, whereas other intermediate fusion methods decrease at least 6\%. Moreover, the accuracy of \textit{Early Fusion} and \textit{Late Fusion} degrade by nearly 20\% in AP@0.7. When the noise is massive (\eg, 0.5~m and 1\degree std), V2X-ViT can still stay around 60\% detection accuracy while the performance of other methods significantly degrades, showing the robustness of V2X-ViT against pose errors.

\noindent\textbf{Time delay analysis.}
We further investigate the impact of time delay with range $[0,400]$~ms. As \cref{fig:pose_error}c shows, the AP of \textit{Late Fusion} drops dramatically below \textit{No Fusion} with only 100~ms delay. \textit{Early Fusion} and other intermediate fusion methods are relatively less sensitive, but they still drop rapidly when delay keeps increasing and are all below the baseline after 400~ms. Our V2X-ViT, in contrast,  exceeds \textit{No Fusion} by 6.8\% in AP@0.7 even under 400~ms delay, which is much larger than usual transmission delay in real-world system\cite{tsukada2020autoc2x}. This clearly demonstrates its great robustness against time delay.

\begin{figure*}[!t]
\centering
\includegraphics[width=0.98\textwidth]{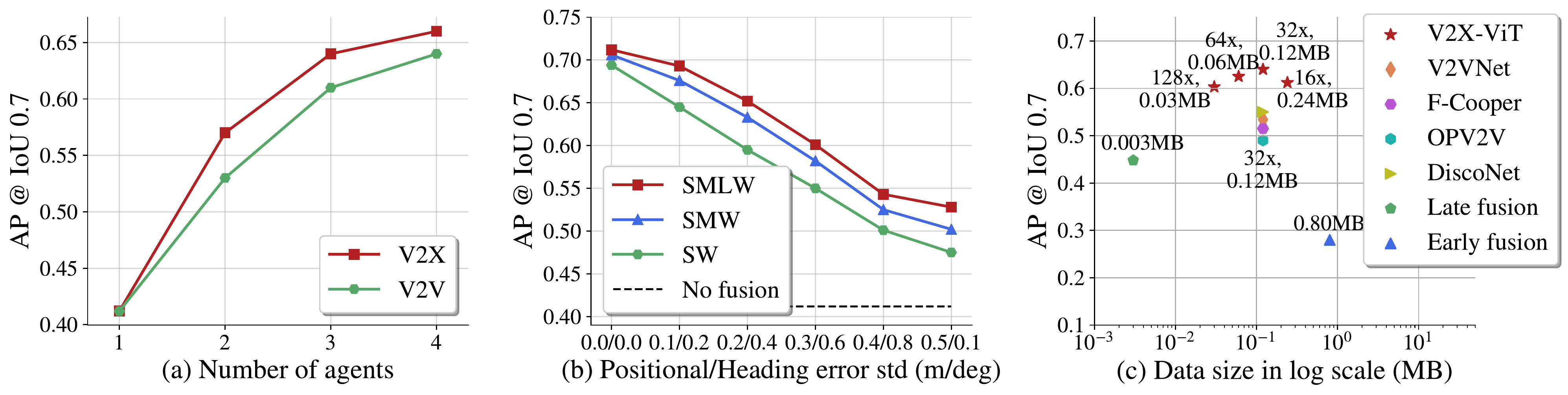}
\vspace{-2mm}
\caption{\textbf{Ablation studies.} (a) AP \vs number of agents. (b) MSwin for localization error with window sizes: $4^2$ (S), $8^2$ (M), $16^2$ (L). (c) AP \vs data size.}
\label{fig:ablation}
\vspace{-2mm}
\end{figure*}

\begin{table*}[!t]
\centering
\setlength{\tabcolsep}{1pt}
\begin{tabular}{@{}cc@{}}
\multirow{2}{*}{
\begin{minipage}[t]{0.54\textwidth}
\centering
\scriptsize
\setlength{\tabcolsep}{4pt}
\renewcommand{\arraystretch}{1.0}
\caption{\textbf{Component ablation study}. MSwin, SpAttn, HMSA, DPE represent adding i) multi-scale window attention, ii) split attention, iii) heterogeneous multi-agent self-attention, and iv) delay-aware positional encoding, respectively.}
\label{table:componet}
\begin{tabular}{ccccc}
\cellcolor{lightgray}  \rot{MSwin} &
 \cellcolor{lightgray} \rot{SpAttn} &
 \cellcolor{lightgray} \rot{HMSA} &
  \cellcolor{lightgray} \rot{DPE} &
 \cellcolor{lightgray}  AP0.5 / AP0.7 \\ \toprule
  &  &  &  & 0.719 / 0.478 \\
\cmark  &  &  &  & 0.748 / 0.519 \\
\cmark & \cmark &  &  & 0.786 / 0.548 \\
 \cmark & \cmark & \cmark &  & 0.823 / 0.601 \\
 \cmark & \cmark & \cmark & \cmark & \textbf{0.836} / \textbf{0.614 }\\ \bottomrule
\end{tabular}
\end{minipage}%
}
&
\hspace{3mm}
\begin{minipage}[t]{0.4\textwidth}
\centering
\scriptsize
\setlength{\tabcolsep}{3pt}
\renewcommand{\arraystretch}{1.0}
\caption{\textbf{Effect of DPE} w.r.t. time delay on AP@0.7.}
\label{tbl:dpe}
\begin{tabular}{ccc}
 \cellcolor{lightgray} {Delay}/{Model} &  \cellcolor{lightgray} w/o DPE &  \cellcolor{lightgray} w/ DPE \\
 \toprule
 100 ms & 0.639 & 0.650 \\
 200 ms & 0.558 & 0.572 \\
 300 ms & 0.496 & 0.514 \\
 400 ms & 0.458 & 0.478 \\
 \bottomrule
\end{tabular}
\end{minipage}%
\\
&
\hspace{3mm}
\begin{minipage}[t]{0.4\textwidth}
\centering
\scriptsize
\setlength{\tabcolsep}{3pt}
\renewcommand{\arraystretch}{1.0}
\caption{\textbf{Inference time} measured on GPU Tesla V100.}
\label{tbl:time}
\begin{tabular}{lcc}
 \cellcolor{lightgray} Model &  \cellcolor{lightgray} Time &  \cellcolor{lightgray} AP0.7(prf/nsy) \\
 \toprule
 V2X-ViT$_S$ & 28ms & 0.696 / 0.591 \\
 V2X-ViT & 57ms & 0.712 / 0.614 \\
 \bottomrule
\end{tabular}
\end{minipage}%
\end{tabular}
\vspace{-4mm}
\end{table*}

\noindent\textbf{Infrastructure \vs vehicles. }To analyze the effect of infrastructure in the V2X system, we evaluate the performance between V2V, where only vehicles can share information, and V2X, where infrastructure can also transmit messages. We denote the number of agents as the total number of infrastructure and vehicles that can share information.
As shown in \cref{fig:ablation}a, both V2V and V2X have better performance when the number of agents increases. The V2X system has better APs compared with V2V in our collected scenes. We argue this is due to the better sight-of-view and less occlusion of infrastructure sensors, leading to more informative features for reasoning the environmental context. 

\noindent\textbf{Effects of transmission size.}
The size of the transmitted message can significantly affect the transmission delay, thereby affecting the detection performance. Here we study the model's detection performance with respect to transmitted data size. The data transmission time is calculated by  $t_c = f_s/v $,  where $f_s$ denotes the feature size and transmission rate $v$ is set to 27~Mbps~\cite{arena2019overview}.  Following~\cite{rauch2011analysis}, we also include another system-wise asynchronous delay that follows a uniform distribution between 0 and 200~ms. See supplementary materials for more details.  From \cref{fig:ablation}c, we can observe: 1) Large bandwidth requirement can eliminate the advantages of cooperative perception quickly, \eg, \textit{Early Fusion} drops to 28\%, indicating the necessity of compression; 2) With the default compression rate~(32x), our V2X-ViT outperforms other intermediate fusion methods substantially; 3) V2X-ViT is insensitive to large compression rate. Even under a 128x compression rate, our model can still maintain high performance.

\begin{figure*}[!t]
\centering
    \begin{subfigure}[c]{0.24\linewidth}
        \centering{\includegraphics[width=1\linewidth]{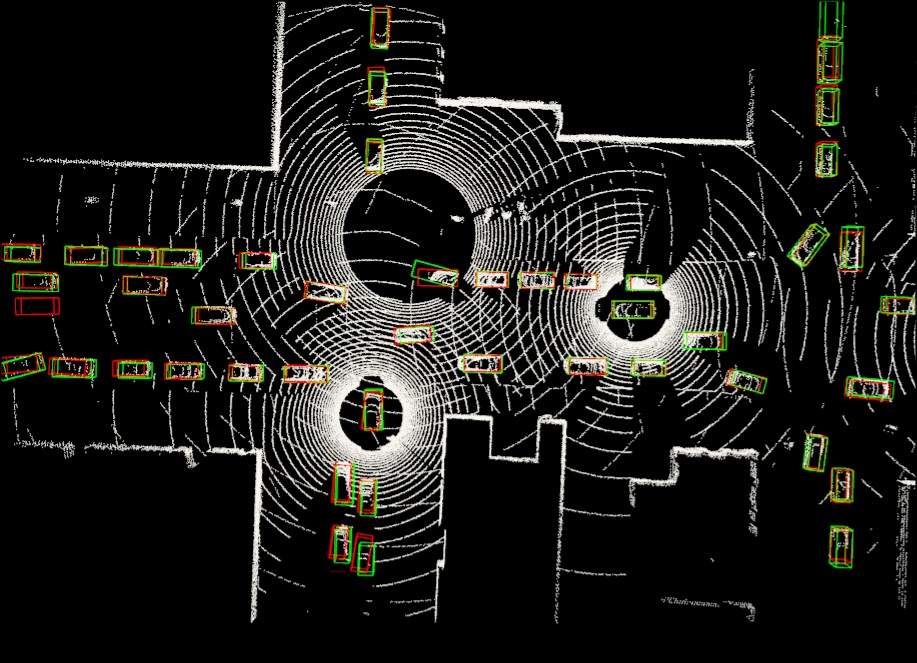}}
        \centering{\includegraphics[width=1\linewidth]{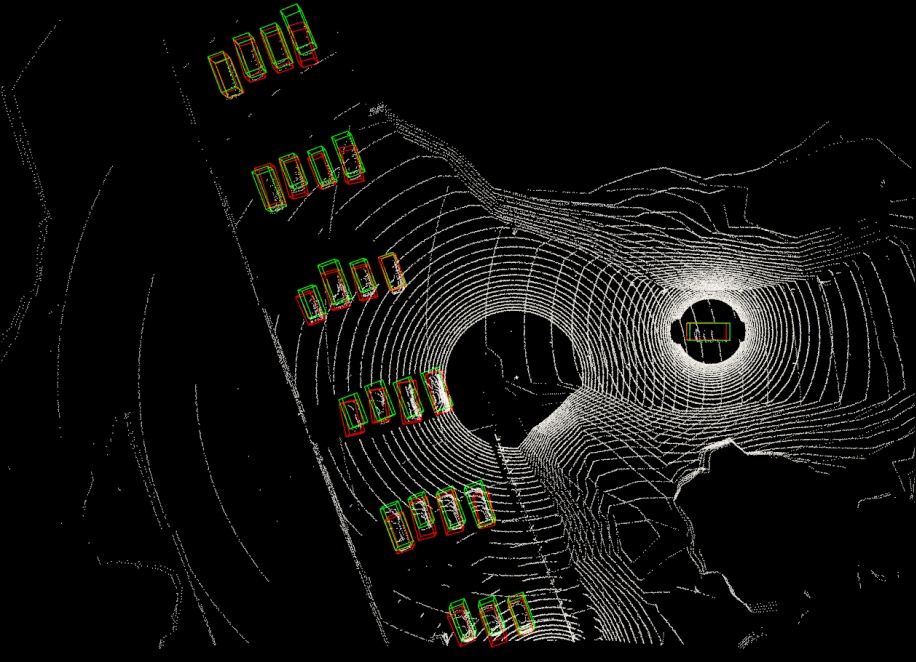}}
        \caption{OPV2V~\cite{xu2021opv2v}}
        \label{fig:output-a}
    \end{subfigure}
    \begin{subfigure}[c]{0.24\linewidth}
        \centering{\includegraphics[width=1\linewidth]{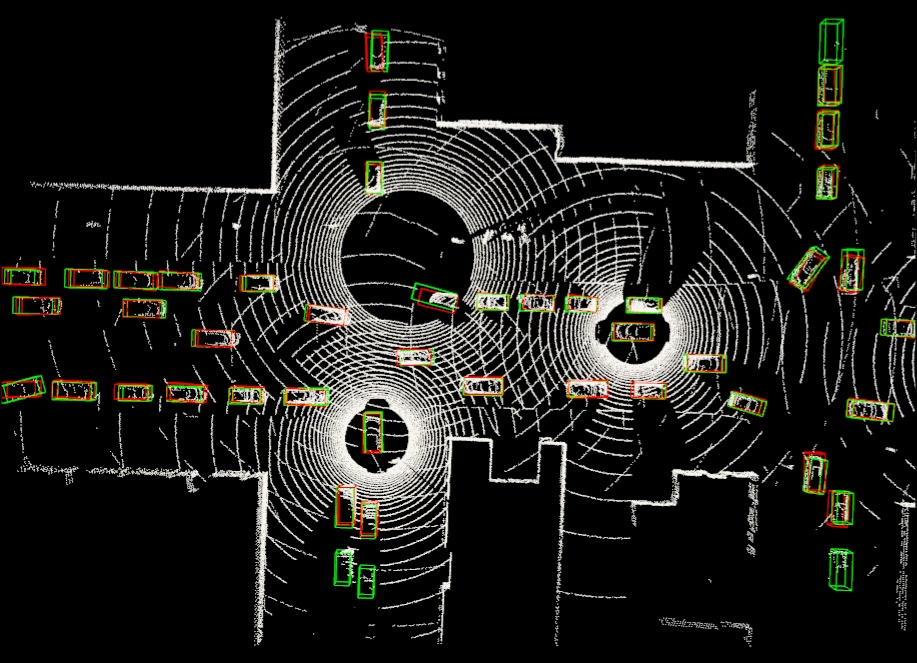}}
        \centering{\includegraphics[width=1\linewidth]{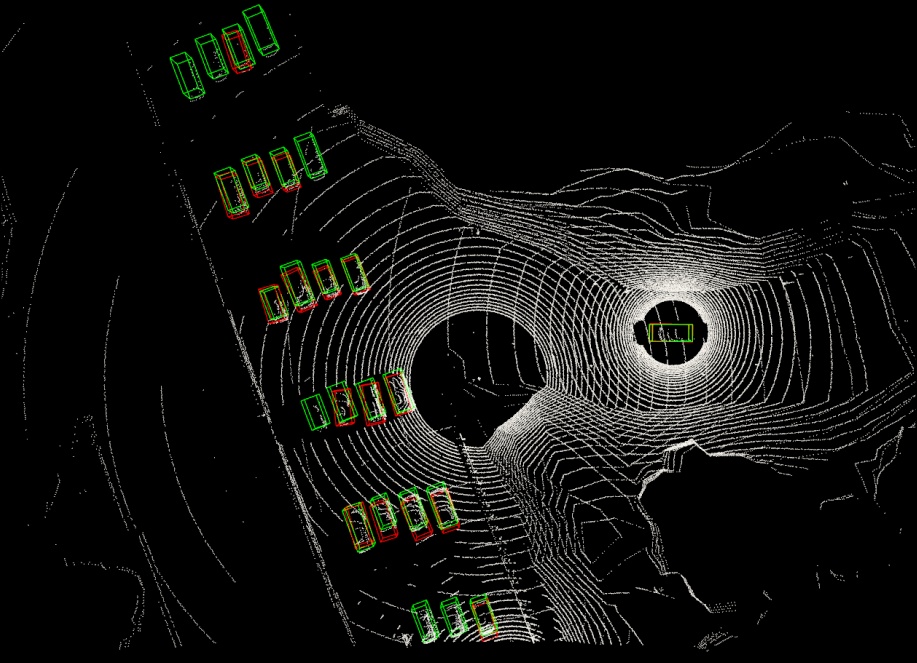}}
        \caption{V2VNet~\cite{wang2020v2vnet}}
        \label{fig:output-b}
    \end{subfigure}
    \begin{subfigure}[c]{0.24\linewidth}
        \centering{\includegraphics[width=1\linewidth]{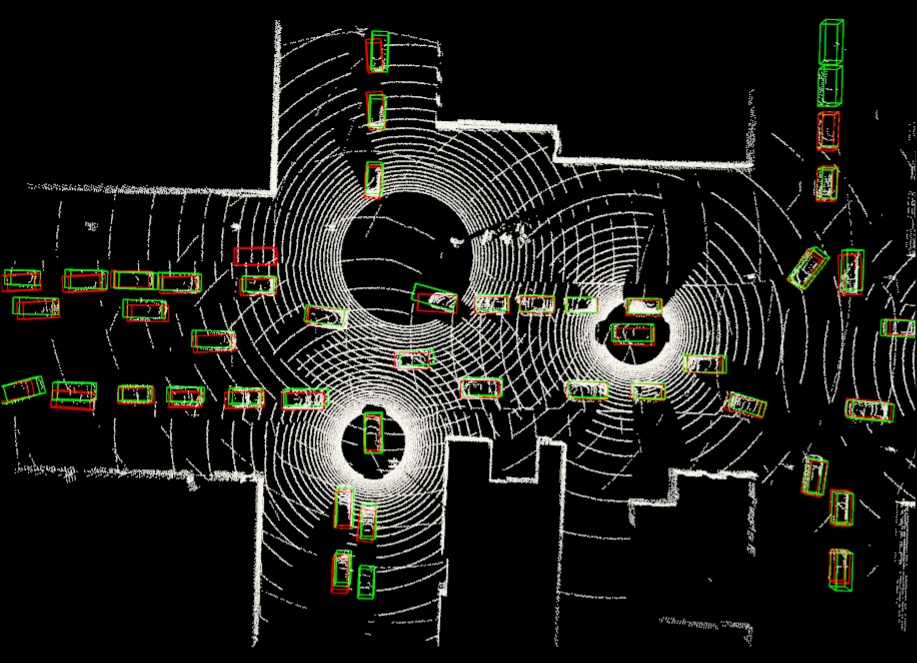}}
        \centering{\includegraphics[width=1\linewidth]{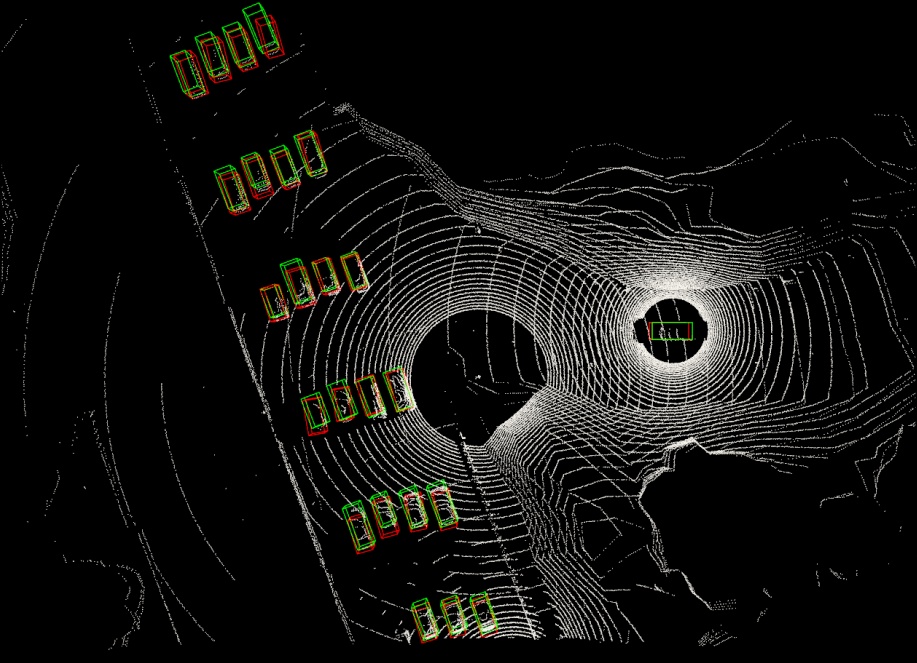}}
        \caption{DiscoNet~\cite{li2021learning}}
        \label{fig:output-c}
    \end{subfigure}
    \begin{subfigure}[c]{0.24\linewidth}
        \centering{\includegraphics[width=1\linewidth]{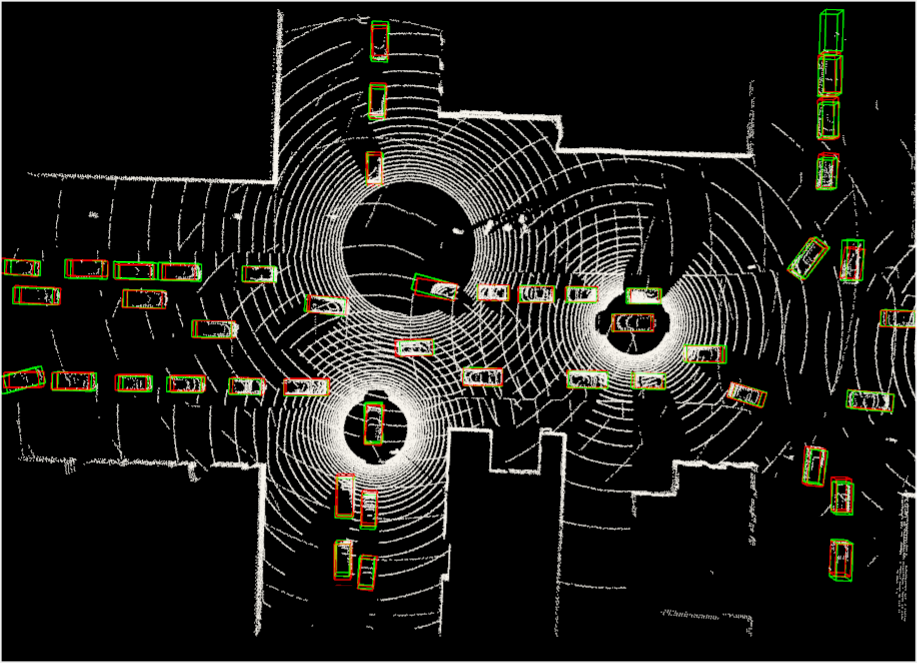}}
        \centering{\includegraphics[width=1\linewidth]{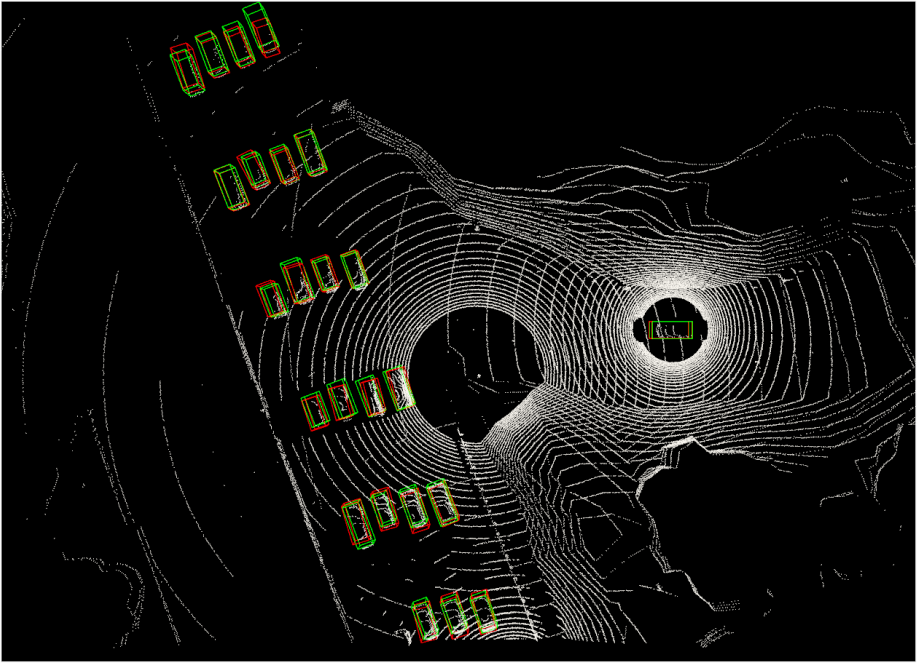}}
        \caption{V2X-ViT (ours)}
        \label{fig:output-d}
    \end{subfigure}
    \vspace{-2mm}
    \caption{\textbf{Qualitative comparison in a congested intersection and a highway entrance ramp.} \textcolor{green}{Green} and \textcolor{red}{red} 3D bounding boxes represent the ground truth and prediction respectively. Our method yields more accurate detection results. More visual examples are provided in the supplementary materials.}
    \label{fig:qualitive}
    \vspace{-2mm}
\end{figure*}

\begin{figure}[!tb]
     \centering
     \begin{subfigure}[b]{0.24\textwidth}
         \centering
         \includegraphics[width=\textwidth]{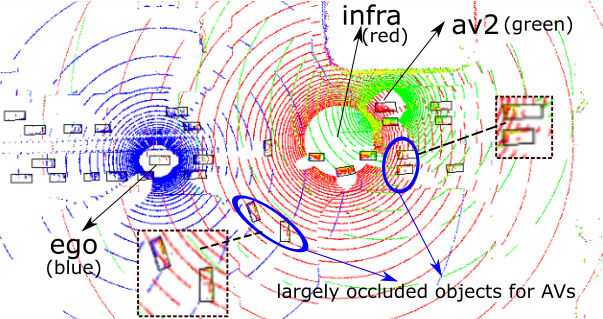}
         \caption{LiDAR points (better zoom-in)}
         \label{fig:att_lidar}
     \end{subfigure}
     \begin{subfigure}[b]{0.24\textwidth}
         \centering
         \includegraphics[width=\textwidth]{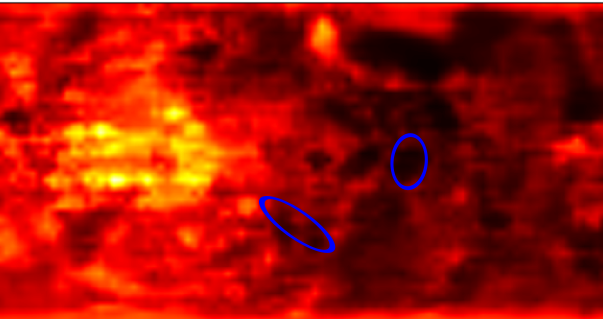}
         \caption{attention weights ego paid to ego}
         \label{fig:att_ego}
     \end{subfigure}
     \begin{subfigure}[b]{0.24\textwidth}
         \centering
         \includegraphics[width=\textwidth]{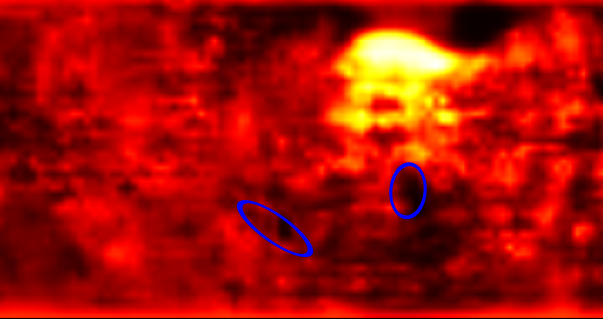}
         \caption{attention weights ego paid to av2}
         \label{fig:att_cav}
     \end{subfigure}
     \begin{subfigure}[b]{0.24\textwidth}
         \centering
         \includegraphics[width=\textwidth]{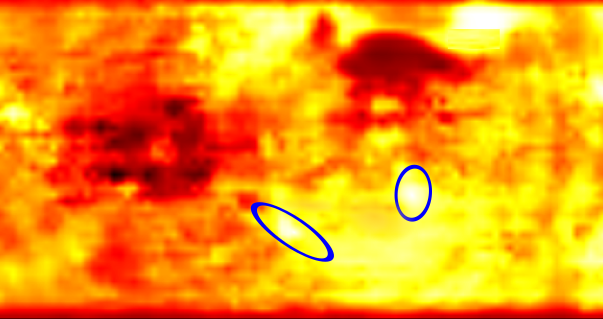}
         \caption{attention weights ego paid to infra}
         \label{fig:att_infra}
     \end{subfigure}
     
    \vspace{-2mm}
     \caption{\textbf{Aggregated LiDAR points and  attention maps for ego.} Several objects are occluded (\textcolor{blue}{blue} circle) from both AV's perspectives, whereas infra can still capture rich point clouds. V2X-ViT learned to pay more attention to infra on occluded areas, shown in (d). We provide more visualizations in Appendix.}
     \label{fig:infra}
     \vspace{-4mm}
\end{figure}
\subsection{Qualitative evaluation}
\noindent\textbf{Detection visualization.} \cref{fig:qualitive} shows the detection visualization of  OPV2V, V2VNet, DiscoNet, and V2X-ViT in two challenging scenarios under \textit{Noisy setting}. Our model predicts highly accurate bounding boxes which are well-aligned with ground truths, while other approaches exhibit larger displacements. More importantly, V2X-ViT can identify more dynamic objects~(more ground-truth bounding boxes have matches), which proves its capability of effectively fusing all sensing information from nearby agents. Please see Appendix for more results.

\noindent\textbf{Attention map visualization. } To understand the importance of infra, we also visualize the learned attention maps in~\cref{fig:infra}, where brighter color means more attention ego pays. As shown in \cref{fig:att_lidar}, several objects are largely occluded~(circled in \textcolor{blue}{blue}) from both ego and AV2's perspectives, whereas infrastructure can still capture rich point clouds. Therefore, V2X-ViT pays much more attention to infra on occluded areas~(\cref{fig:att_infra}) than other agents~(\cref{fig:att_ego,fig:att_cav}), demonstrating the critical role of infra on occlusions. Moreover, the attention map for infra is generally brighter than the vehicles, indicating more importance on infra seen by the trained V2X-ViT model.
\subsection{Ablation studies}
\noindent\textbf{Contribution of major components in V2X-ViT.}
Now we investigate the effectiveness of individual components in V2X-ViT. Our base model is PointPillars with naive multi-agent self-attention fusion, which treats vehicles and infrastructure equally. We evaluate the impact of each component by progressively adding i) MSwin, ii) split attention, iii) HMSA, and iv) DPE on the \textit{Noisy Setting}. As~\cref{table:componet} demonstrates, all the modules are beneficial to the performance gains, while our proposed MSwin and HMSA have the most significant contributions by increasing the AP@0.7 4.1\% and 6.6\%, respectively. 

\noindent\textbf{MSwin for localization error.}
To validate the effectiveness of the multi-scale design in MSwin on localization error, we compare three different window configurations: i) using a single small window branch~(SW), ii) using a small and a middle window~(SMW), and iii) using all three window branches~(SMLW). We simulate the localization error by combining different levels of positional and heading noises. From \cref{fig:ablation}b, we can clearly observe that using a large and small window in parallel remarkably increased its robustness against localization error, which validates the design benefits of MSwin.

\noindent\textbf{DPE Performance under delay.}
\cref{tbl:dpe} shows that DPE can improve the performance under various time delays. The AP gain increases as delay increases. 

\section{Conclusion}
In this paper, we propose a new vision transformer (V2X-ViT) for V2X perception. Its key components are two novel attention modules \ie HMSA and MSwin, which can capture heterogeneous inter-agent interactions and multi-scale intra-agent spatial relationship. 
To evaluate our approach, we construct V2XSet, a new large-scale V2X perception dataset.
Extensive experiments show that V2X-ViT can significantly boost cooperative 3D object detection under both perfect and noisy settings. 
This work focuses on LiDAR-based cooperative 3D vehicle detection, limited to single sensor modality and vehicle detection task.
Our future work involves multi-sensor fusion for joint V2X  perception and prediction.

\vspace{2mm}
\noindent\textbf{Broader impacts and limitations.} The proposed model can be deployed to improve the performance and robustness of autonomous driving systems by incorporating V2X communication using a novel vision Transformer. 
However, for models trained on simulated datasets, 
there are known issues on data bias and generalization ability to real-world scenarios. 
Furthermore, although the design choice of our communication approach~(i.e., project LiDAR to others at the beginning) has an advantage of accuracy~(see supplementary for details), its scalability is limited.
In addition, new concerns around privacy and adversarial robustness may arise during data capturing and sharing, which has not received much attention. 
%
This work facilitates future research on fairness, privacy, and robustness in visual learning systems for autonomous vehicles.  

\vspace{2mm}
\noindent\textbf{Acknowledgement.} This material is supported in part by the
Federal Highway Administration Exploratory Advanced Research (EAR) Program, and by the US National Science Foundation through Grants CMMI \# 1901998. We thank Xiaoyu Dong for her insightful discussions.


\clearpage
%
%

\appendix
\section*{Appendix}
In this supplementary material, we first discuss the design choice and scalability issue of our method (\cref{sec:design-choice}). Then, we provide more model details and analysis (\cref{sec:model-details}) in regards to the proposed architecture, including the mathematical details of the spatial-temporal correction module (\cref{sec:stcm}), details of the proposed MSwin (\cref{sec:mswin}), and the overall architectural specifications (\cref{ssec:arch-config}). Afterwords, additional information and visualizations of the proposed V2XSet dataset are shown in~\cref{sec:opv2x-dataset}. In the end, we present more quantitative experiments, qualitative detection results, attention map visualizations, and  details about the effects of the transmission size experiment in~\cref{sec:additional-results}.

\section{Discussion of design choice}
\label{sec:design-choice}
\noindent\textbf{Scalability of ego vehicles.}
Our approach can be scalable in two ways: 1) {De-centralized}: the ablation studies conducted in this paper~(Fig. 5a) and OPV2V~\cite{xu2021opv2v} indicate that, when the number of collaborators is larger than 4, the performance gain becomes marginal while the computation still increases linearly. In practice, each agent only needs to share features with a limited number of agents. For example, Who2Com~\cite{liu2020who2com} studies which agent to request/transmit data, largely reducing computation. Moreover, the computation of selected PointPillar backbone is efficient, e.g., around 4 ms for 4 agents with full parallelization and 16~ms in sequence computing on RTX3090. 2) {Centralized}: Within a certain communication range, only one ego agent is selected to aggregate all the features from neighbors to predict bounding boxes and share the results with other agents. This solution requires only one computation node for a group of agents, thus being scalable. 
\begin{table}[h]
\vspace{-4mm}
    \centering
    \footnotesize
    \def\xwidth{0.245}
    \renewcommand\thetable{T0}
    \caption{Comparison between our design choice and broadcasting approach.}
    \setlength{\tabcolsep}{2pt}
    \begin{tabular}{c|c|c}
         \cellcolor{lightgray}{} &\cellcolor{lightgray}{DiscoNet (broad. / ours)}&\cellcolor{lightgray}{V2X-ViT (broad. / ours)}  \\
        \toprule
         AP@0.7 (perfect)&  0.610 / 0.695 & 0.623 / 0.712  \\
         \bottomrule
    \end{tabular}
    \label{tab:design_choice}
\vspace{-3mm}
\end{table}

\noindent\textbf{Design choices for communication. }
Compared to the broadcasting approach~(\ie, compute the features in each cav's own space and transform the feature maps directly on the ego side), our approach has more advantages in terms of detection accuracy. Most LiDAR detection methods often largely crop the LiDAR range based on the evaluation range to reduce computation. As the figure below shows, the CAVs crop the LiDAR data based on their own evaluation range in the broadcasting method, which leads to redundant data. Our approach, on the contrary, always does cropping based on the ego's evaluation range, thus guaranteeing more effective feature transmission. We further validate this by comparing our framework with the broadcasting approach. The Tab.~\ref{tab:design_choice} below shows that our design outperforms broadcasting by 8.5\% and 8.9\% for DiscoNet and V2X-ViT.
\begin{figure}[h]
\centering
\includegraphics[width=0.9\linewidth]{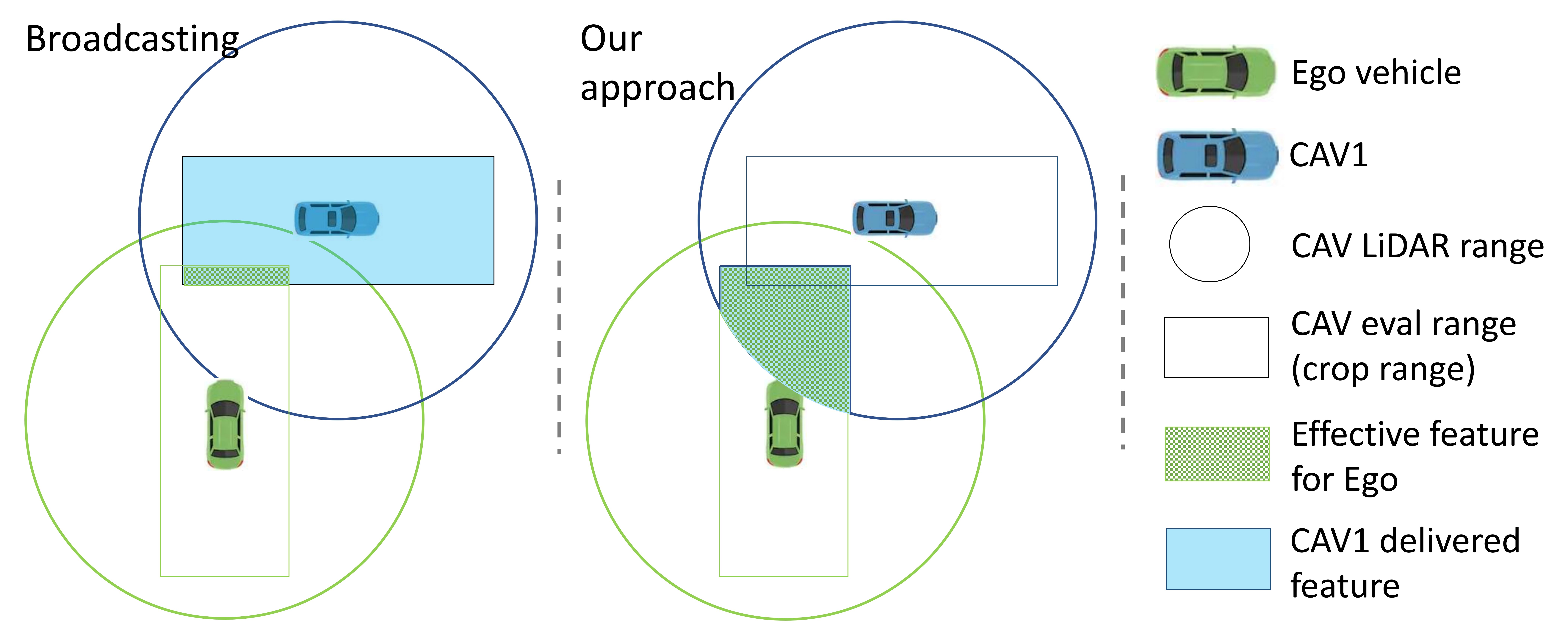}
\label{fig:broadcast}
\end{figure}

\section{Model Details and Analysis}
\label{sec:model-details}

\subsection{Spatial-Temporal Correction Module}
\label{sec:stcm}
During the early stage of collaboration, when each connected agent $i$ receives ego vehicle's pose at time $t_i$, the observed point clouds of agent $i$ will be projected to ego vehicle's pose $x_{e}^{t_i}$ at time $t_i$ before feature extraction. However, due to the time delay, the ego vehicle observes the data at a different time $t_e$. Thus, the received features from connected agents are centered around a delayed ego vehicle's pose (\ie, $x_{e}^{t_i}$) while the ego vehicle's features are centered around current pose (\ie, $x_{e}^{t_e}$), leading to a delay-induced spatial misalignment. To correct this misalignment between the received features and ego-vehicle's features, a global transformation ${\xi}_{x_{e}^{t_i}, x_{e}^{t_e}}\in \mathfrak{s e}(3)$ from ego vehicle's past pose $x_e^{t_i}$ to its current pose $x_e^{t_e}$ is required. To this end, we employ a differential 2D transformation ${\Gamma}_{\xi}\left(\cdot\right)$ to warp the intermediate features spatially~\cite{jaderberg2015spatial}. To be more specific, we will transform features' positions by using affine transformation:
\begin{equation}
    \begin{bmatrix}
    x_s\\y_s
    \end{bmatrix}={\Gamma}_{\xi}(
    \begin{bmatrix}
    x_t\\y_t\\1
    \end{bmatrix})\\
    =\begin{bmatrix}
    R_{11}&R_{12}&\delta x\\
    R_{21}&R_{22}&\delta y\\
    \end{bmatrix}\begin{bmatrix}
    x_t\\y_t\\1
    \end{bmatrix}
\end{equation}
where $(x_s,y_s)$ and $(x_t,y_t)$ are the source and target coordinates. As the calculated coordinates may not be integers, we use bilinear interpolation to sample input feature vectors. An ROI mask is also calculated to prevent the network from paying attention to the padded zeros caused by the spatial warp. This mask will be used in heterogeneous multi-agent self-attention to mask padded values' attention weights as zeros. 

\subsection{Multi-Scale Window Attention (MSwin)}
\label{sec:mswin}
\noindent\textbf{Detailed formulation.}
Let $\mathbf{H}\in\mathbb{R}^{H\times W\times C}$ be an input feature of a single agent. Let $h_j$ be the number of attention heads used in branch $j$ (\ie head dimension $d_{h_j}=C/h_j$), applying self-attention within each non-overlapping window $P_j\times P_j$ for branch $j$ out of $k$ branches on feature $\mathbf{H}$ can be formulated as:
\begin{align}
\label{eq:mswin-block}
\mathbf{H}&=[\mathbf{H}^1,\mathbf{H}^2,...,\mathbf{H}^{HW/(P_j)^2}],\ &\text{for branch}\ j\\
\hat{\mathbf{H}}^i_{m}&=\mathsf{Attention}(\mathbf{H}^i\mathbf{W}^Q_{m},\mathbf{H}^i\mathbf{W}^K_{m},\mathbf{H}^i\mathbf{W}^V_{m}),\ &i=1,...,HW/(P_j)^2
\\
\mathbf{Y}_{m}&=[\hat{\mathbf{H}}^1_{m},\hat{\mathbf{H}}^2_{m},...,\hat{\mathbf{H}}^{HW/(P_j)^2}_{m}], &m=1,...,h_j
\\
\mathbf{Y}^{j}&=[\mathbf{Y}_{1},\mathbf{Y}_{2},...,\mathbf{Y}_{h_j}],
\end{align}
where $\hat{\mathbf{H}}^i_m\in\mathbb{R}^{P_j^2\times d_{h_j}}$ and $\mathbf{W}^Q_m,\mathbf{W}^K_m,\mathbf{W}^V_m$ represent the query, key, and value projection matrices. $\mathbf{Y}_{m}$ is the output of the $m$-th head for branch $j$. Afterwards, the outputs for all heads $1,2,...,h_j$ are concatenated to obtain the final output $\mathbf{Y}^j$. Here the $\mathsf{Attention}$ operation denotes the relative self-attention, similar to the usage in Swin~\cite{liu2021swin}:
\vspace{-2mm}
\begin{equation}
\label{eq:attention}
\mathsf{Attention}(\mathbf{Q},\mathbf{K},\mathbf{V})=\mathsf{softmax}((\frac{\mathbf{Q}\mathbf{K}^T}{\sqrt{d}}+\mathbf{B})\mathbf{V})
\vspace{-2mm}
\end{equation}
where $\mathbf{Q},\mathbf{K},\mathbf{V}\in\mathbb{R}^{P_j^2\times d}$ denote the query, key, and value matrices. $d$ is the dimension of query/key, while $P_j^2$ denotes the window size for branch $j$.
Following~\cite{liu2021swin,hu2019local}, we also consider an additional relative positional encoding $\mathbf{B}$ that acts as a bias term added to the attention map. As the relative position along each axis lies in the range $[-P_j+1,P_j-1]$,  we take $\mathbf{B}$ from a parameterized matrix $\hat{\mathbf{B}}\in\mathbb{R}^{(2P_j-1)\times(2P_j-1)}$.
To adaptively fuse features from all the $k$ branches, we adopt the split-attention module~\cite{zhang2020resnest} for the parallel feature aggregation:
\begin{equation}
\label{eq:split-attention}
\mathbf{Y}=\mathsf{SplitAttention}(\mathbf{Y}^{1},\mathbf{Y}^{2},...\mathbf{Y}^{k}),
\end{equation}

\noindent\textbf{Time complexity. }As mentioned in the paper, we have $k$ parallel branches. Each branch has $P_j\times P_j$ window size and $h_j$ heads where $P_j=jP$ and $h_j=h/j$. After partitioning, the input tensor $\mathbf{H}\in\mathbb{R}^{H\times W\times C}$ is split into $h_j$ features with shape $(\frac{H}{P_j}\times\frac{W}{P_j},P_j\times P_j,C/h_j)$. Following~\cite{liu2021swin}, we focus on the computation for vector-matrix multiplication and attention weight calculation. Thus, the complexity of MSwin can be written as:
\begin{align*}
    &\mathcal{O}(\sum_{j=1}^k \frac{HW}{P_j^2}\times\frac{C}{h_j}\times (P_j\times P_j)^2\times h_j+4\frac{HW}{P_j^2}\times P_j^2\times (\frac{C}{h_j})^2\times h_j)\\
    &=\mathcal{O}(\sum_{j=1}^k P_j^2HWC+\frac{4HWC^2}{h_j})=\mathcal{O}(\sum_{j=1}^k j^2P^2HWC+\frac{4HWC^2j}{h})\\
    &=\mathcal{O}(\frac{1}{3}k^3P^2HWC+\frac{2HWC^2k^2}{h}) \numberthis \label{0}
\end{align*}
where the first term corresponds to attention weight calculation, the second term is associated with vector-matrix multiplication, and the last equality is due to the fact that $\sum_{j=1}^k j^2=\mathcal{O}(\frac{k^3}{3})$ and $\sum_{j=1}^k j=\mathcal{O}(\frac{k^2}{2})$. Thus the overall complexity is 
\begin{equation}
\label{eq:mswin-complexity}
\mathrm{FLOPs}(\mathsf{MSwin})=\mathcal{O}((\frac{k^3P^2C}{3}+\frac{2k^2C^2}{h})HW)\sim \mathcal{O}(HW),
\end{equation}
which is linear with respect to the image size. The comparison of time complexity of different types of transformers is shown in \cref{tab:complexity-compare} where $N$ denotes the number of input pixels, or  (here $N=HW$). Our MSwin obtains multi-scale spatial interactions with a linear complexity with respect to $N$, while other long-range attention mechanisms like ViT~\cite{dosovitskiy2020image}, Axial~\cite{wang2020axial}, and CSwin~\cite{dong2021cswin} requires more than linear complexity, which are not scalable to high-resolution dense prediction tasks such as object detection and segmentation.

\begin{table}[!ht]
\centering
\footnotesize
\def\xwidth{0.245}
\renewcommand\thetable{T1}
    \caption{Computational complexity comparisons of our proposed MSwin attention with (a) full attention in ViT~\cite{dosovitskiy2020image}, (b) Axial~\cite{wang2020axial}, (c) Swin~\cite{liu2021swin}, (d) CSwin~\cite{dong2021cswin}. }
    \label{tab:complexity-compare}
\setlength{\tabcolsep}{8pt}
    \begin{tabular}{l|c}
 \cellcolor{lightgray}  Attention Models &  \cellcolor{lightgray}  Complexity  \\
 \toprule
ViT~\cite{dosovitskiy2020image} & $\mathcal{O}(4HWC^2 + 2(HW)^2C)\sim \mathcal{O}(N^2)$ \\
Axial~\cite{wang2020axial} & $\mathcal{O}(HWC(4C+H+W))\sim \mathcal{O}(N\sqrt{N})$ \\
Swin~\cite{liu2021swin} & $\mathcal{O}(4HWC^2 + 2P^2HWC)\sim \mathcal{O}(N)$  \\
CSwin~\cite{dong2021cswin} & $\mathcal{O}(HWC(4C+sH+sW))\sim \mathcal{O}(N\sqrt{N})$\\
\hline
MSwin (ours) &  $\mathcal{O}(\frac{1}{3}k^3P^2HWC+\frac{2HWC^2k^2}{h})\sim \mathcal{O}(N)$ \\
\bottomrule
\end{tabular}
\vspace{-5mm}
\end{table}

\begin{figure}[!ht]
\centering
\includegraphics[width=0.96\columnwidth]{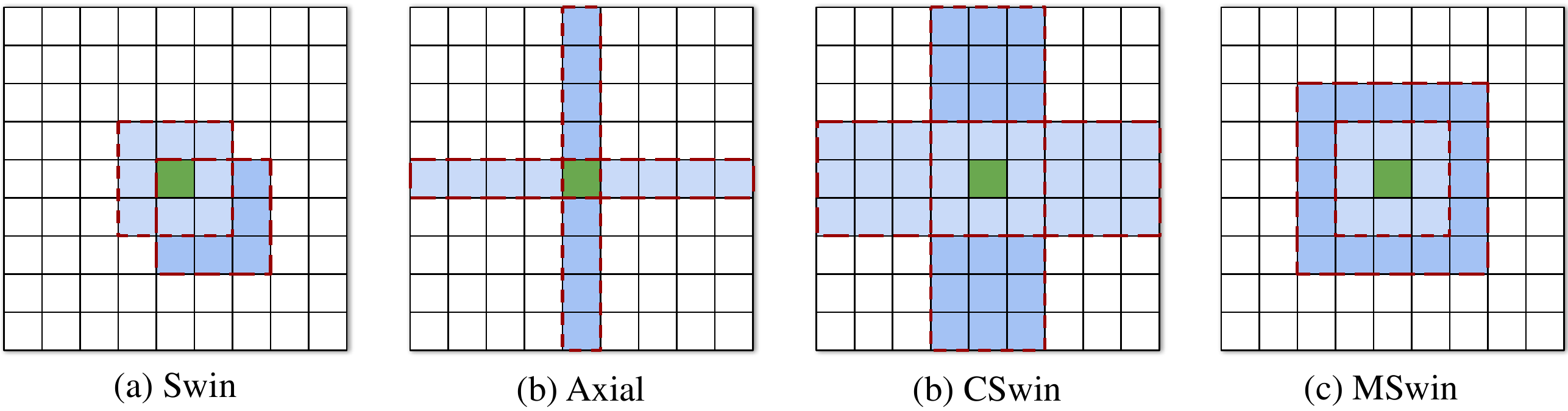}
\caption{Visualizations of approximated receptive fields (\textcolor{blue}{blue} shaded pixels) for the \textcolor{green}{green} pixel for (a) Swin~\cite{liu2021swin} (b) Axial~\cite{wang2020axial}, (c) CSwin~\cite{dong2021cswin} and (d) MSwin attention. MSwin obtains multi-scale long-range interactions at linear complexity.}
\label{fig:visualization-mswin}
\vspace{-5mm}
\end{figure}

\noindent\textbf{Effective receptive field. }The comparisons of receptive fields between different transformers are shown in \cref{fig:visualization-mswin}. Swin~\cite{liu2021swin} enlarge the receptive fields by using shifted window but it requires sequential blocks to accumulate. Axial Transformer~\cite{wang2020axial} conducts attention on both row-wise and column-wise directions. Similarly, CSwin~\cite{dong2021cswin} proposes to perform attention on horizontal and vertical stripes with asymmetrical receptive range in different directions, but requires polynomial time complexity--$\mathcal{O}(N^{1.5})$. In contrast, our proposed MSwin can aggregate features from multi-scale branches to increase fields in parallel, which has more symmetrical receptive fields and linear complexity with respect to $N$.

\subsection{Architectural Configurations}
\label{ssec:arch-config}

Given all these definitions, the entire V2X-ViT model can be formulated as:
\begin{align}
\label{eq:tfm-block}
{z}_i & = \text{PointPillar}({x}_i),  \quad\quad\quad\quad\quad  {x}_i\in\mathbb{R}^{P\times 4},{z_i}\in\mathbb{R}^{H\times W\times C}\ & \text{for agent}\ i  \\
\mathbf{z}_0 & = \text{STCM}([z_0,...,z_M])+\text{DPE}([\Delta t_0,...,\Delta t_M]),  & \text{for ego AV}  \\
\mathbf{z}'_\ell & = \mathbf{z}_{\ell-1} + \text{MSwin}(\text{HSMA}(\text{LN}(\mathbf{z}_0))),\quad \mathbf{z}_0\in\mathbb{R}^{M\times H\times W\times C}  & \ell=1,...,L \\
\mathbf{z}_\ell & = \mathbf{z}'_\ell + \text{MLP}(\text{LN}(\mathbf{z}'_\ell)), & \ell=1,...,L \\
\mathbf{y}&=\text{Head}(\mathbf{z}_L),
\end{align}
where the input ${x}_i$ denotes the raw LiDAR point clouds captured on each agent, which are fed into the PointPillar Encoder~\cite{lang2019pointpillars}, yielding visually informative 2D features ${z}_i$ for each agent $i$. These tensors are then compressed, shared,  decompressed, and further fed into the spatial-temporal correction module~(STCM) to spatially warp the features. Then, we add the delay-aware positional encoded (DPE) features conditioned on each agent's time delay $\Delta t_i$ to the output of STCM. Afterwords, the gathered features from $M$ agents are processed using our proposed V2X-ViT, which consists of $L$ layers of V2X-ViT blocks. Each V2X-ViT block contains a HSMA, a MSwin, and a standard MLP network~\cite{dosovitskiy2020image}. Following~\cite{dosovitskiy2020image,liu2021swin}, we use the Layer Normalization~\cite{ba2016layer} before feeding into the Transformer/MLP module. We show the detailed specifications of V2X-ViT architecture in Table~\ref{tab:arch-spec}.

\begin{table}[!t]
\centering
\footnotesize
\setlength{\tabcolsep}{2pt}
\renewcommand\thetable{T2}
\caption{\textbf{Detailed architectural specifications for V2X-ViT.}}
\label{tab:arch-spec}
\begin{tabular}{c|c|c}
 \cellcolor{lightgray} &  \cellcolor{lightgray} Output size &  \cellcolor{lightgray} V2X-ViT framework  \\
 \toprule
\multirow{4}{*}{\begin{tabular}{c}PointPillar\\ Encoder\\
\end{tabular}} & $M\times 352\times 96\times 256$ & 
\begin{tabular}{c}
$\left[ \begin{array}{c} \text{Voxel samp. reso. 0.4m, Scatter, 64}\end{array}\right]$ \\
$\left[ \begin{array}{c} \text{Conv3x3, 64, stride 2, BN, ReLU}\end{array}\right] \times 3$ \\
$\left[ \begin{array}{c} \text{Conv3x3, 128, stride 2, BN, ReLU}\end{array}\right] \times 5$ \\
$\left[ \begin{array}{c} \text{Conv3x3, 256, stride 2, BN, ReLU}\end{array}\right] \times 8$ \\
$\left[ \begin{array}{c} \text{ConvT3x3, 128, stride 1, BN, ReLU}\end{array}\right] \times 1$ \\
$\left[ \begin{array}{c} \text{ConvT3x3, 128, stride 2, BN, ReLU}\end{array}\right] \times 1$ \\
$\left[ \begin{array}{c} \text{ConvT3x3, 128, stride 4, BN, ReLU}\end{array}\right] \times 1$
\end{tabular}
\\ \cline{2-3}

&  $M\times 176\times 48\times 256$ & 
\begin{tabular}{c}
$\left[ \begin{array}{c} \text{Concat3, 384}\end{array}\right]$  \\
$\left[ \begin{array}{c} \text{Conv3x3, 256, stride 2, ReLU} \\
\text{Conv3x3, 256, stride 1, ReLU}
\end{array}\right] \times 1$
\end{tabular}
\\ \hline
\begin{tabular}{c}Delay-aware\\ Pos. Encoding\\
\end{tabular} 
&  $M\times 176\times 48\times 256$ & 
\begin{tabular}{c}
$\left[ \begin{array}{c} \text{sin-cos pos. encoding}\end{array}\right]$  \\
$\left[ \begin{array}{c} 
\text{Linear, 256}
\end{array}\right] \times 1$
\end{tabular}
\\ \hline
\begin{tabular}{c}Transformer\\ Backbone\\
\end{tabular} & $M\times 176\times 48\times 256$ & 
$\left[ \begin{array}{c} \text{HSMA, dim 256, head 8} \\ 
\text{MSwin, dim 256, } \\ 
\text{head \{16, 8, 4\},} \\
\text{ws. }\{4\times 4,8\times 8,16\times 16\} \\
\text{MLP, dim 256}
\end{array}\right] \times 3$
\\ \hline

\begin{tabular}{c}Detection\\ Head\\
\end{tabular} & $176\times 48\times 16$ &
\begin{tabular}{c}
Cls. head: $\left[ \begin{array}{c}
\text{Conv1x1, 2, stride 1}
\end{array}\right]$ \\
Regr. head: $\left[ \begin{array}{c}
\text{Conv1x1, 14, stride 1}
\end{array}\right]$
\end{tabular}
\\
\bottomrule
\end{tabular}
\end{table}

\section{V2XSet Dataset}
\label{sec:opv2x-dataset}
\noindent\textbf{Statistics} We gather 55 representative scenes covering 5 different roadway types and 8 towns in CARLA. Each scene is limited to 25 seconds, and in each scene, there are at least 2 and at most 7 intelligent agents that can communicate with each other. Each agent is equipped with 32-channel LiDAR and has 120 meters data range. We mount sensors on top of each AV while we only deploy infrastructure sensors in the intersection, mid-block, and entrance ramp at the height of 14~feet since these scenarios are typically more congested and challenging~\cite{guo2020evaluating}. We record LiDAR point clouds at 10 Hz and save the corresponding positional data and timestamp.

\noindent\textbf{Infrastructure deployment. }The infrastructure sensors
are installed on the traffic light poles or steet light poles at the intersection, mid-block, and entrance ramp at the height of 14 feet. For road type like rural curvy road, there is no infrastructure installed and only V2V collaboration exists.

\begin{figure}[!ht]
\centering
\includegraphics[width=0.7\textwidth]{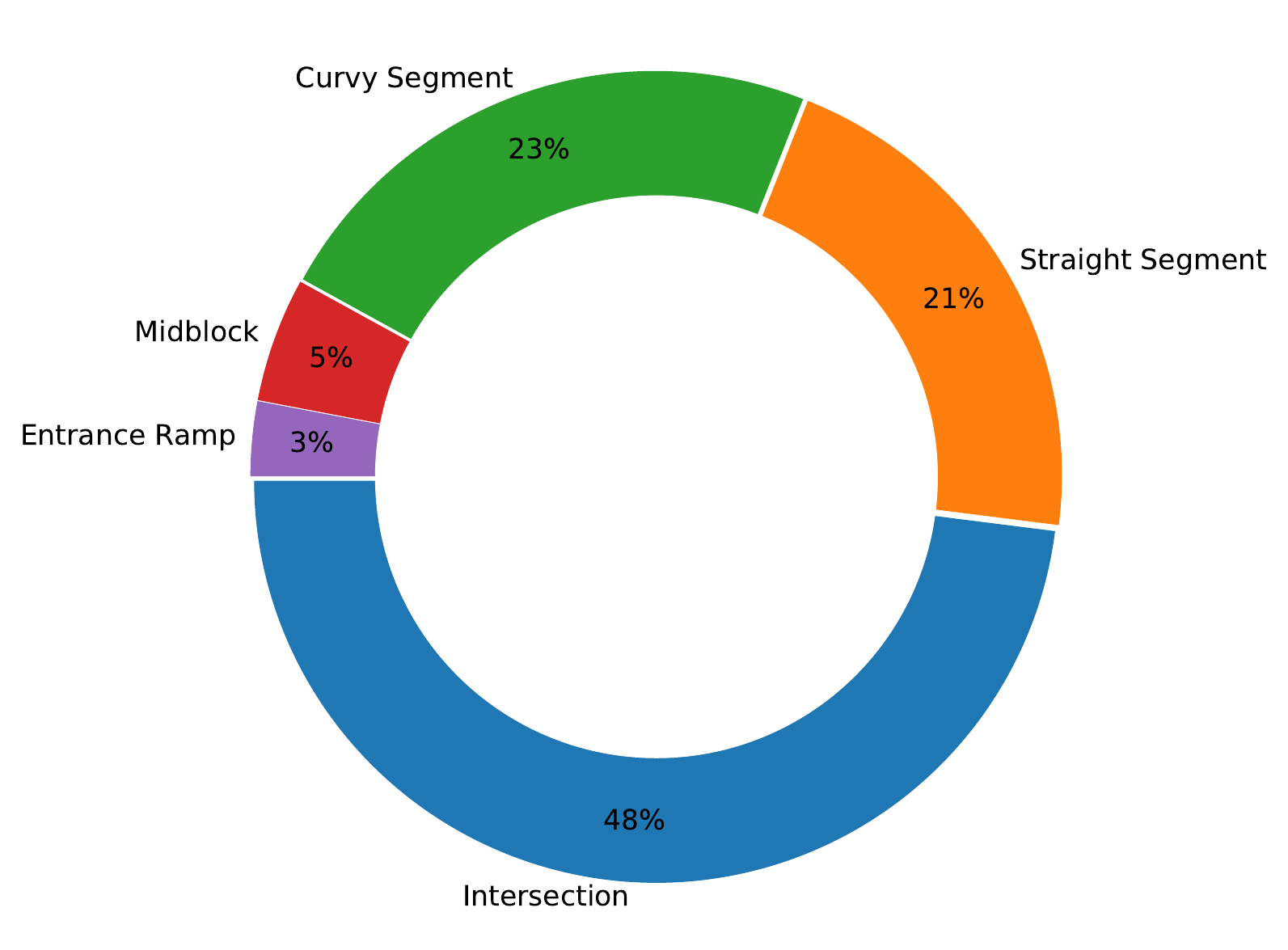}
\caption{Data distribution of 5 roadway types in the proposed dataset.}
\label{fig:chart}
\end{figure}

\begin{figure*}[!ht]
\centering
\footnotesize
\def\xwidth{0.88}
\begin{tabular}{c}
\includegraphics[width=\xwidth\linewidth]{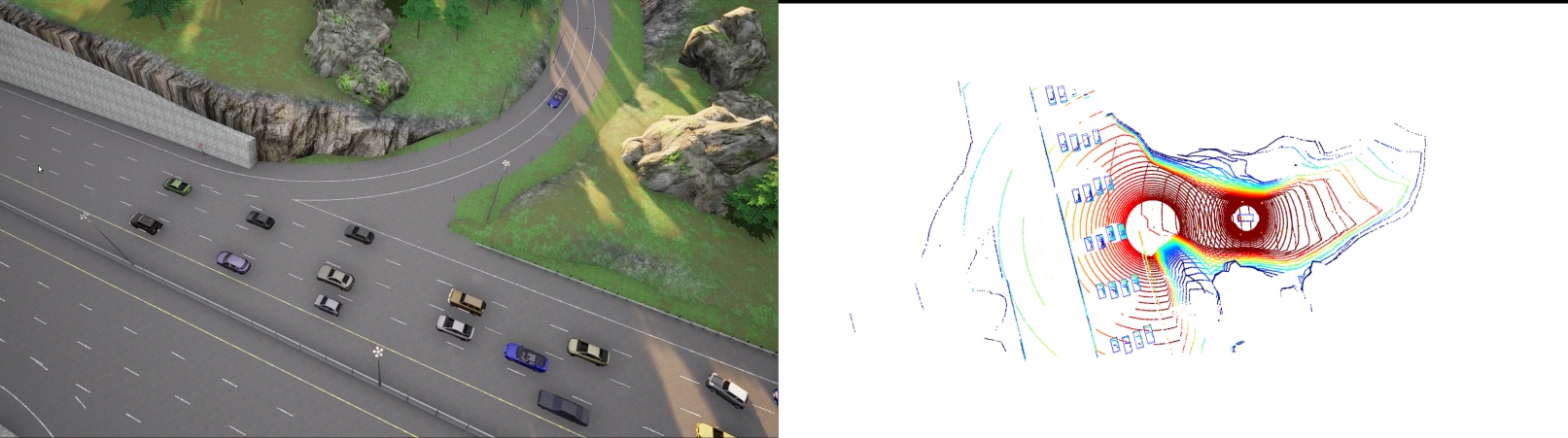} \\
(a) Entrance ramp \\
\includegraphics[width=\xwidth\linewidth]{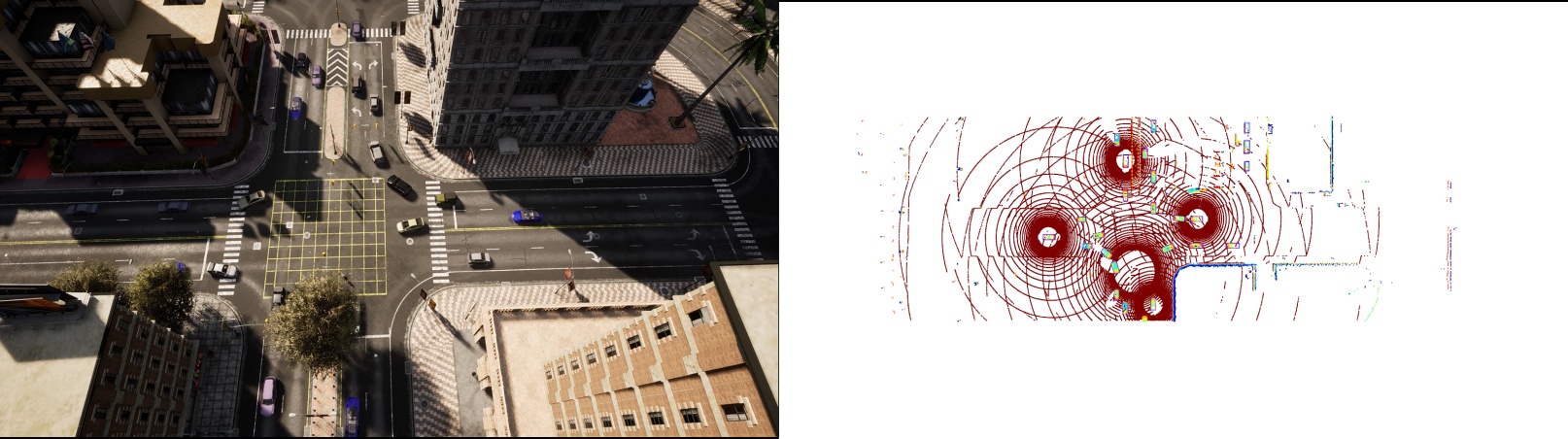} \\
(b) Intersaction \\
\includegraphics[width=\xwidth\linewidth]{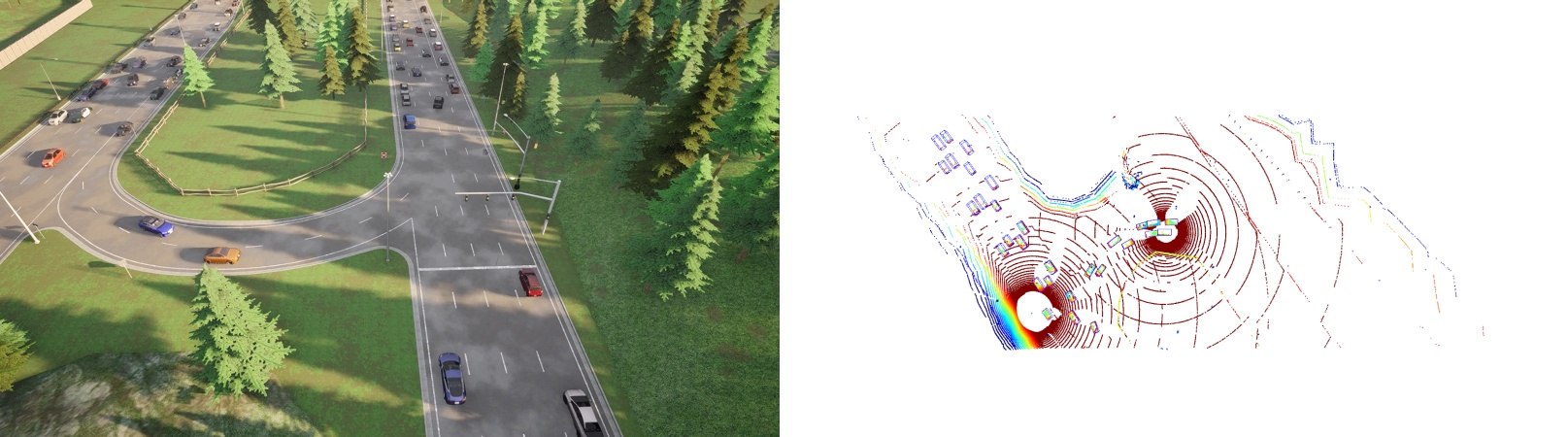} \\
(c) Mid-block \\
\includegraphics[width=\xwidth\linewidth]{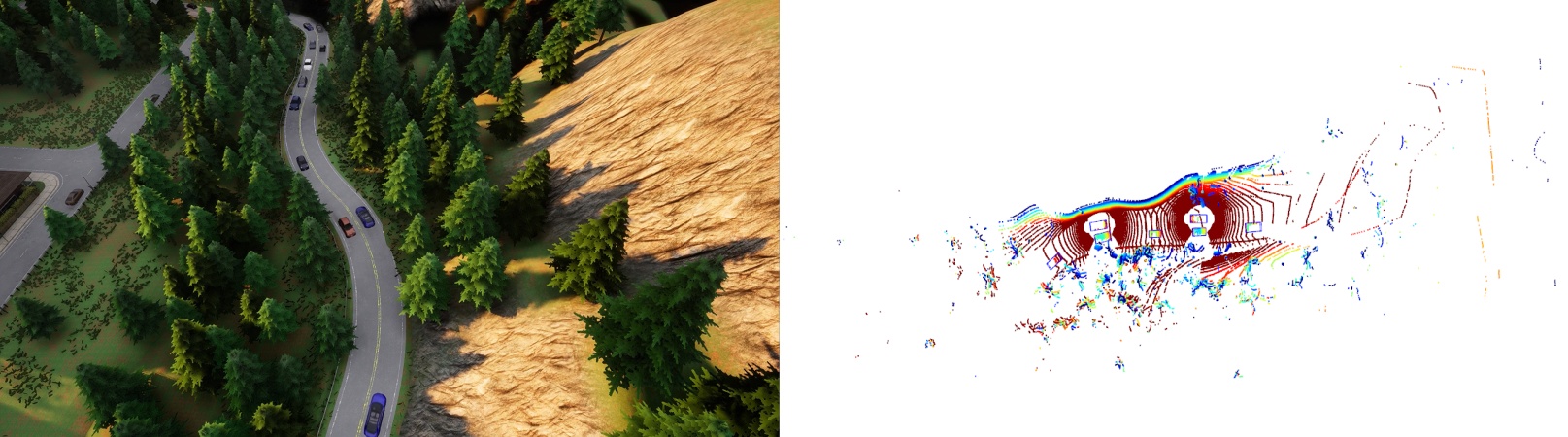} \\
(d) Rural curvy road \\
\includegraphics[width=\xwidth\linewidth]{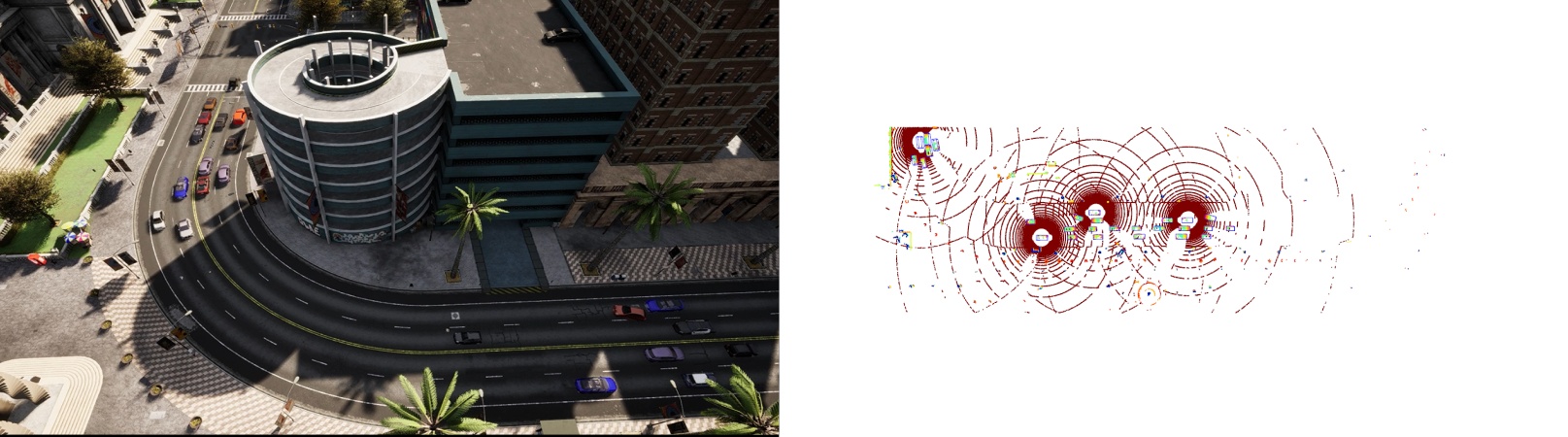} \\
(e) Urban curvy street\\
\end{tabular}
\caption{\textbf{Data samples of 5 different roadway types.} Left is the snapshot of simulation and right is the corresponding aggregated LiDAR point clouds from multiple agents.}
\label{fig:sample}
\end{figure*}

\noindent\textbf{Dataset visualization. }
As~\cref{fig:chart} displays, there are 5 different roadway types in V2XSet dataset (\ie, straight segment, curvy segment, midblock, entrance ramp, and intersection),
covering the most common driving scenarios in real life. We collect more intersection scenes than other types as it is usually more challenging due to the high traffic volume and severe occlusions. Data samples from different roadway types can be found in \cref{fig:sample}. From the figure, we can observe that the infrastructure sensors at the entrance ramp and intersection have different measurement patterns especially near its installation position compared with vehicle sensors. This is caused by the different installation heights between vehicle and infrastructure sensors. Such observation again shows the necessity of capturing the heterogeneity nature of V2X system. 

\section{More Experimental Results}
\label{sec:additional-results}

\subsection{Performance for identifying dynamic objects}
We group the test set based on object speeds v~(km/h) and compare AP@IoU=0.7 under the noisy setting for all intermediate fusion models. As shown in~\cref{tbl:dynamic-objects}, V2X-ViT outperforms all other intermediate fusion methods under various speed range. It is noticeable that the objects with higher speed range generally have lower AP scores as the same time delay can produce more positional mis-alignments for the high-speed vehicles. 

\begin{table}[!htb]
\centering
\footnotesize
\setlength{\tabcolsep}{10pt}
\renewcommand\thetable{T3}
\caption{Perception performance for objects with different speed (km/h), measured in AP@0.7 under noisy setting.}
\label{tbl:dynamic-objects}
\begin{tabular}{l|c|c|c}
 \cellcolor{lightgray} Model &   \cellcolor{lightgray}  $v<20$ &  \cellcolor{lightgray}  $20\le v\le 40$&  \cellcolor{lightgray}  $v>40$\\ \toprule 
 F-Cooper & 0.539 & 0.487 & 0.354 \\
 OPV2V  & 0.552 & 0.498 & 0.346\\
 V2VNet  & 0.598 & 0.518 & 0.406 \\
DiscoNet  & 0.639 & 0.580 & 0.420 \\
 V2X-ViT   & \textbf{0.693} & \textbf{0.634} &  \textbf{0.488} \\  \bottomrule
\end{tabular}
\end{table}

\subsection{Performance for different road types}
We also group the test scenes based on their road types and calculate the AP@IoU=0.7 scores under the noisy setting. As shown in~\cref{tbl:road-types}, V2X-ViT ranks the first for all 5 road categories, demonstrating its detection robustness on different scenes.

\begin{table}[!htb]
\centering
\footnotesize
\setlength{\tabcolsep}{8pt}
\renewcommand\thetable{T4}
\caption{Perception performance for different road types, measured in AP@0.7 under noisy setting.}
\label{tbl:road-types}
\begin{tabular}{l|c|c|c|c|c}
  \cellcolor{lightgray}  Model &  \cellcolor{lightgray}  Straight &  \cellcolor{lightgray}  Curvy &  \cellcolor{lightgray}  Intersection &  \cellcolor{lightgray}  Midblock &  \cellcolor{lightgray}  Entrance    \\ \toprule 
 F-Cooper  & 0.483 &	0.558 &	0.458 &	0.431 &	0.375 \\
 OPV2V   & 0.478	 & 0.604	 & 0.492	 & 0.460  &	0.380 \\
 V2VNet  & 0.496 &	0.556 &	0.517 &	0.489 &	0.360\\ 
DiscoNet  & 0.519 &	0.594 &	0.572 &	0.472 &	0.440\\
 V2X-ViT   & \textbf{0.645}&	\textbf{0.686} &	\textbf{0.615} &	\textbf{0.530} &	\textbf{0.487}  \\  \bottomrule
\end{tabular}
\end{table}

\subsection{Qualitative results}

\begin{figure*}[!ht]
\centering
\footnotesize
\def\xwidth{0.32}
\def\yheight{0.18}
\def\xem{-2pt}
\def\im_shift{0.01\textwidth}
\setlength{\tabcolsep}{0.5pt}
\begin{tabular}{cccc}
& Scene 1 & Scene 2 & Scene 3\\
\multirow[t]{1}{*}[\im_shift]{\begin{sideways}  F-Cooper~\cite{chen2019f}  \end{sideways}} &
\includegraphics[width=\xwidth\linewidth]{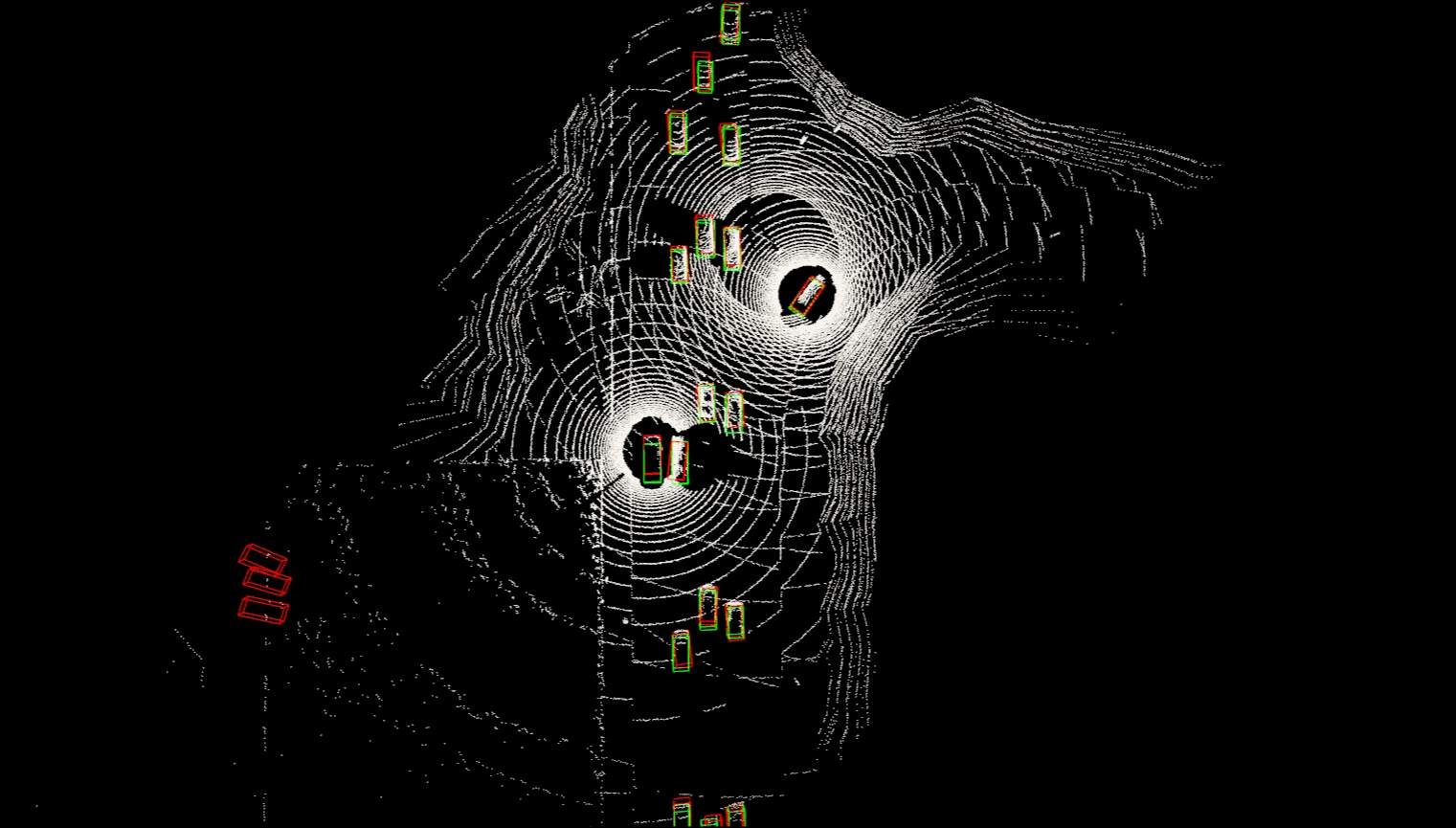}
& \includegraphics[ width=\xwidth\linewidth]{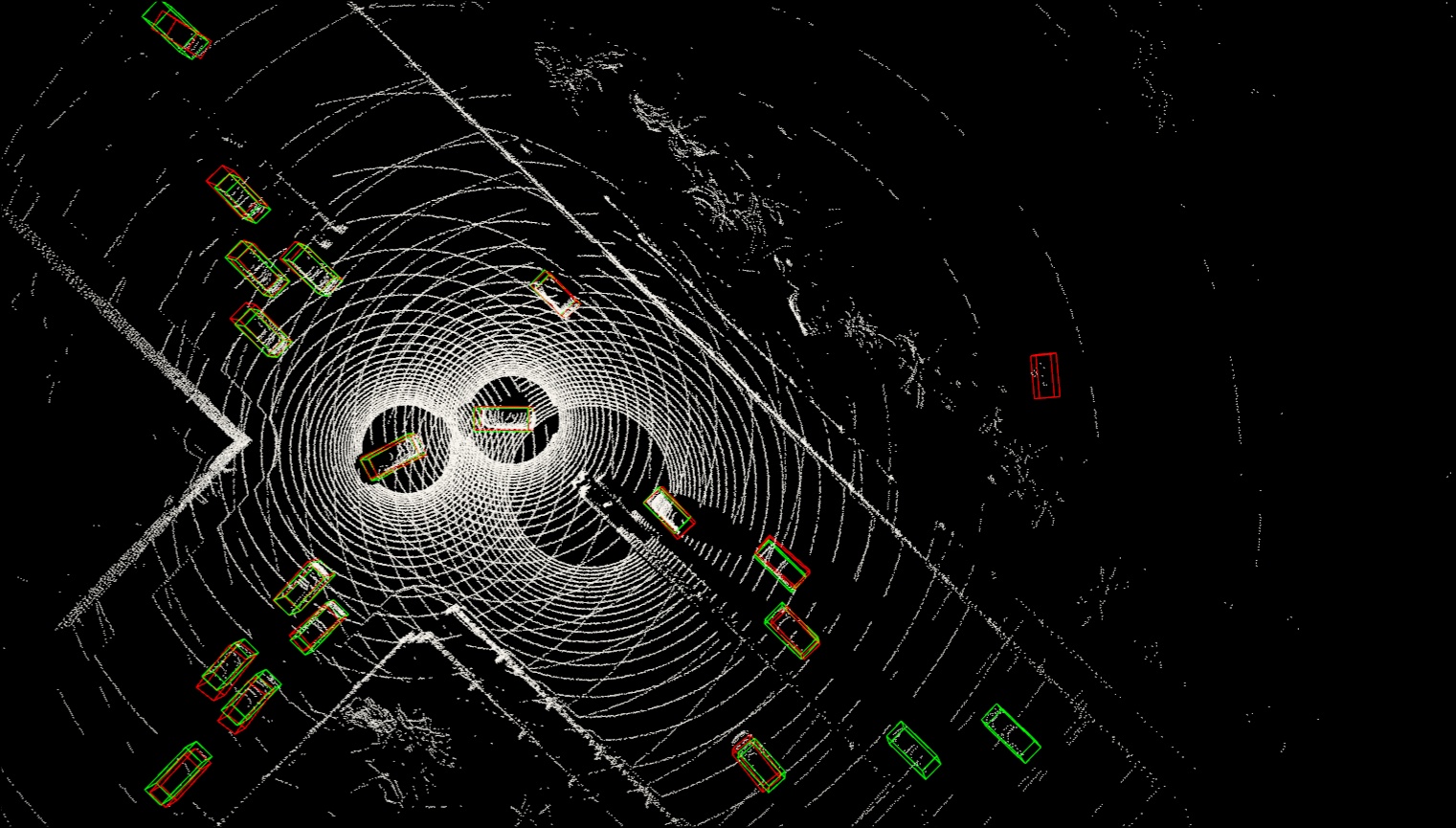}
& \includegraphics[ width=\xwidth\linewidth]{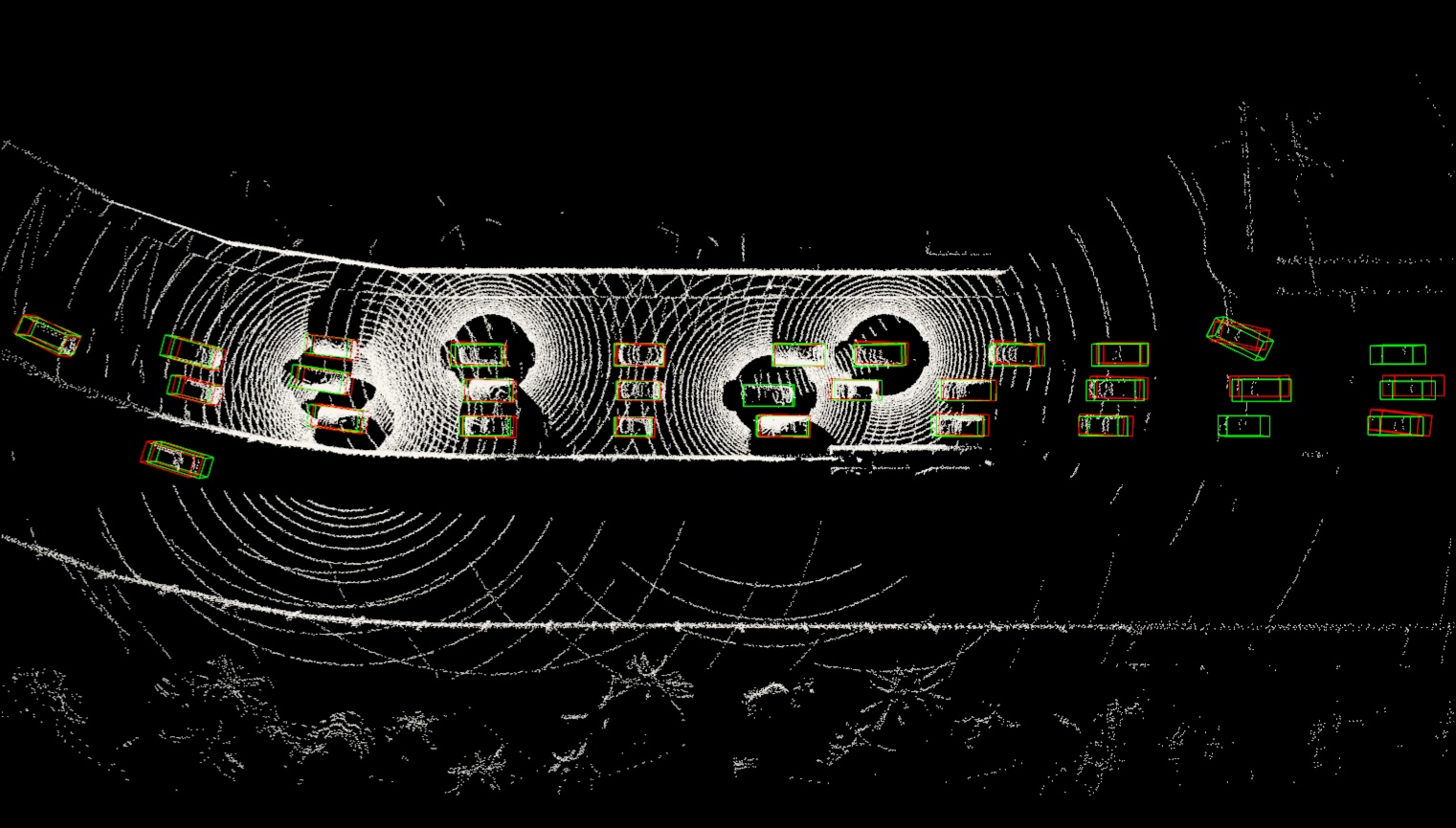} \\
\multirow[t]{1}{*}[\im_shift]{\begin{sideways}  V2VNet~\cite{wang2020v2vnet} \end{sideways}} &
\includegraphics[width=\xwidth\linewidth]{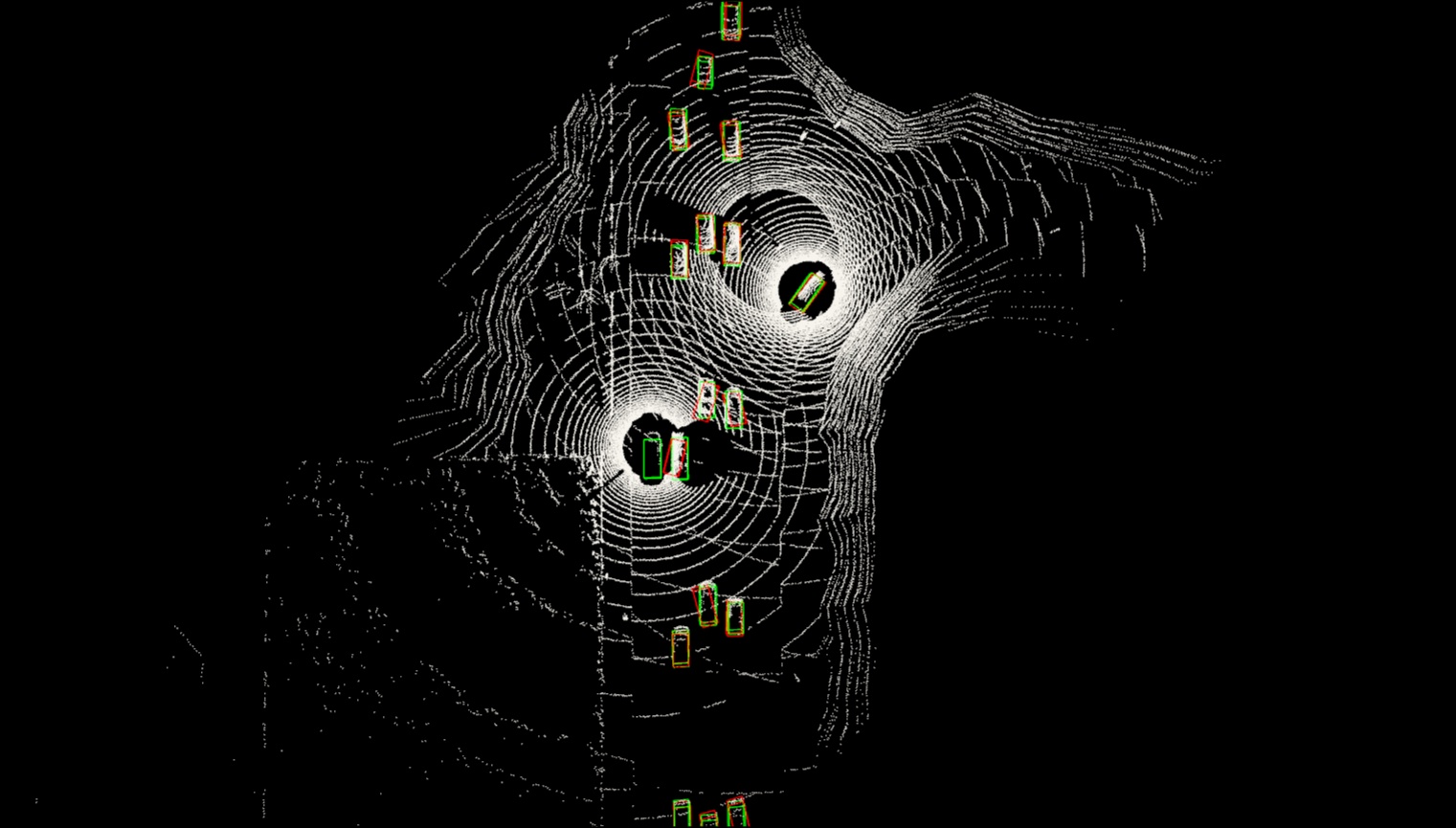}
& \includegraphics[width=\xwidth\linewidth]{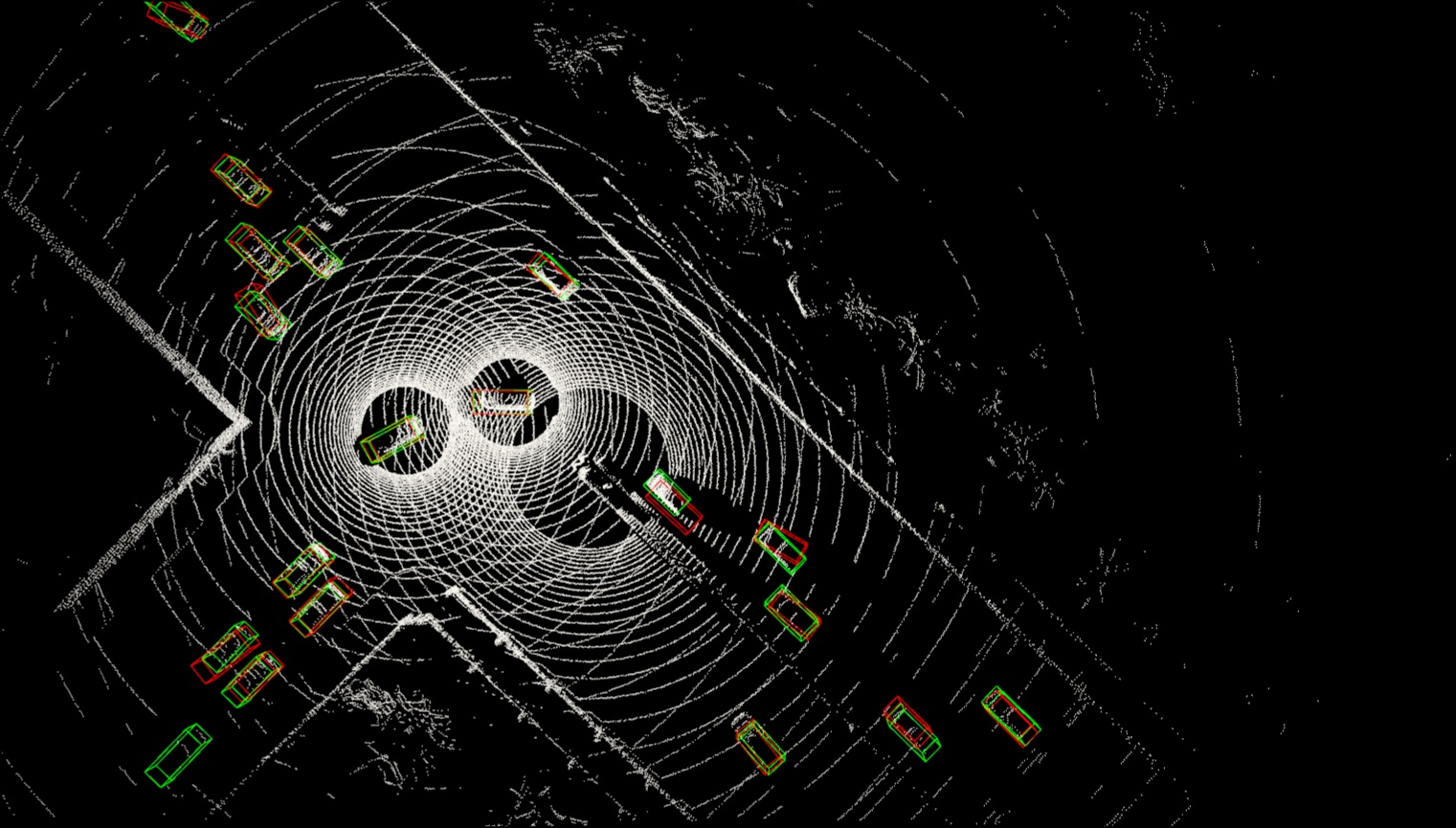} 
& \includegraphics[width=\xwidth\linewidth]{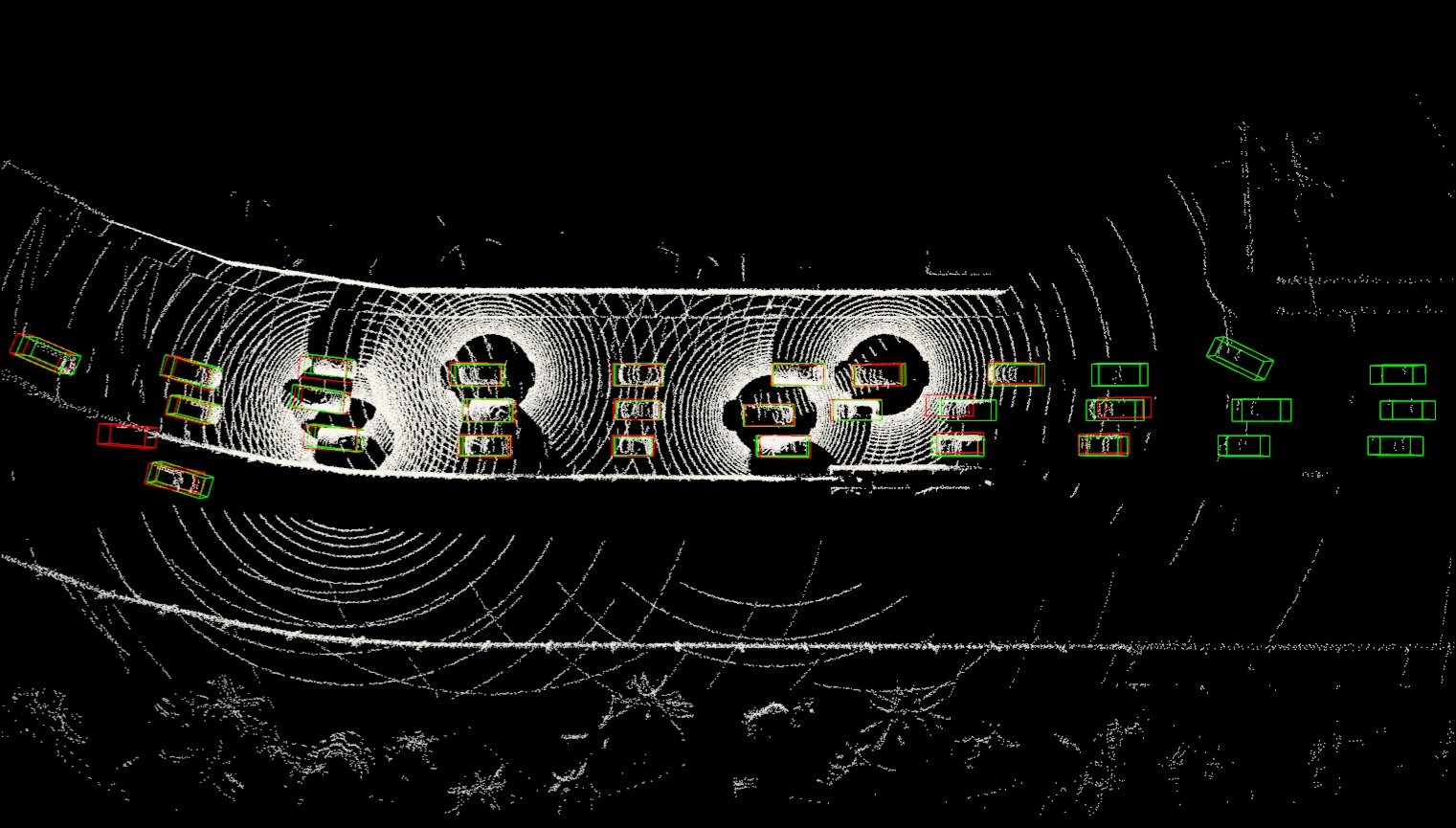}\\
\multirow[t]{1}{*}[\im_shift]{\begin{sideways}  OPV2V~\cite{xu2021opv2v} \end{sideways}} &
\includegraphics[width=\xwidth\linewidth]{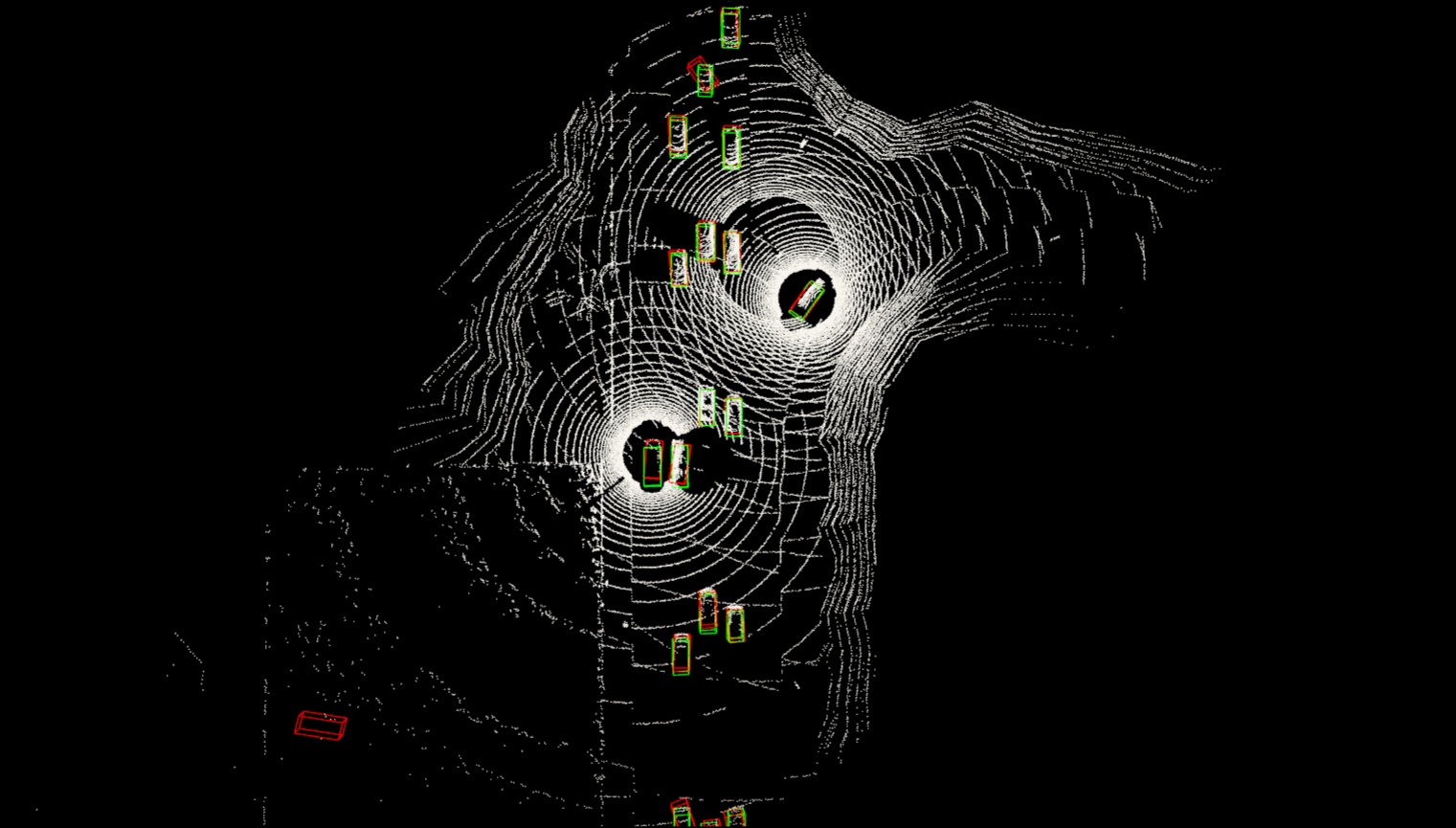}
& \includegraphics[width=\xwidth\linewidth]{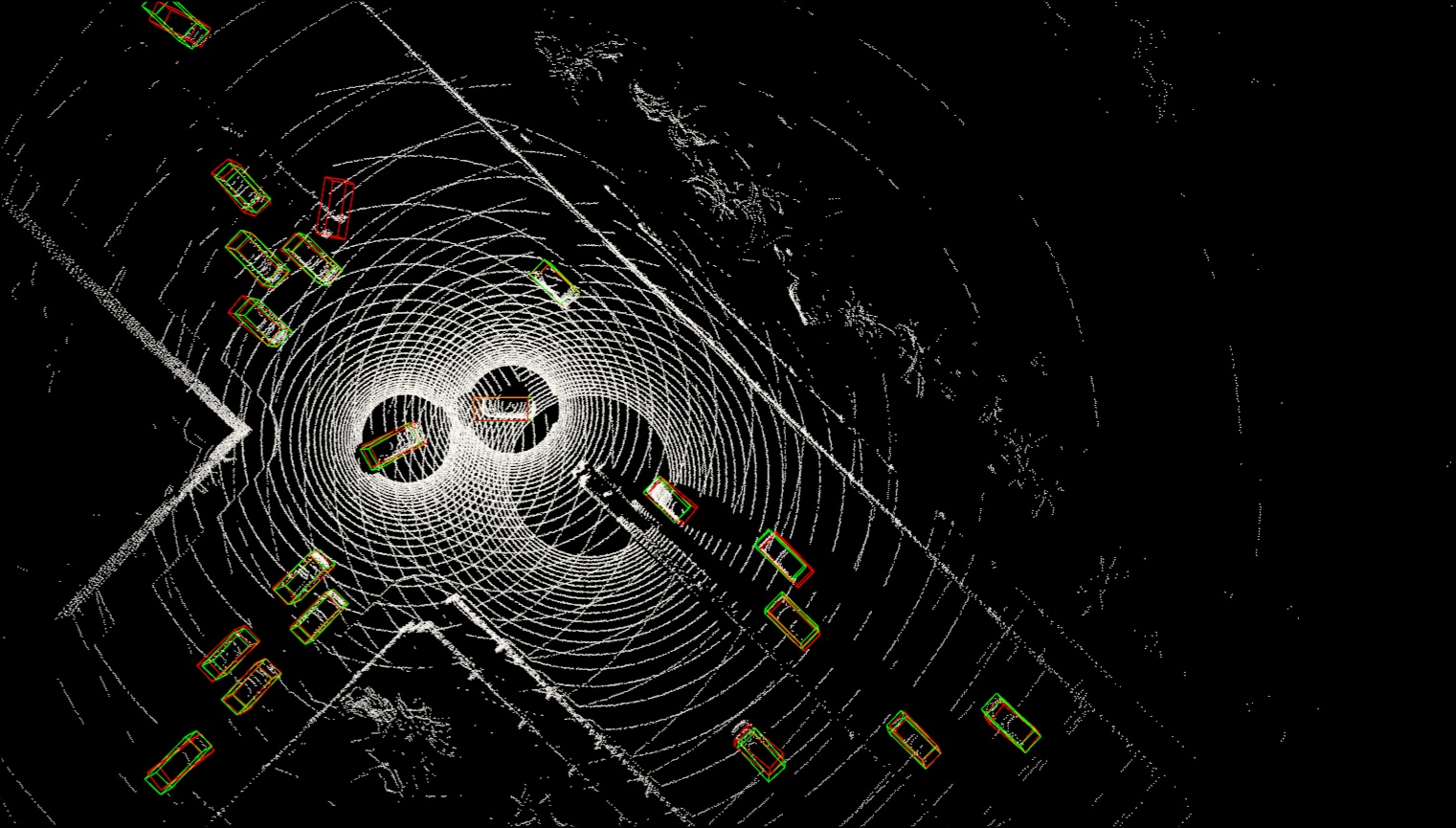}
& \includegraphics[width=\xwidth\linewidth]{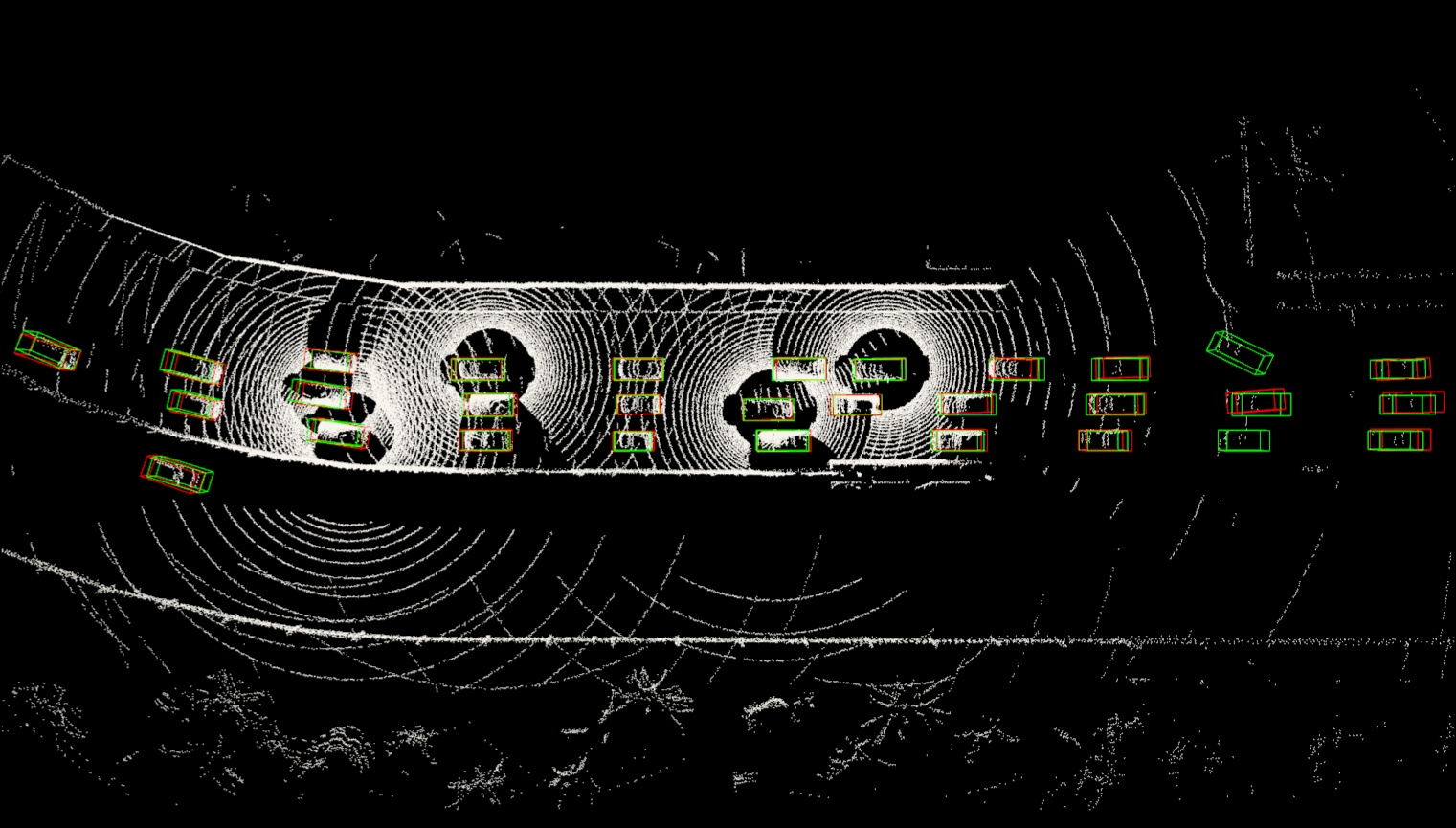} \\
\multirow[t]{1}{*}[\im_shift]{\begin{sideways}  DiscoNet~\cite{li2021learning}  \end{sideways}} 
& \includegraphics[ width=\xwidth\linewidth]{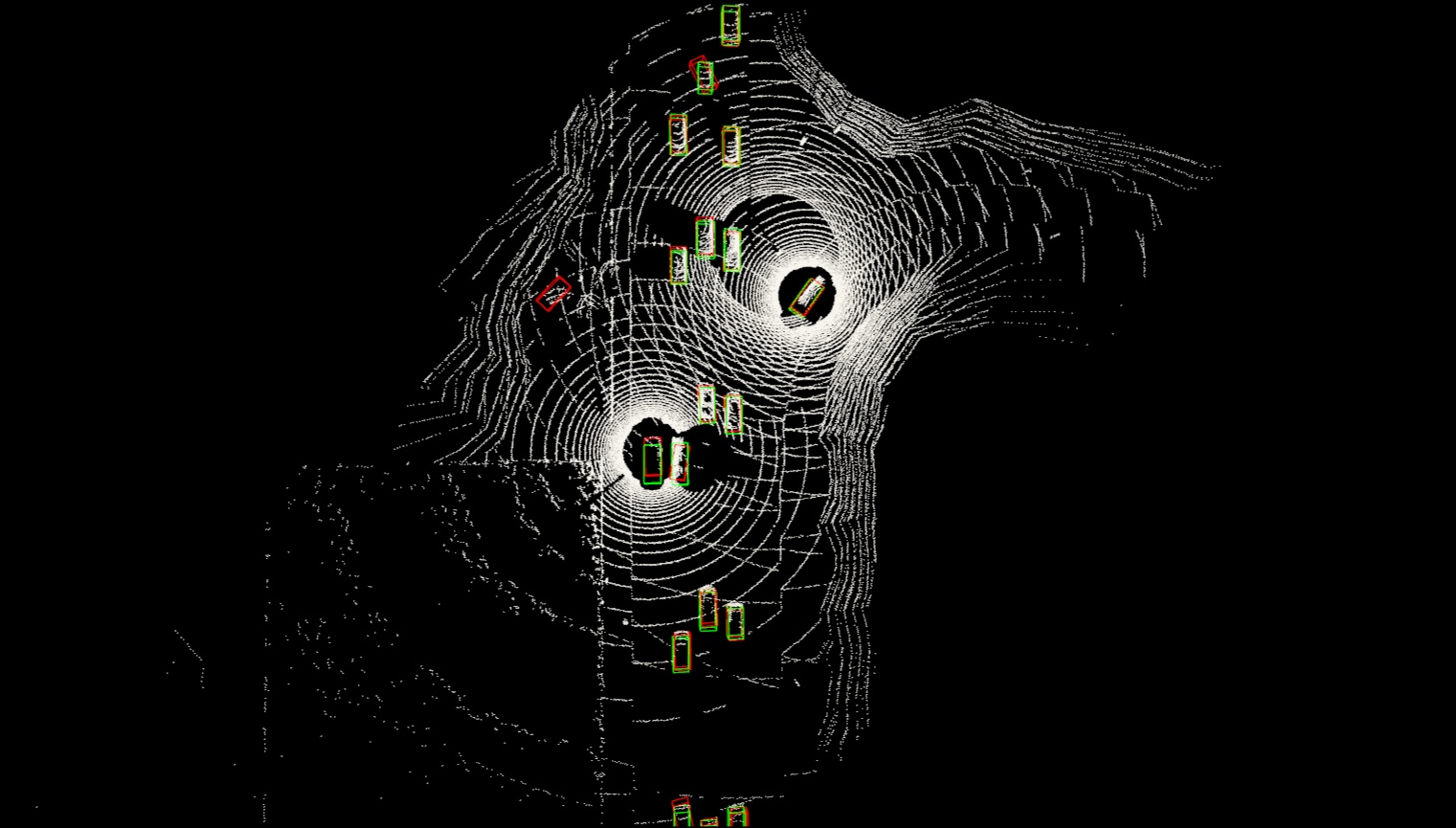}
& \includegraphics[ width=\xwidth\linewidth]{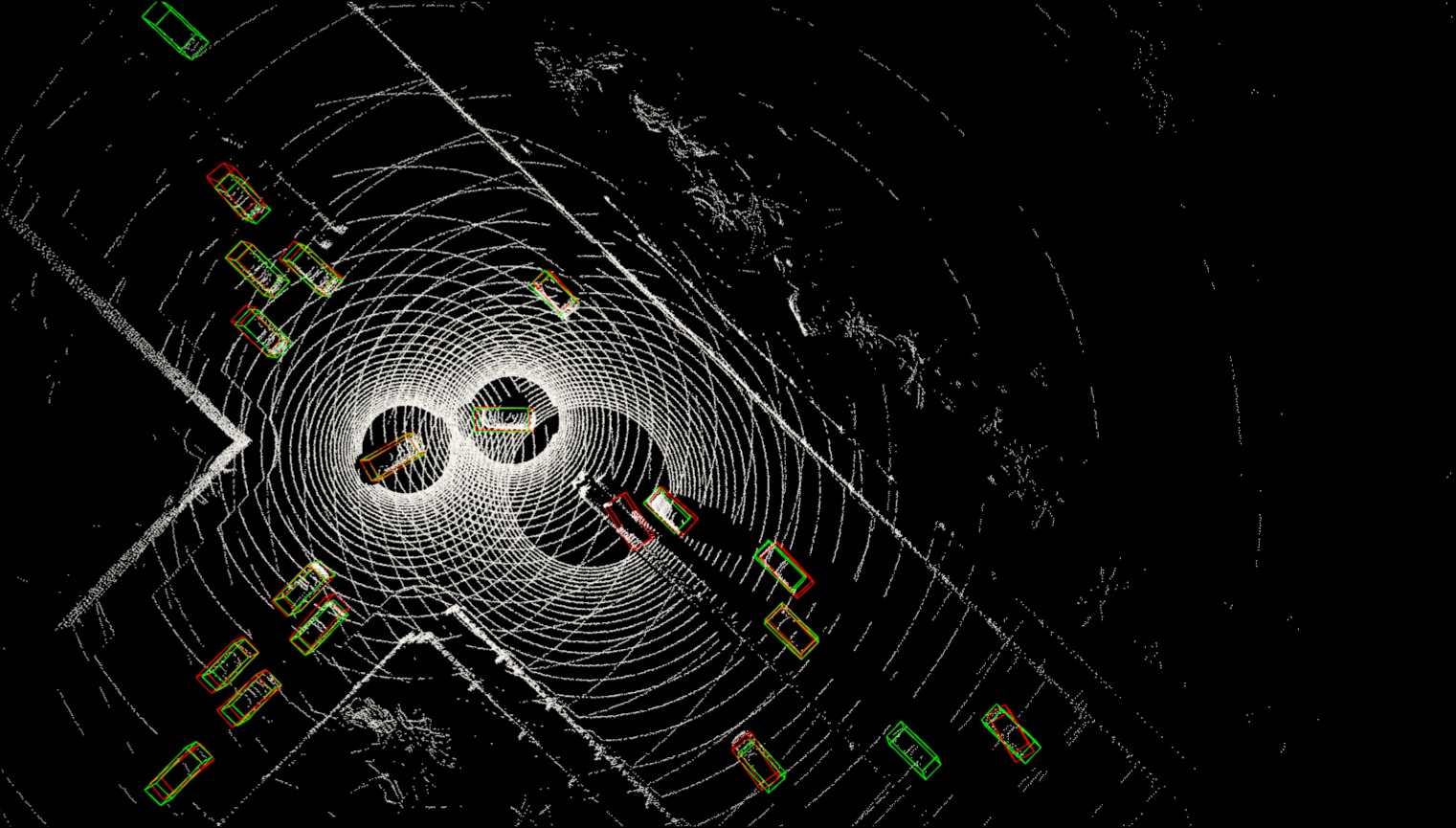}
& \includegraphics[ width=\xwidth\linewidth]{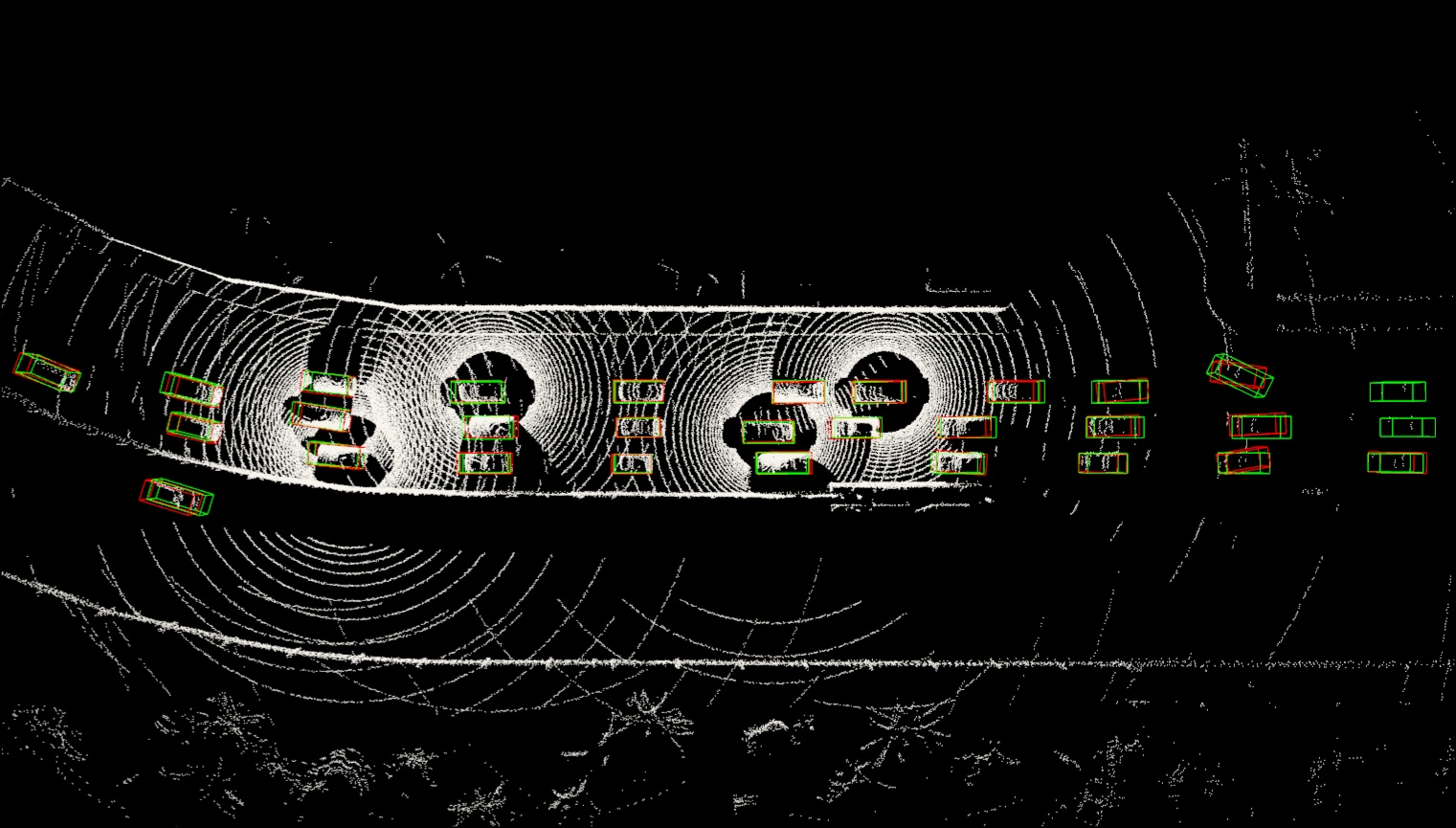} \\
\multirow[t]{1}{*}[0.0\textwidth]{\begin{sideways}  V2X-ViT (ours) \end{sideways}} & \includegraphics[ width=\xwidth\linewidth]{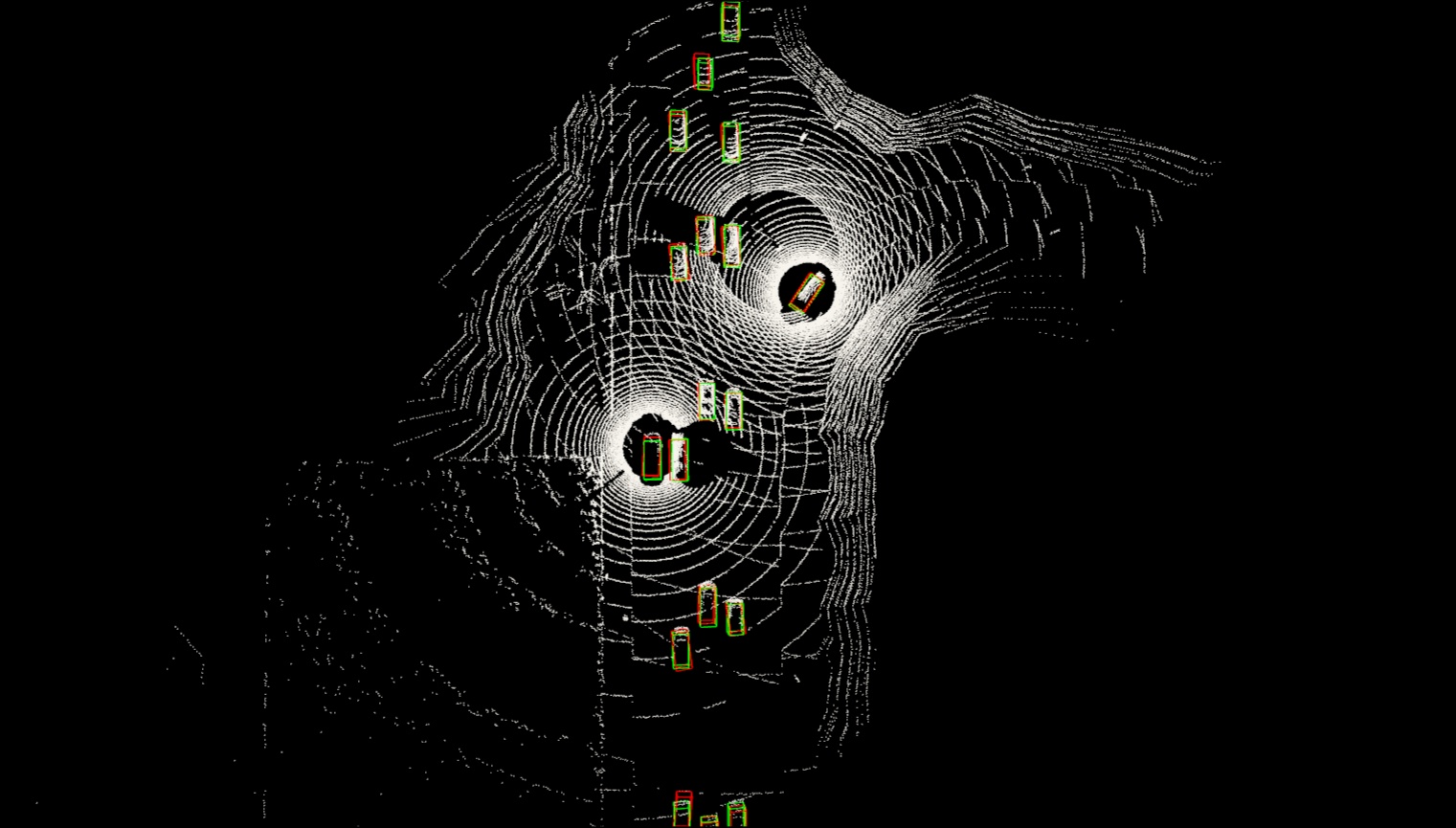}
& \includegraphics[ width=\xwidth\linewidth]{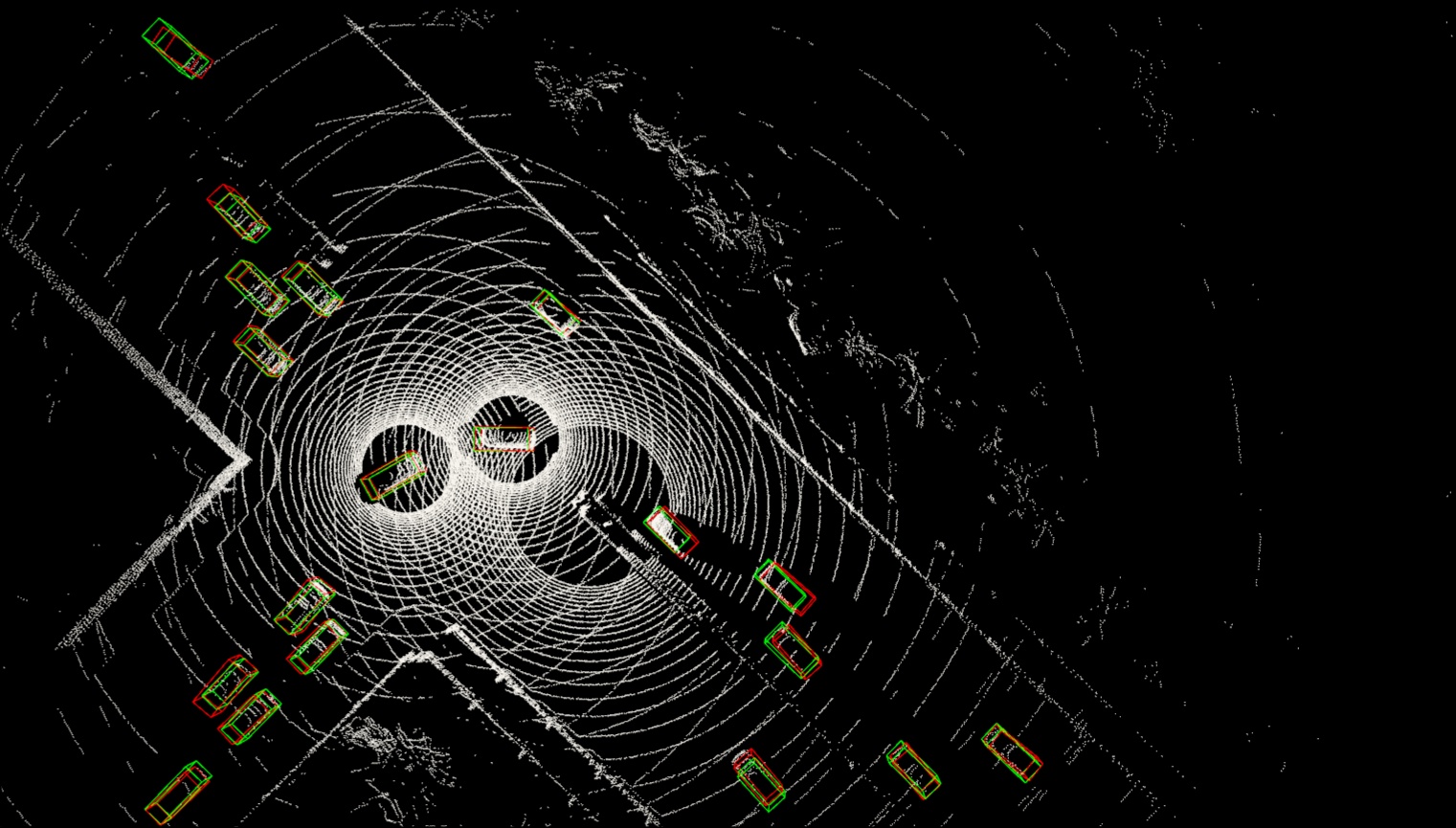}
& \includegraphics[ width=\xwidth\linewidth]{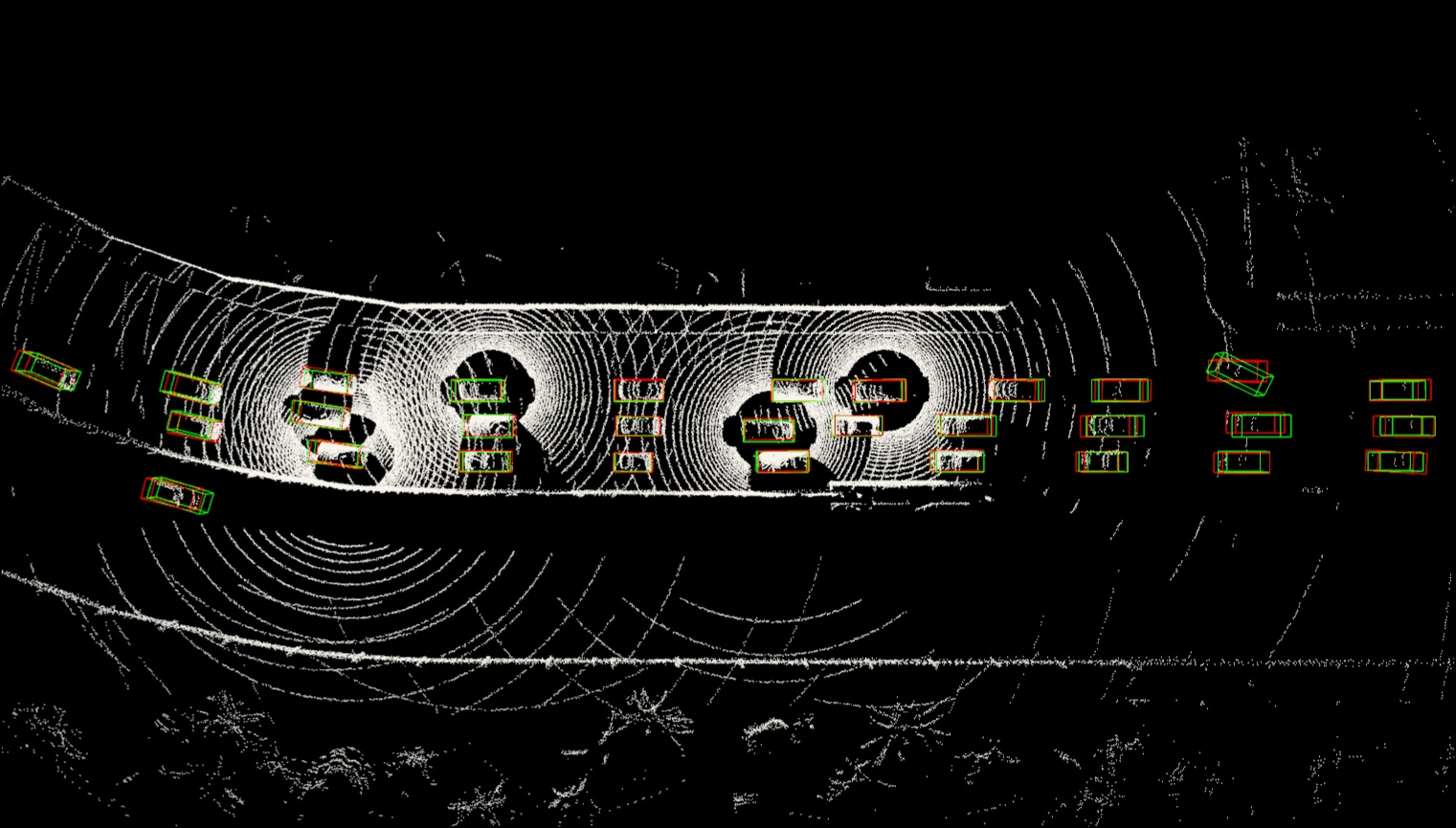} \\
\end{tabular}
\caption{\textbf{Qualitative comparison on scenarios 1-3.} \textcolor{green}{Green} and \textcolor{red}{red} 3D bounding boxes represent the groun truth and prediction respectively. Our method yields more accurate detection results.}
\label{fig:qualitive1}
\end{figure*}

\begin{figure*}[!ht]
\centering
\footnotesize
\def\xwidth{0.32}
\def\yheight{0.18}
\def\xem{-2pt}
\def\im_shift{0.01\textwidth}
\setlength{\tabcolsep}{0.5pt}
\begin{tabular}{cccc}
 & Scene 4 & Scene 5 & Scene 6\\
 \multirow[t]{1}{*}[\im_shift]{\begin{sideways}  F-Cooper~\cite{chen2019f}  \end{sideways}} &
\includegraphics[ width=\xwidth\linewidth]{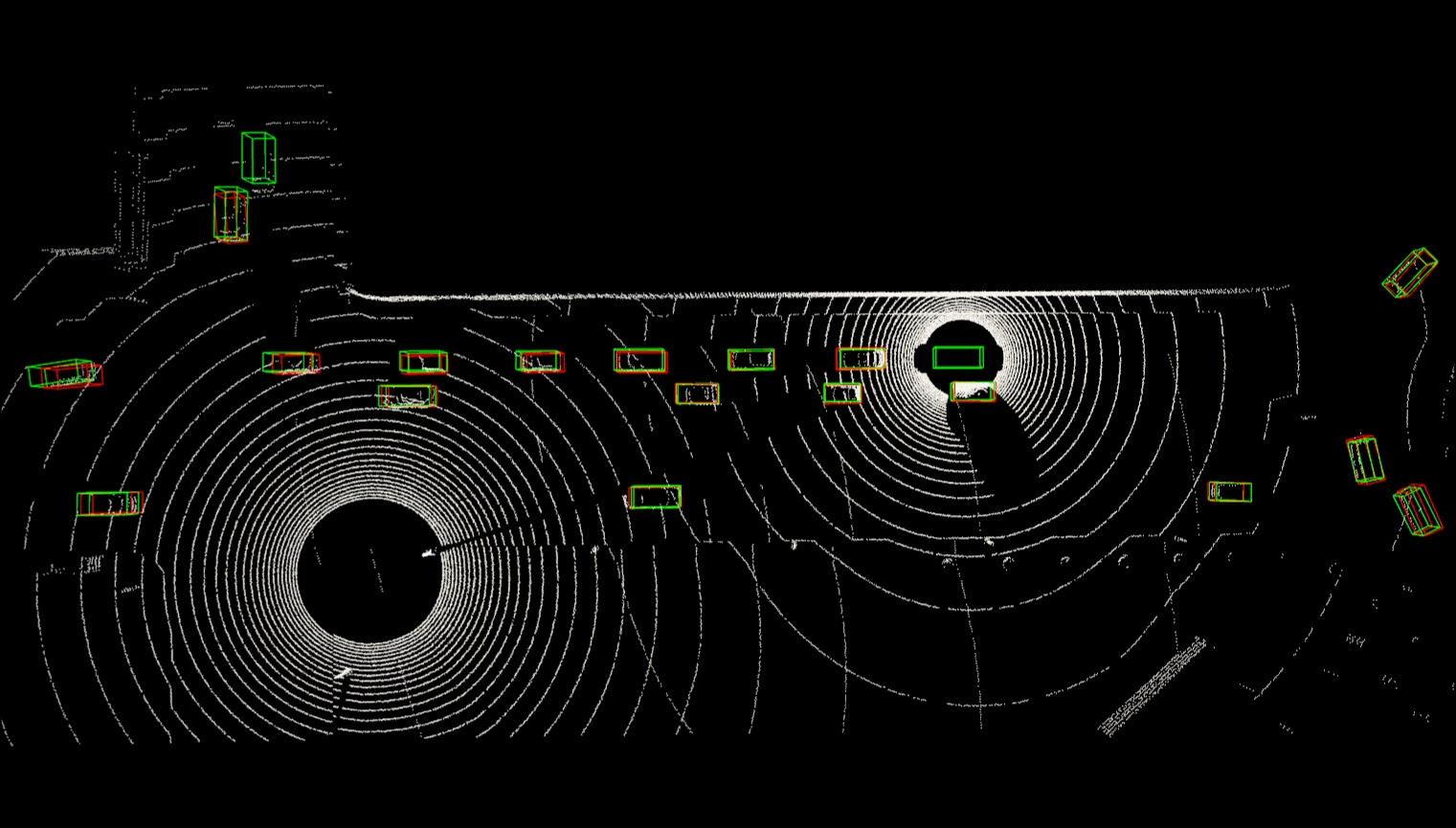}
& \includegraphics[width=\xwidth\linewidth]{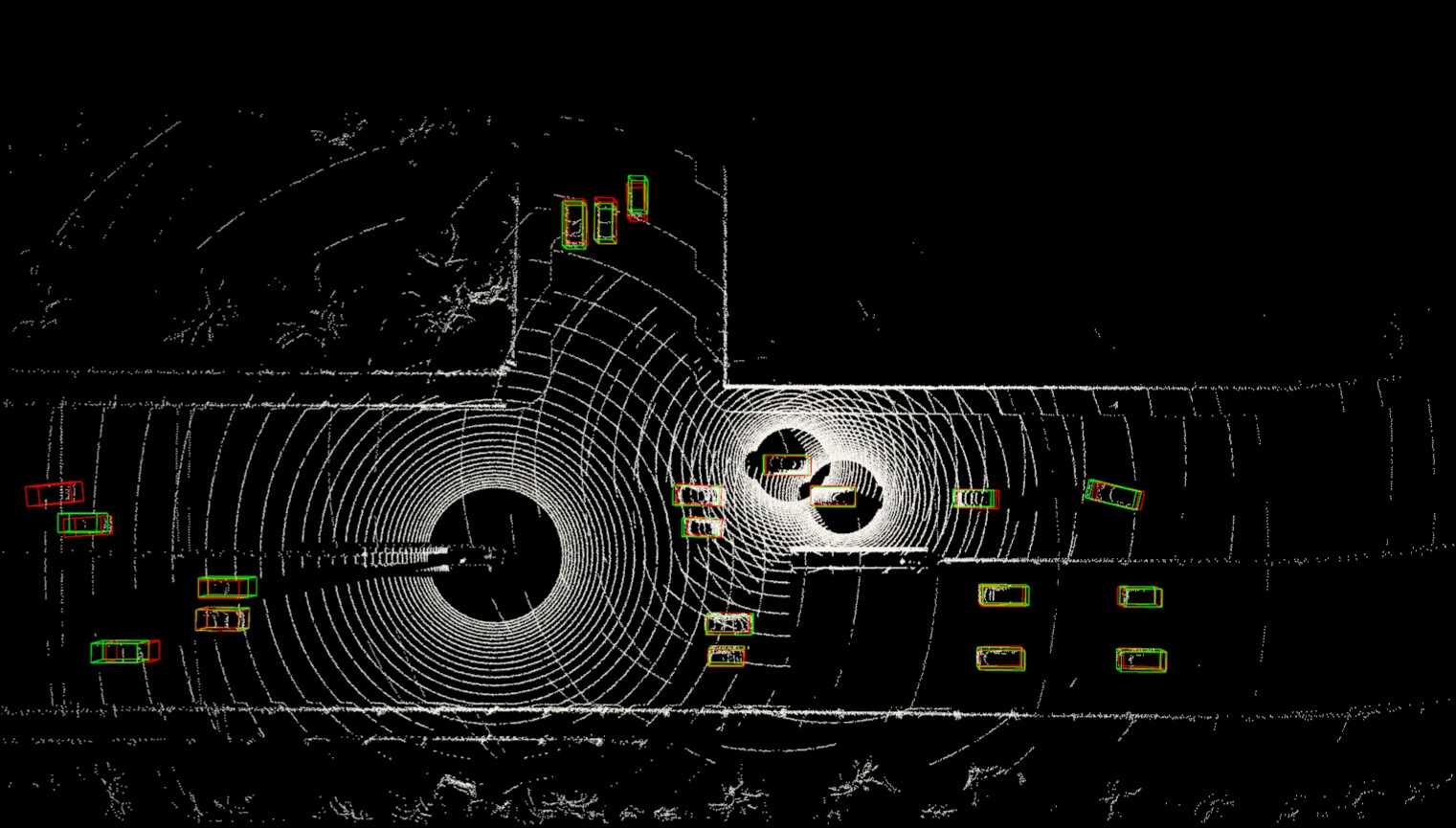}
& \includegraphics[ width=\xwidth\linewidth]{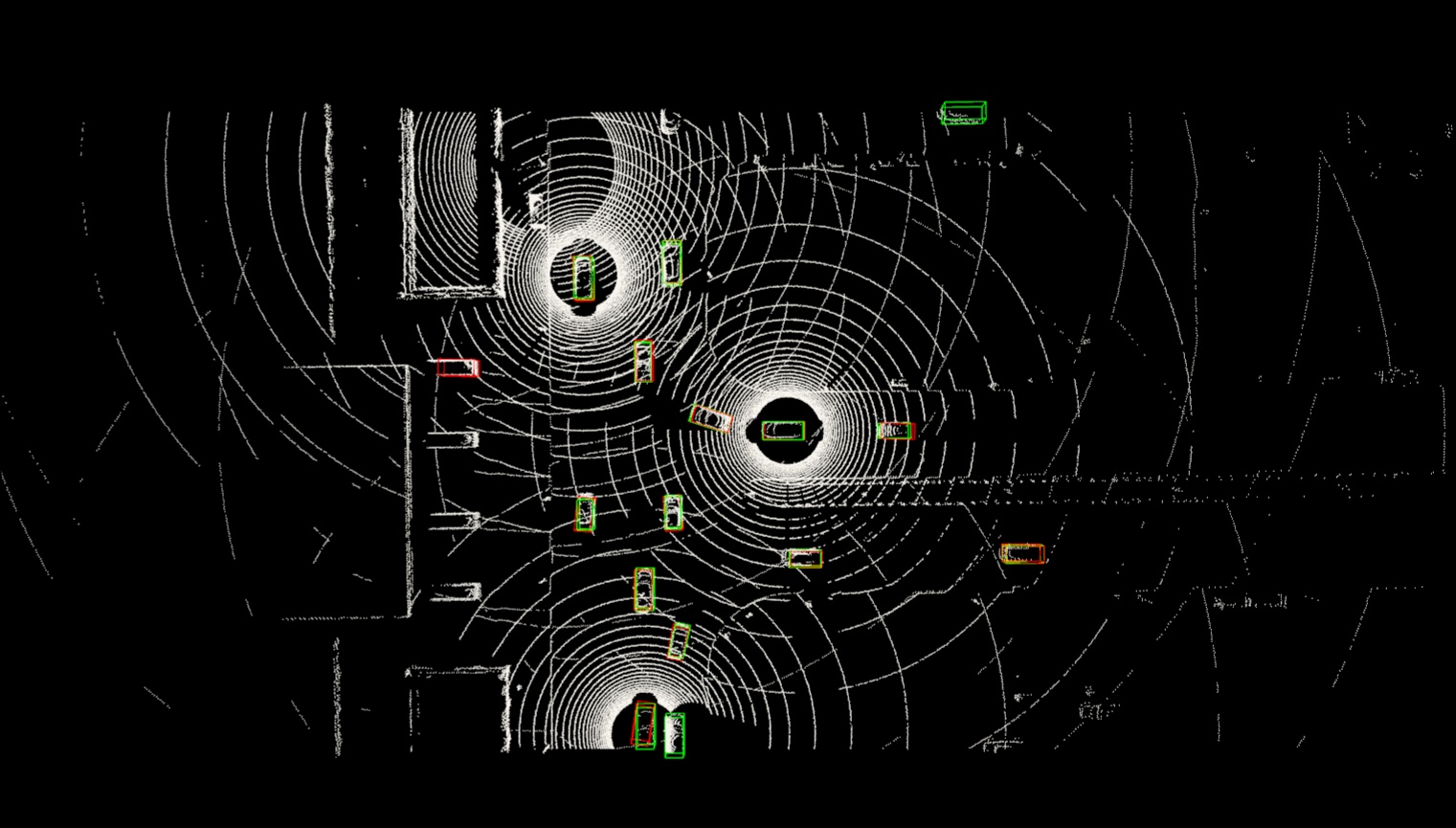}\\
\multirow[t]{1}{*}[\im_shift]{\begin{sideways}  V2VNet~\cite{wang2020v2vnet} \end{sideways}} &
 \includegraphics[width=\xwidth\linewidth]{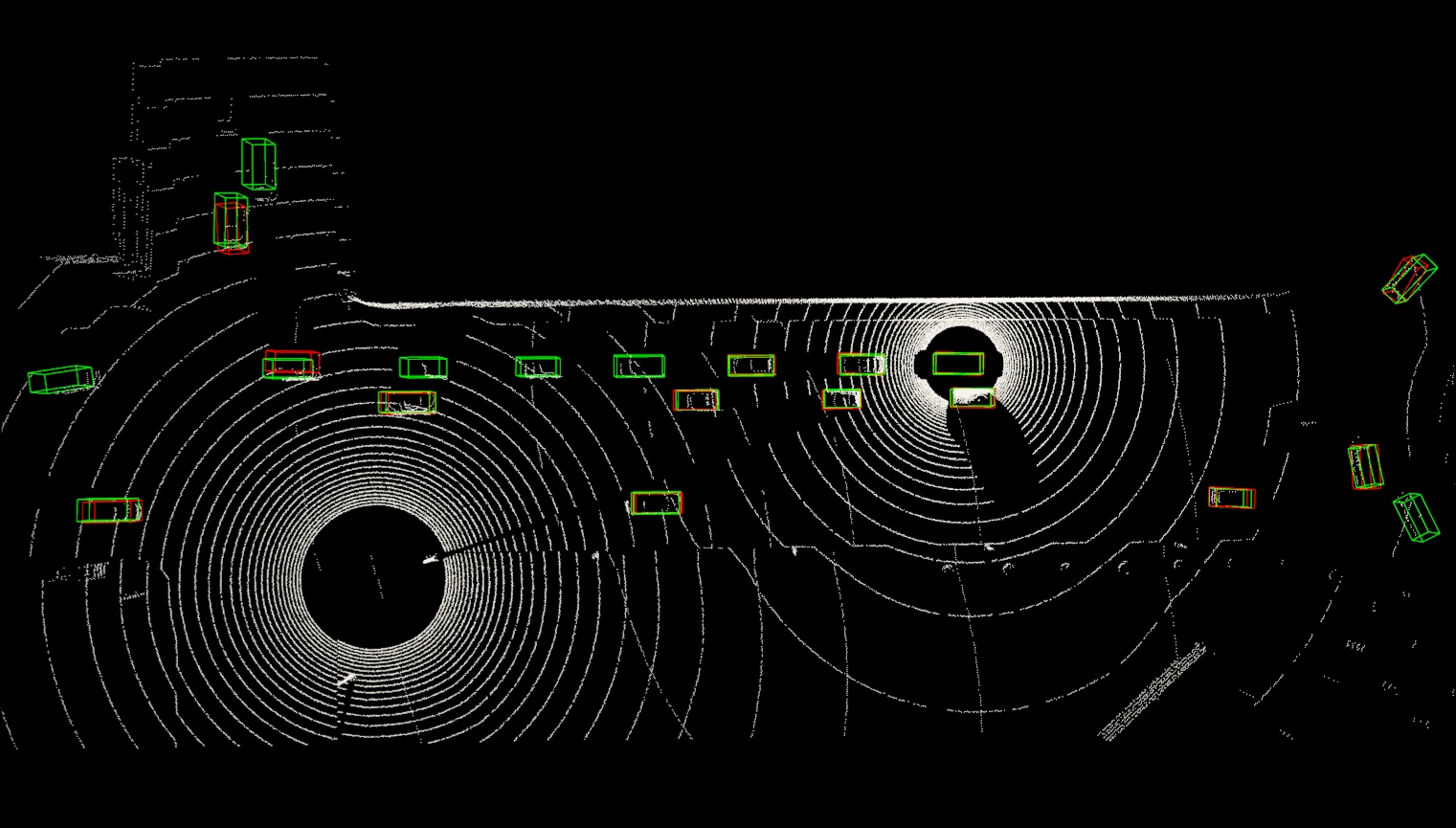}
& \includegraphics[width=\xwidth\linewidth]{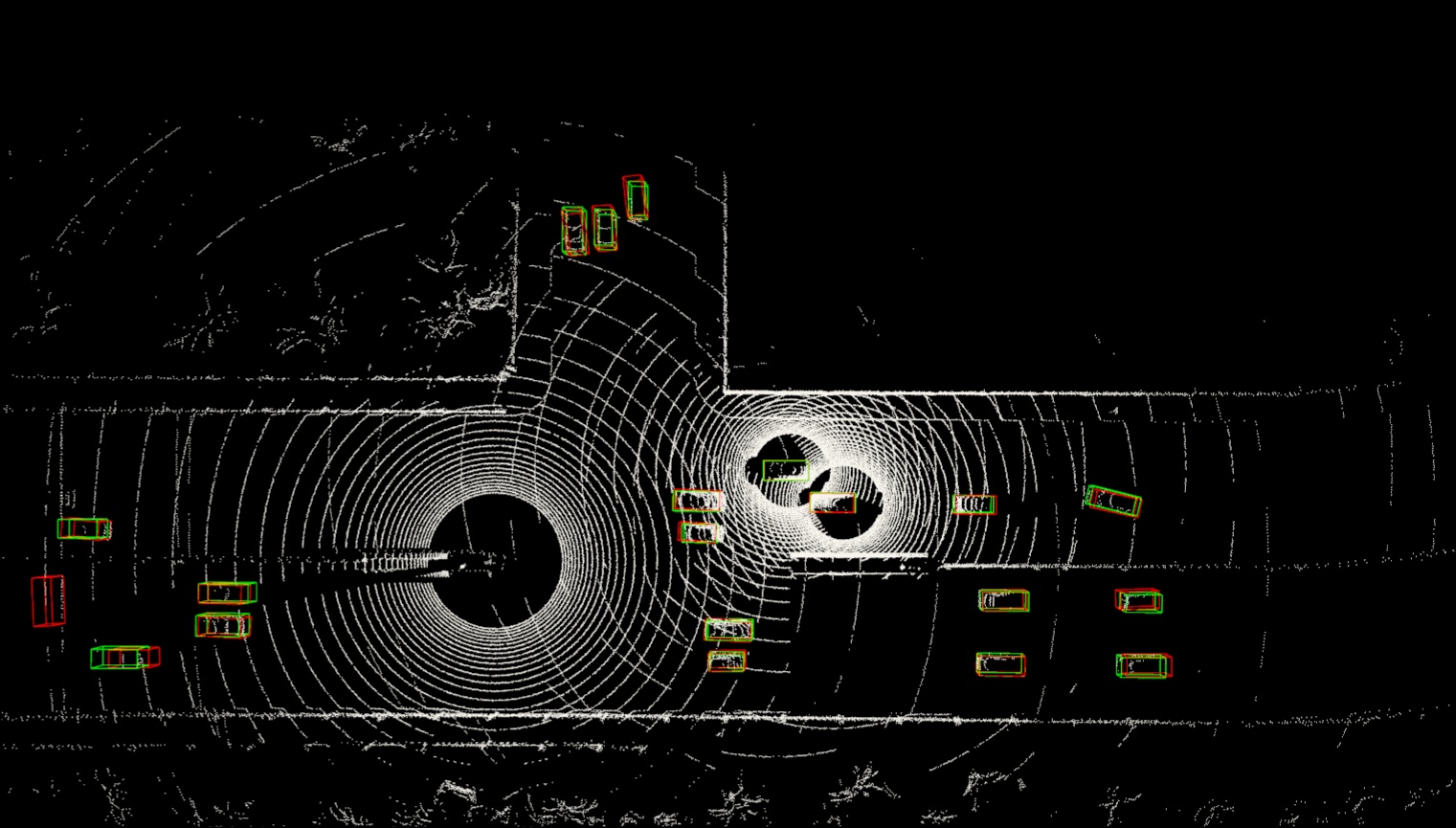}
& \includegraphics[width=\xwidth\linewidth]{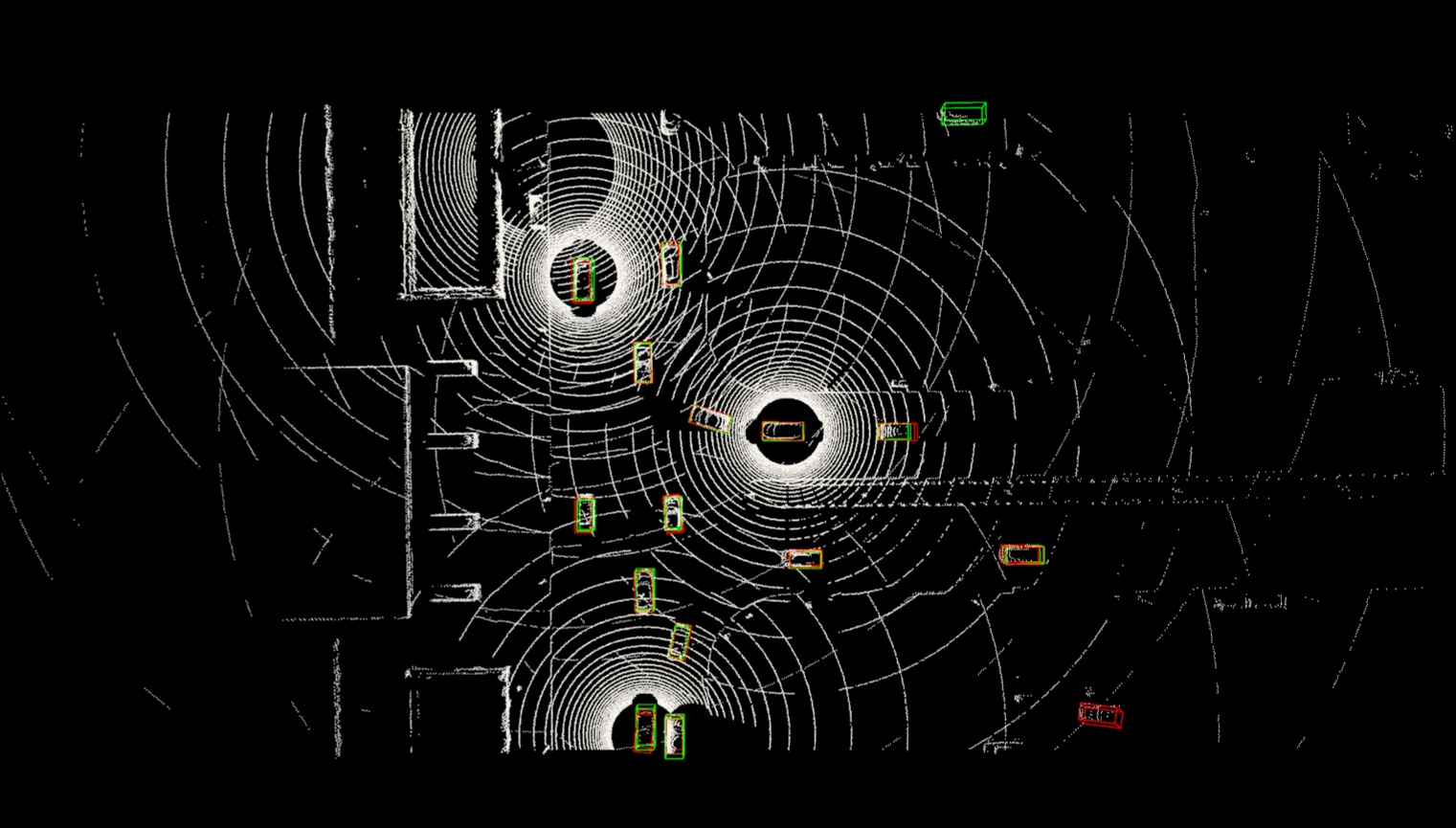} \\
\multirow[t]{1}{*}[\im_shift]{\begin{sideways}  OPV2V~\cite{xu2021opv2v} \end{sideways}} &
 \includegraphics[width=\xwidth\linewidth]{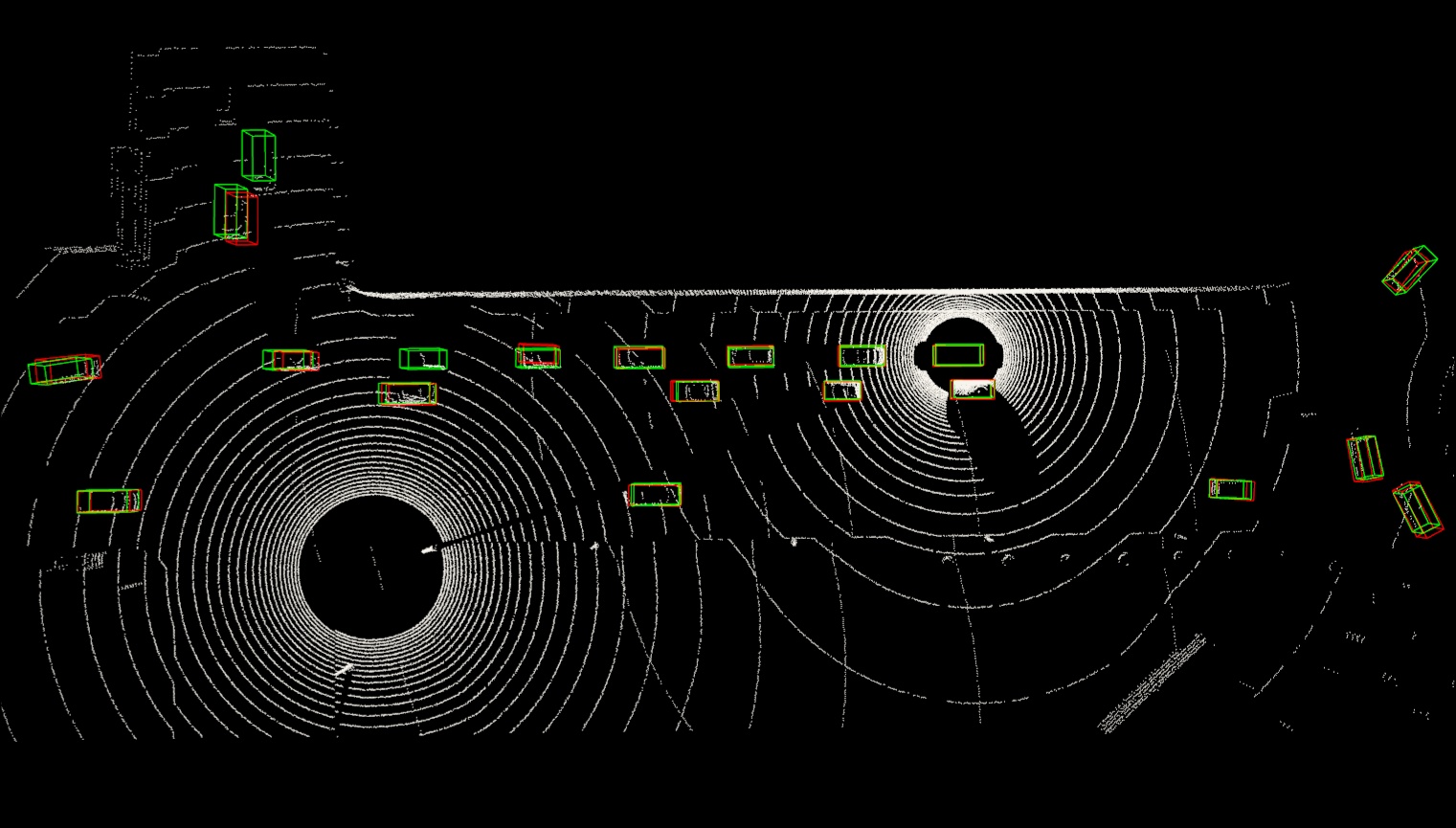}
& \includegraphics[width=\xwidth\linewidth]{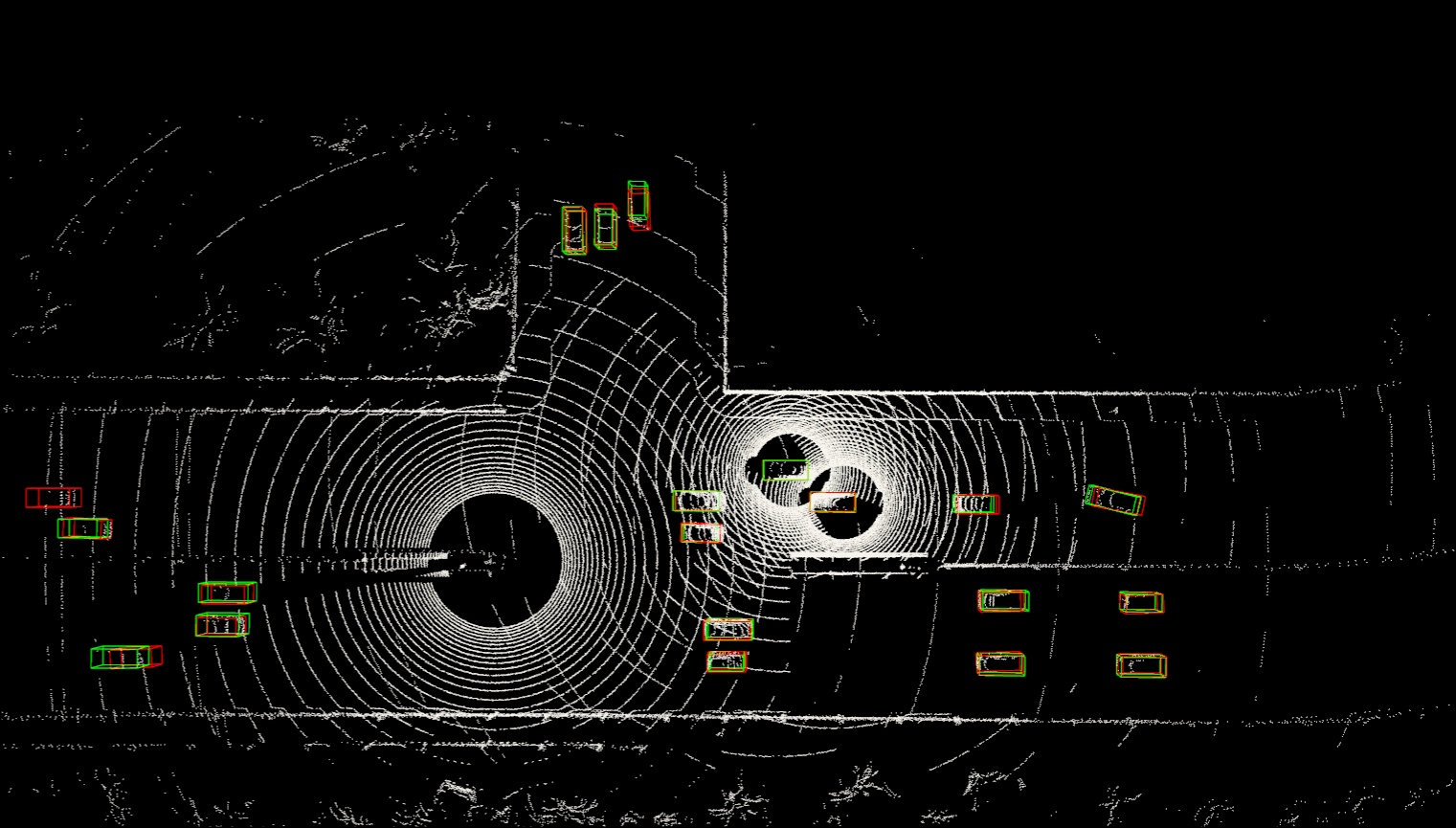}
& \includegraphics[width=\xwidth\linewidth]{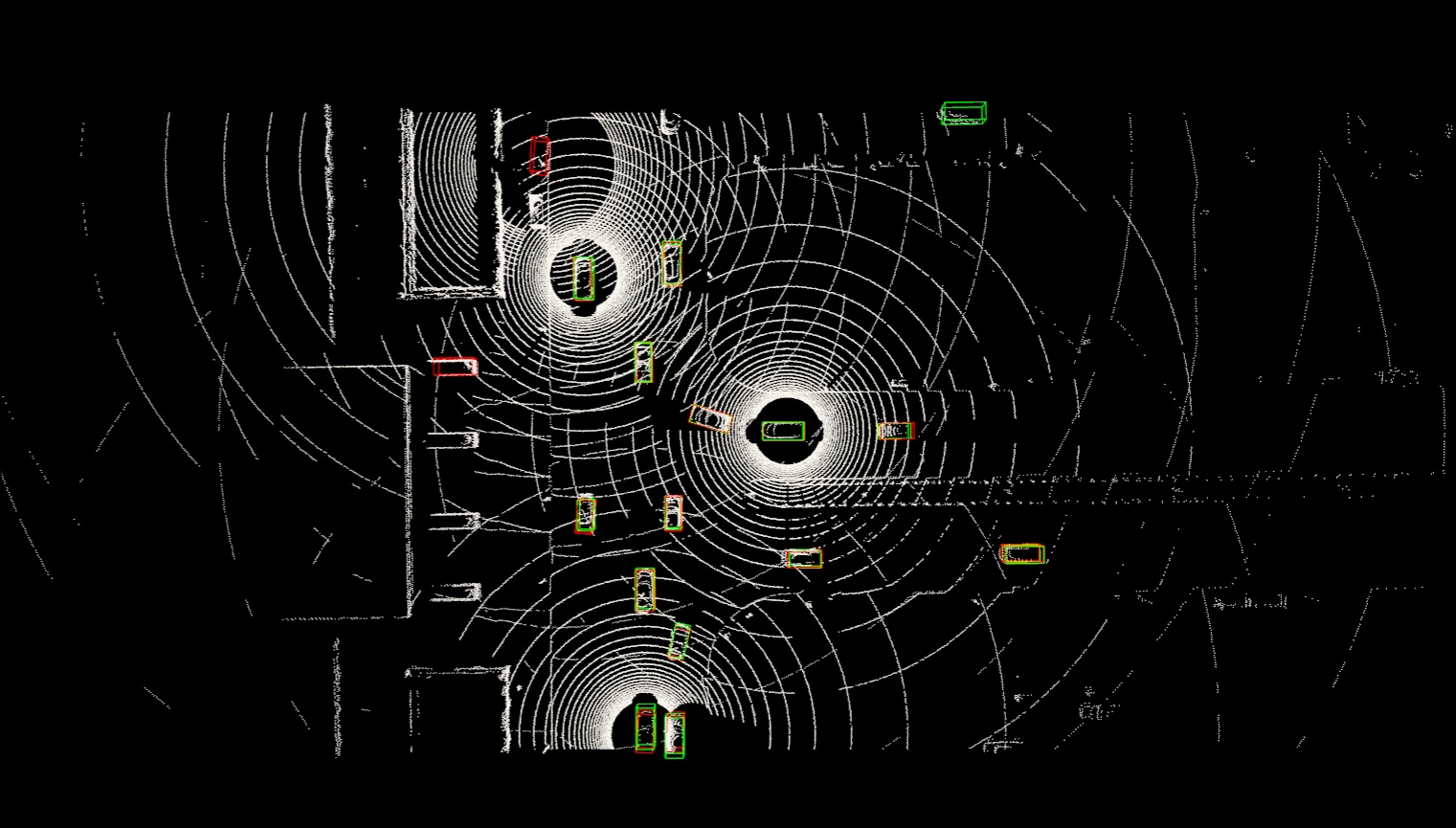} \\
\multirow[t]{1}{*}[\im_shift]{\begin{sideways}  DiscoNet~\cite{li2021learning}  \end{sideways}}  &
 \includegraphics[ width=\xwidth\linewidth]{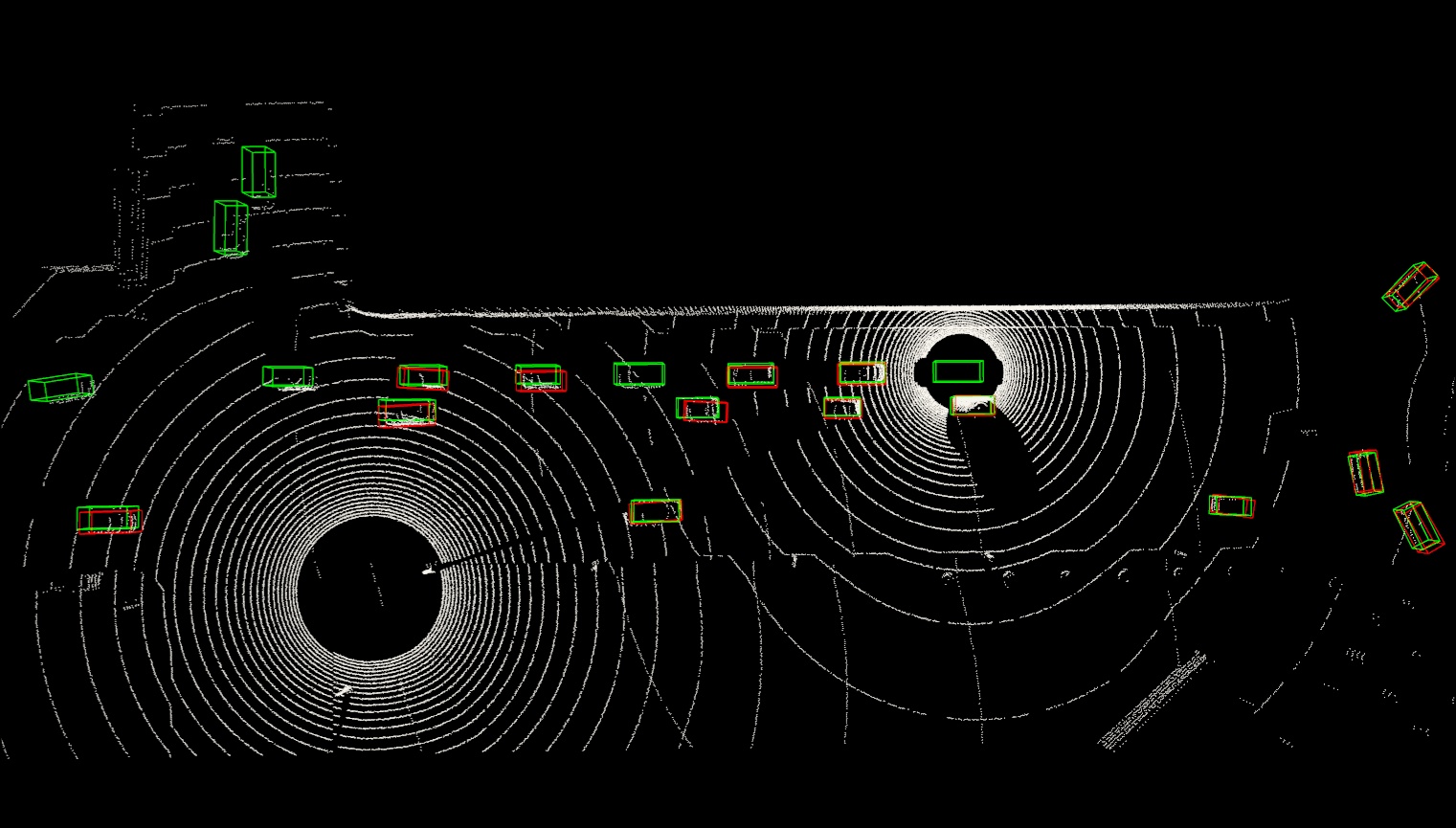}
& \includegraphics[ width=\xwidth\linewidth]{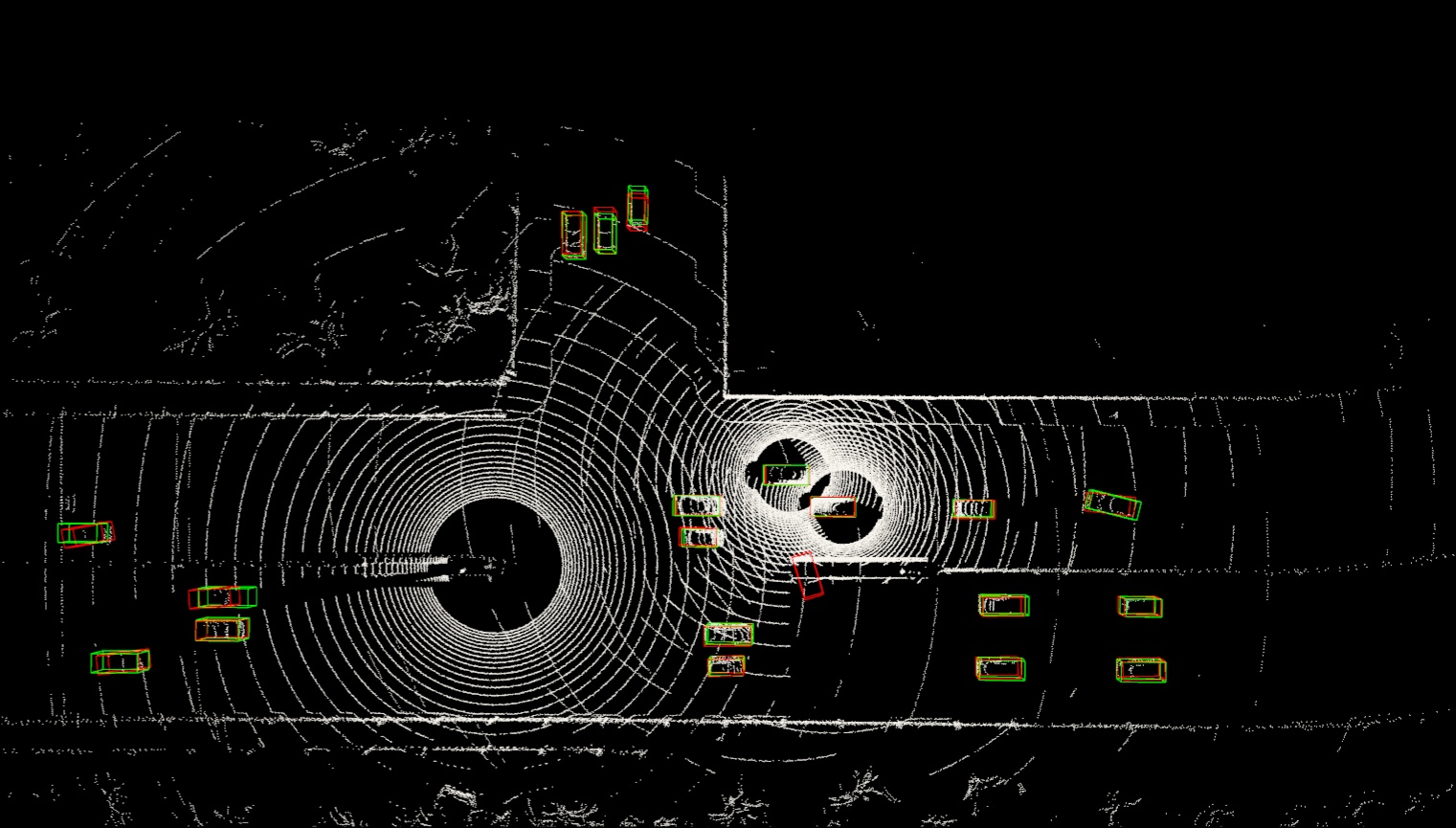}
& \includegraphics[ width=\xwidth\linewidth]{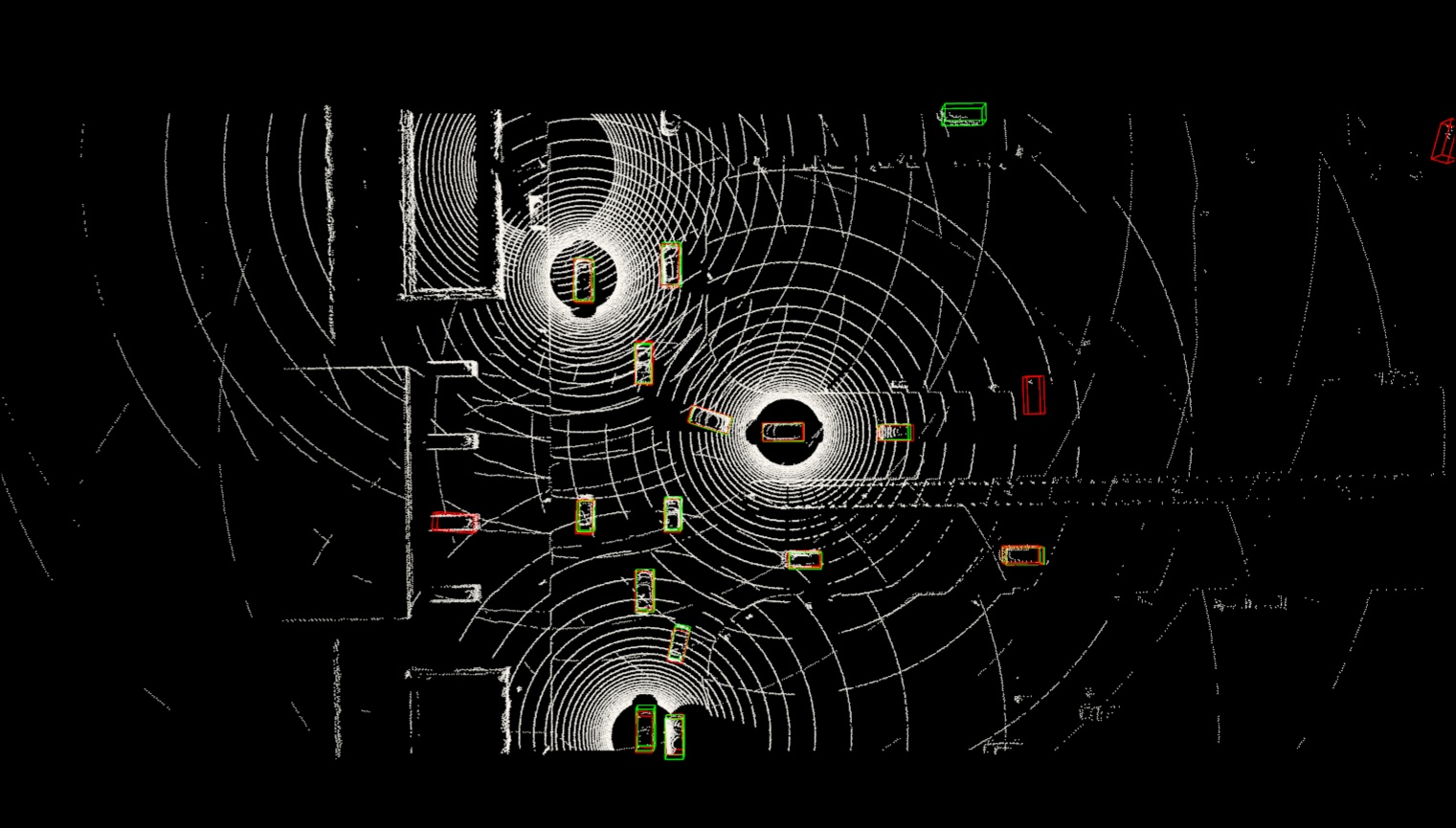}\\
\multirow[t]{1}{*}[0.0\textwidth]{\begin{sideways}  V2X-ViT (ours) \end{sideways}} &
\includegraphics[ width=\xwidth\linewidth]{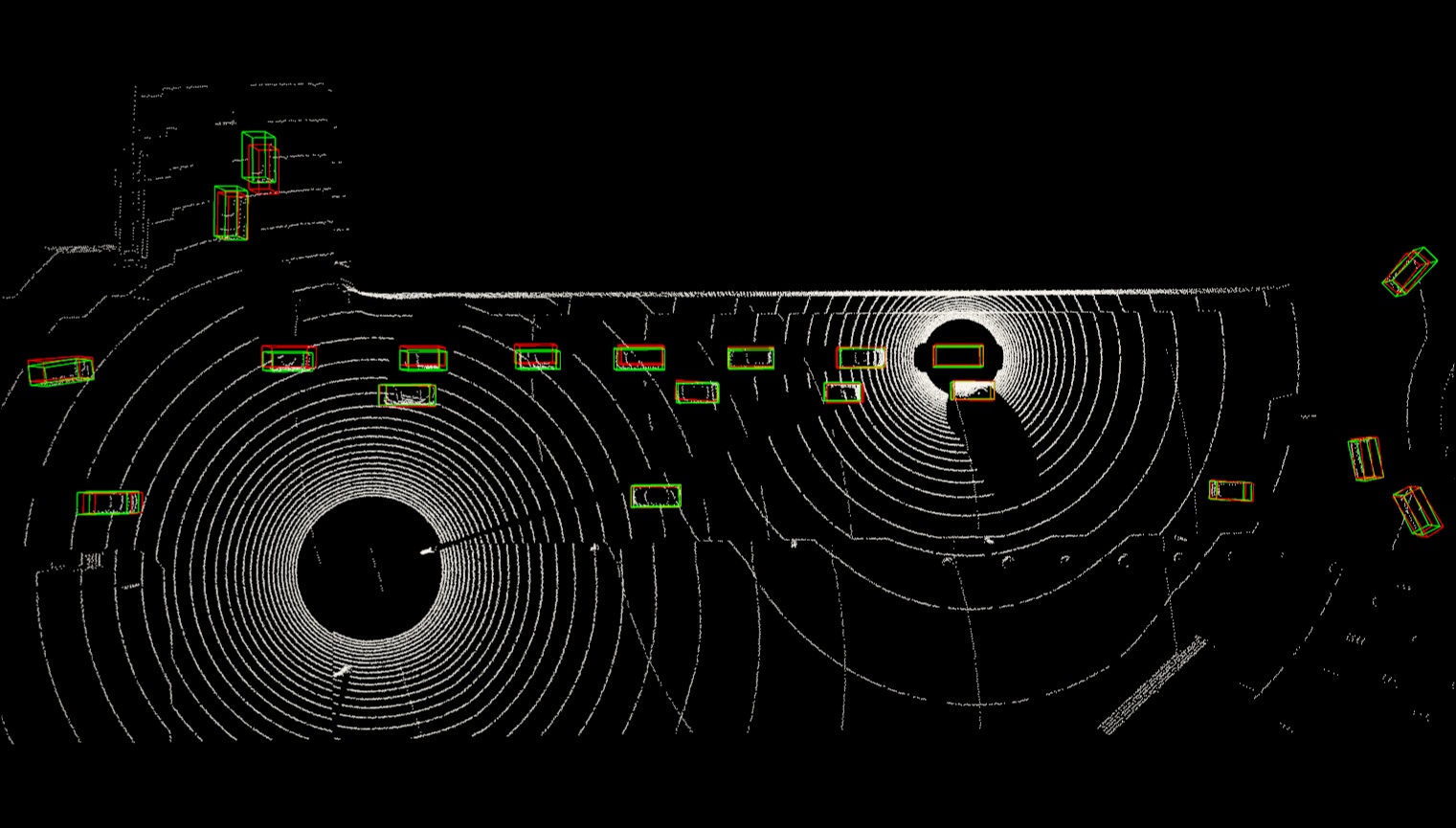}
& \includegraphics[ width=\xwidth\linewidth]{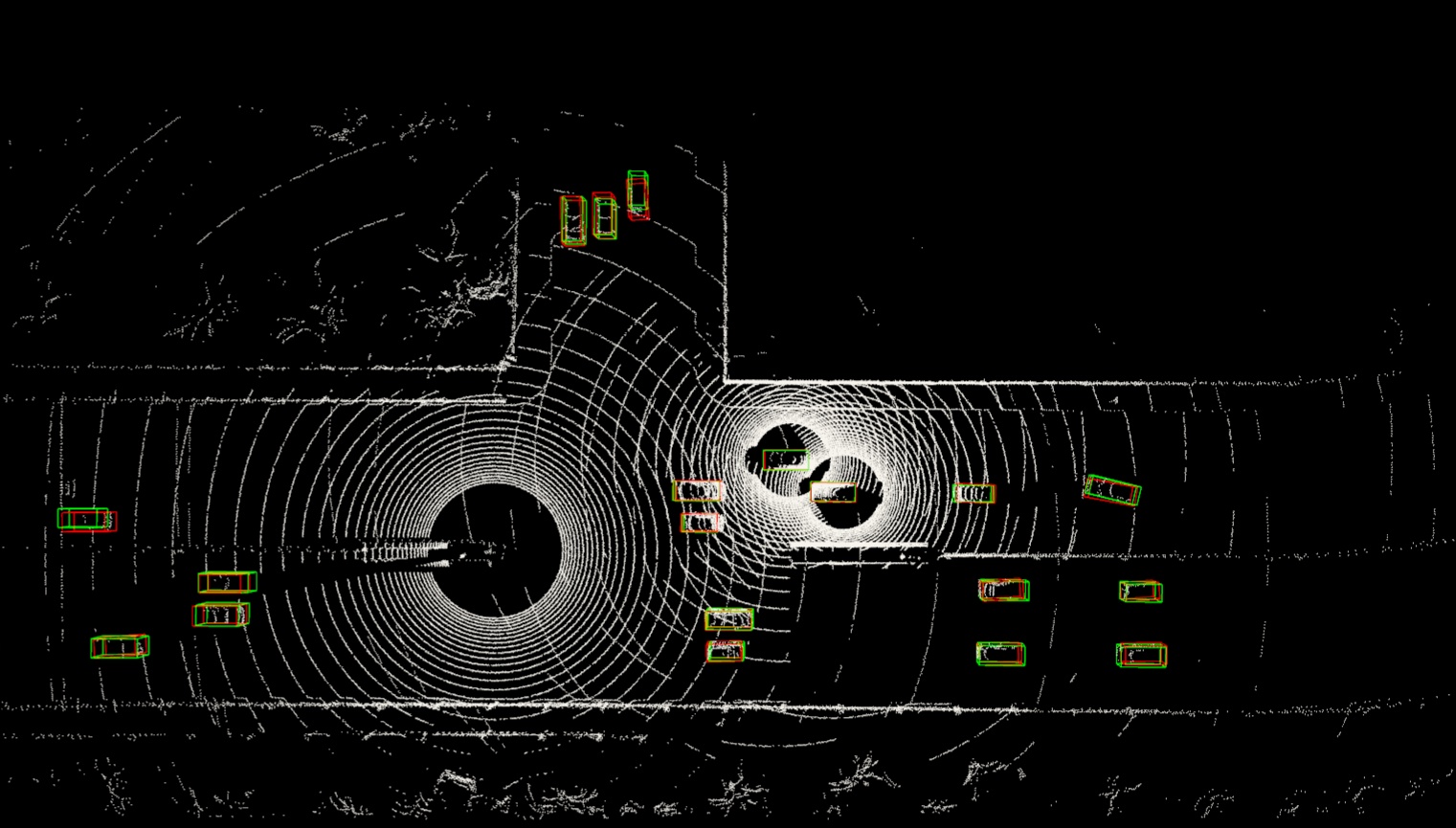}
& \includegraphics[ width=\xwidth\linewidth]{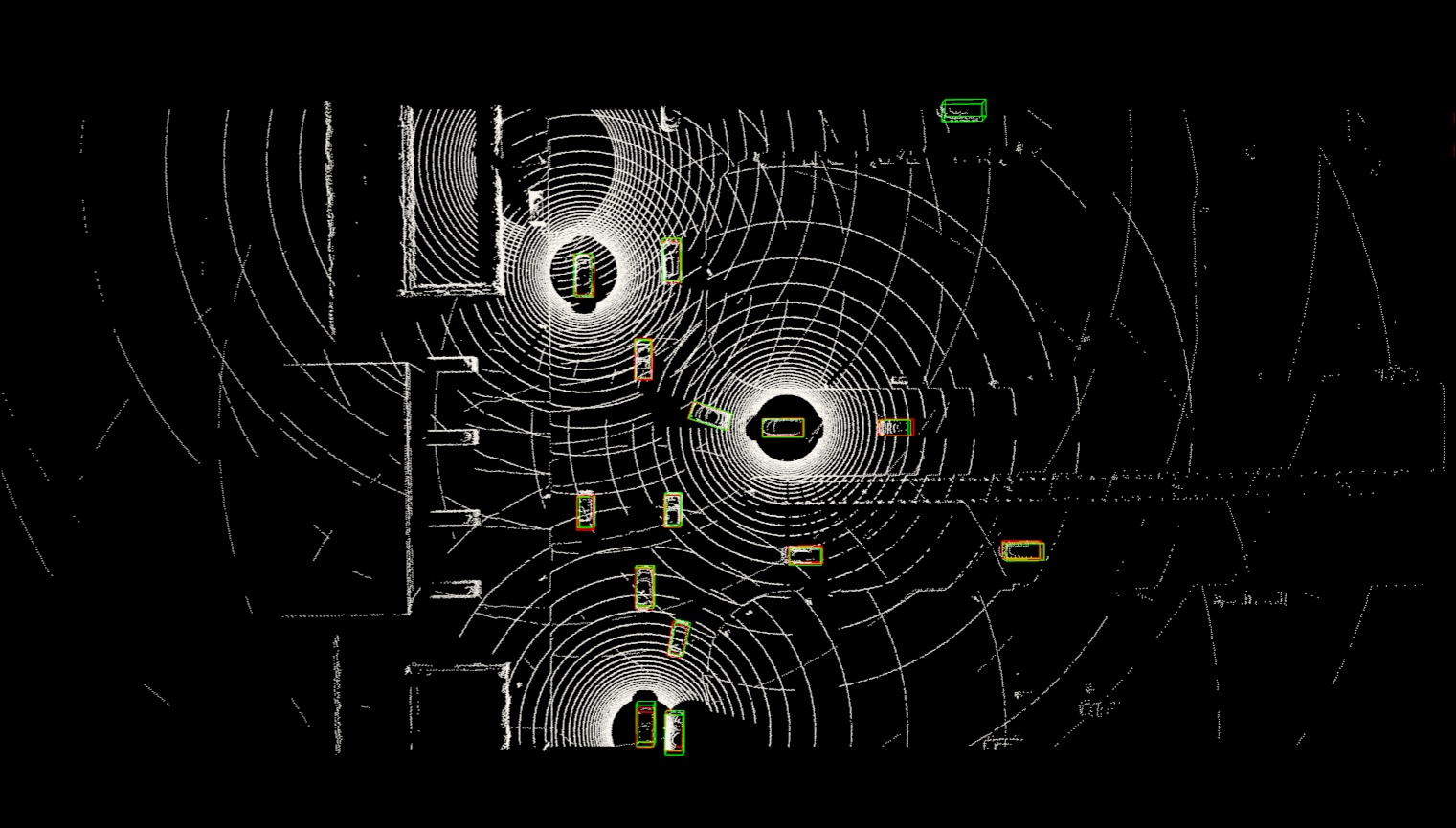}\\
\end{tabular}
\caption{\textbf{Qualitative comparison on scenarios 4-6.} \textcolor{green}{Green} and \textcolor{red}{red} 3D bounding boxes represent the groun truth and prediction respectively.}
\label{fig:qualitive2}
\end{figure*}

\begin{figure*}[!ht]
\centering
\footnotesize
\def\xwidth{0.45}
\def\yheight{0.18}
\def\xem{-2pt}
\def\im_shift{0.05\textwidth}
\setlength{\tabcolsep}{0.5pt}
\begin{tabular}{cccc}
 & Scene 7 & Scene 8\\
 \multirow[t]{1}{*}[\im_shift]{\begin{sideways}  F-Cooper~\cite{chen2019f}  \end{sideways}} &
\includegraphics[ width=\xwidth\linewidth]{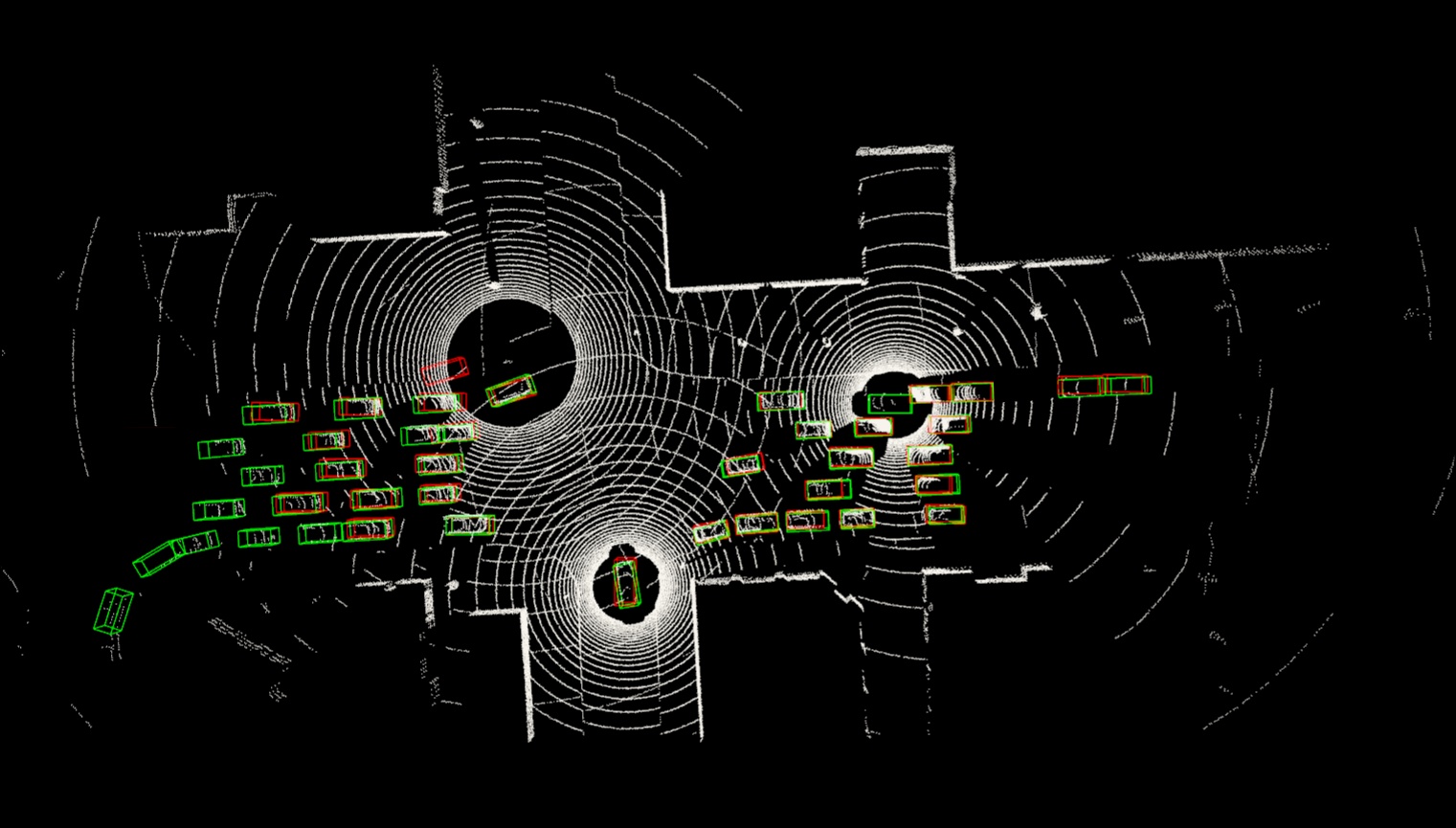}
& \includegraphics[ width=\xwidth\linewidth]{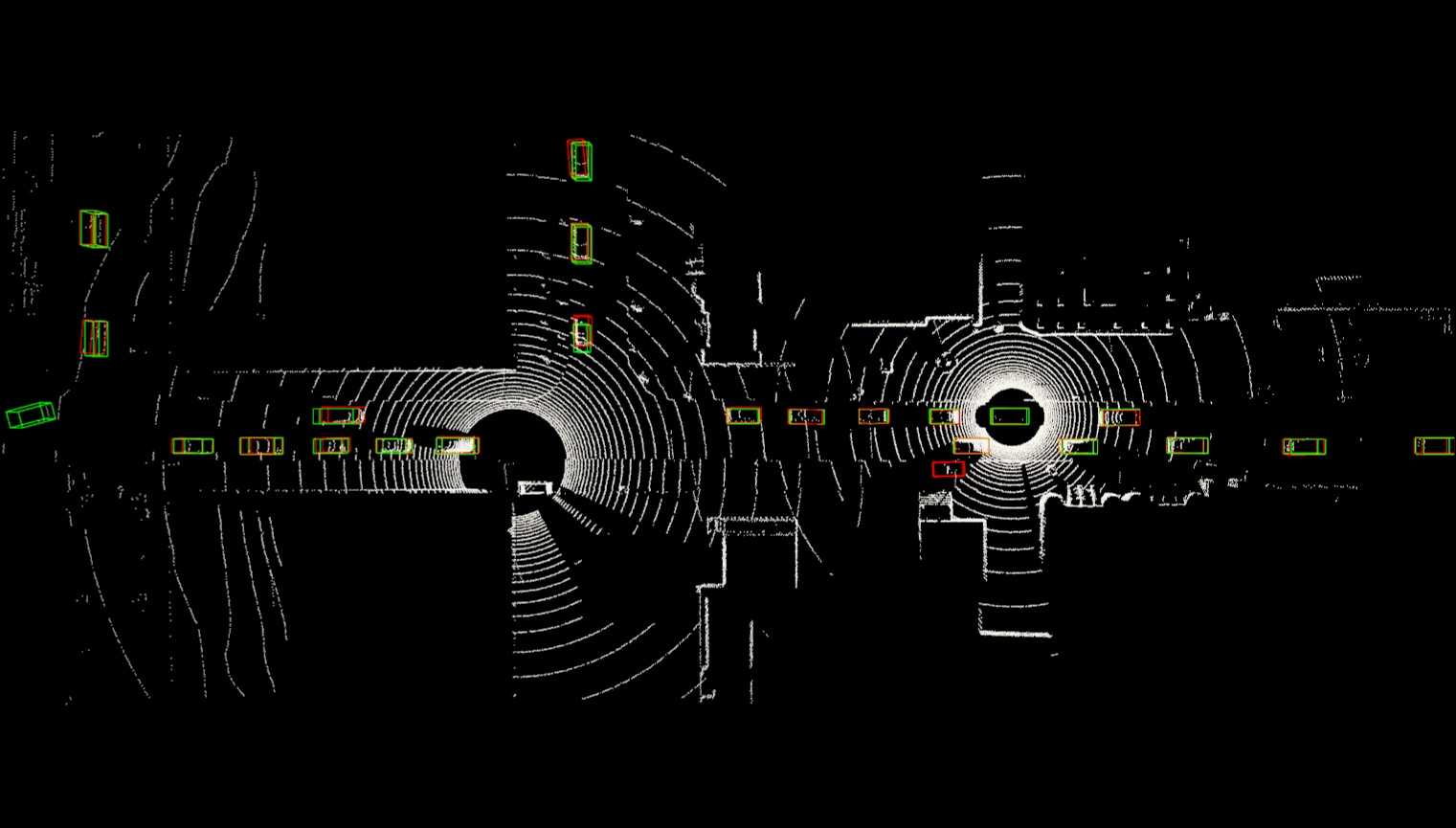}\\
\multirow[t]{1}{*}[\im_shift]{\begin{sideways}  V2VNet~\cite{wang2020v2vnet} \end{sideways}} &
\includegraphics[width=\xwidth\linewidth]{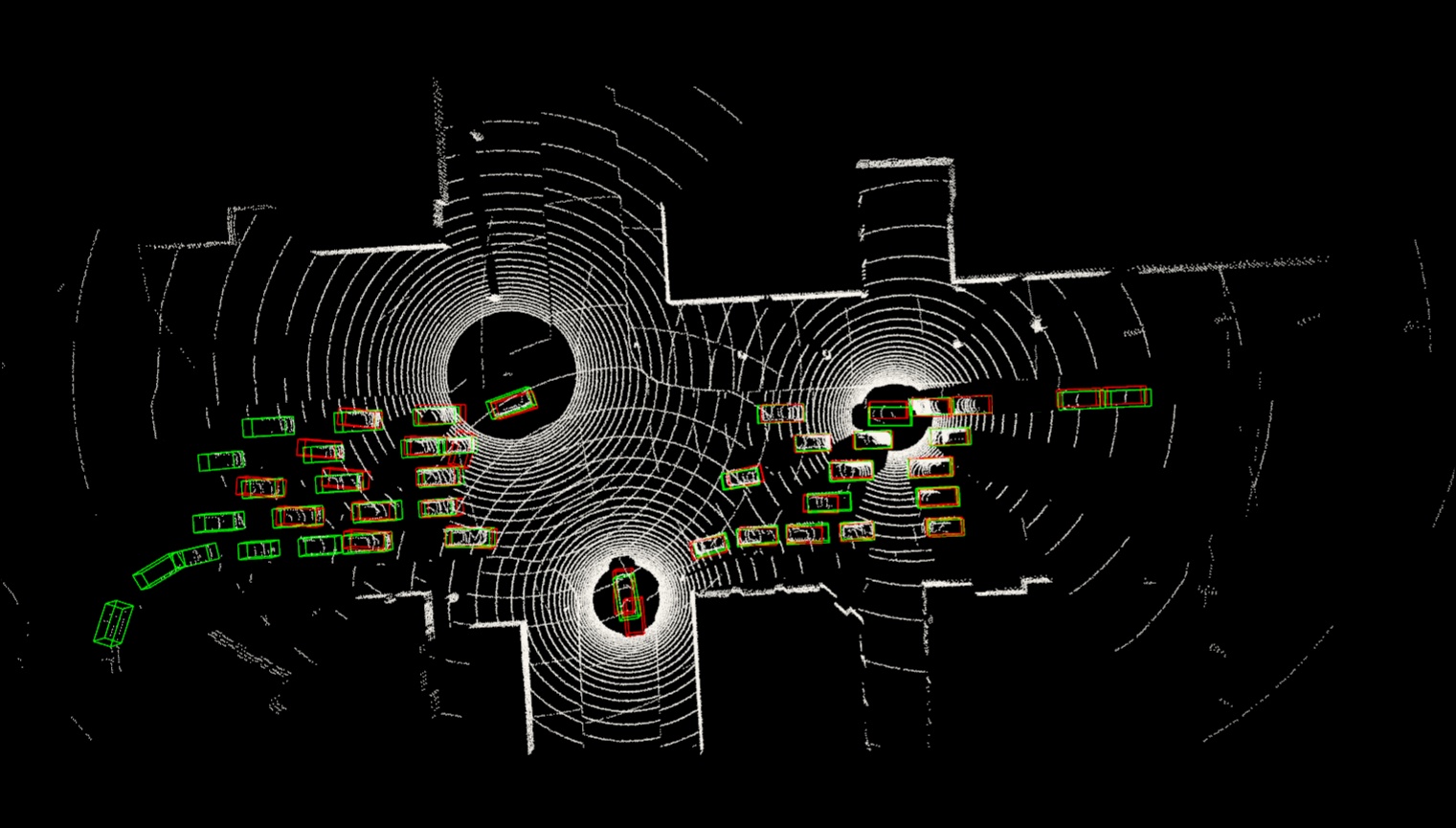}
& \includegraphics[width=\xwidth\linewidth]{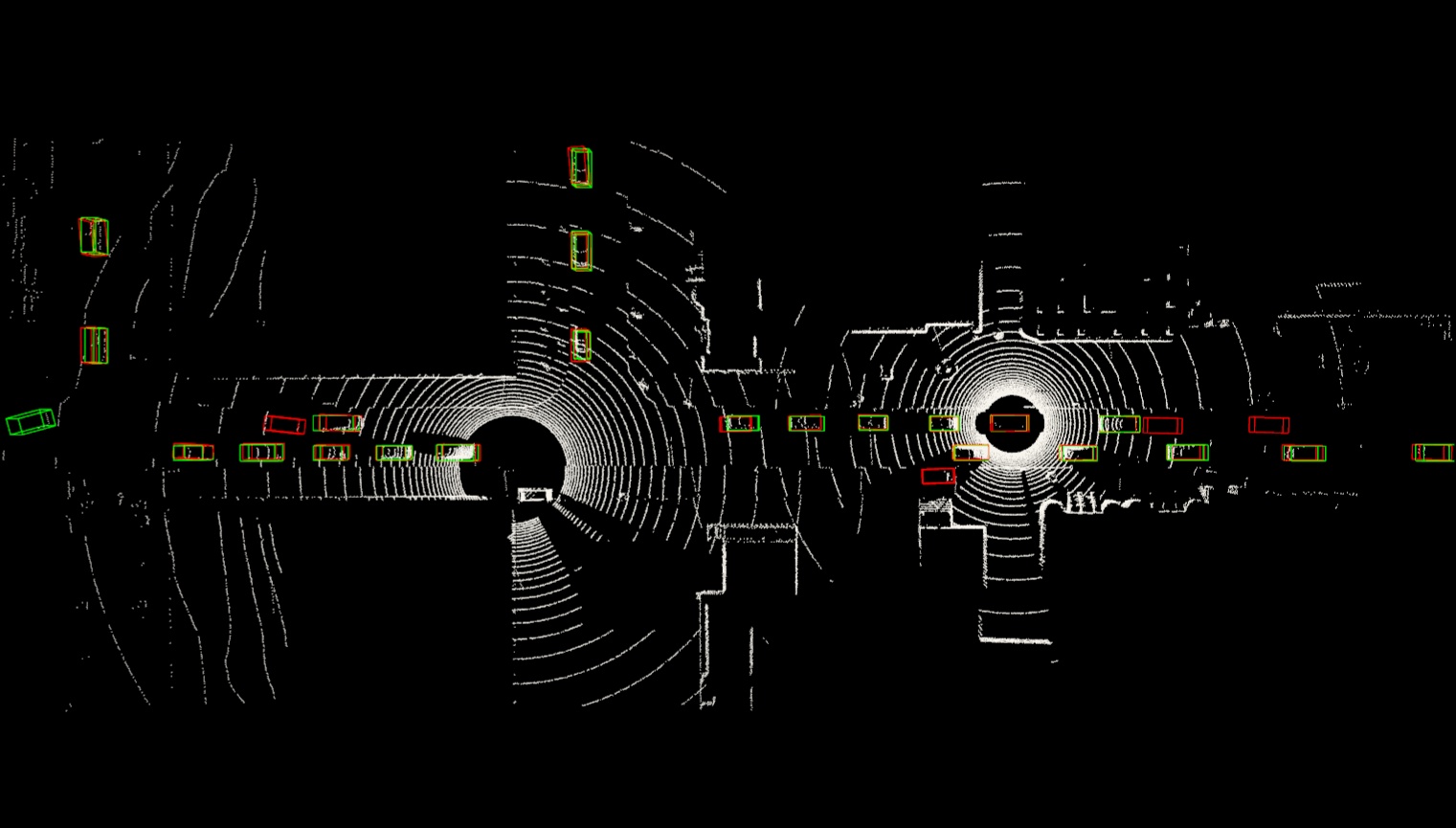}\\
\multirow[t]{1}{*}[\im_shift]{\begin{sideways}  OPV2V~\cite{xu2021opv2v} \end{sideways}} &
 \includegraphics[width=\xwidth\linewidth]{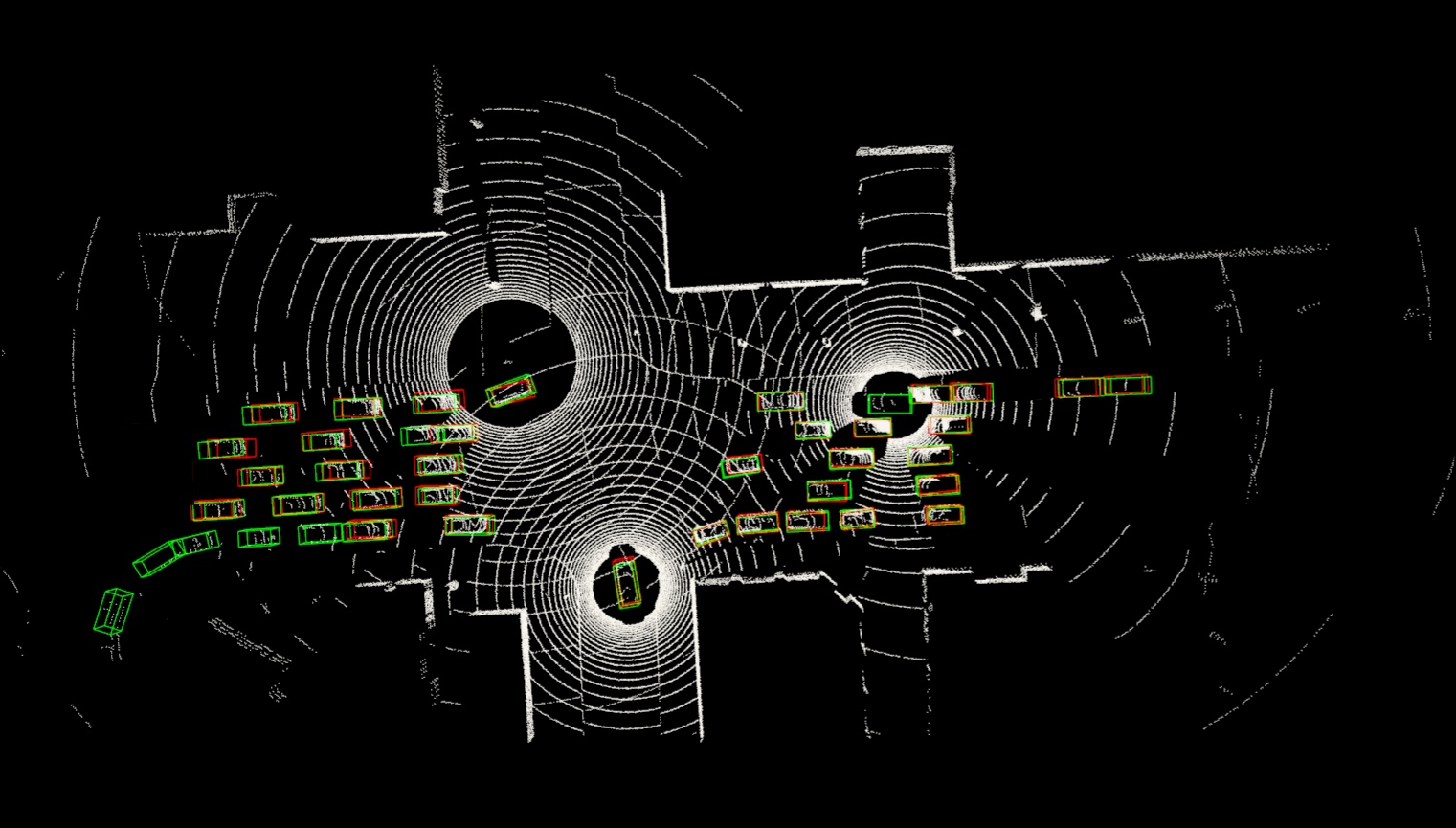}
& \includegraphics[width=\xwidth\linewidth]{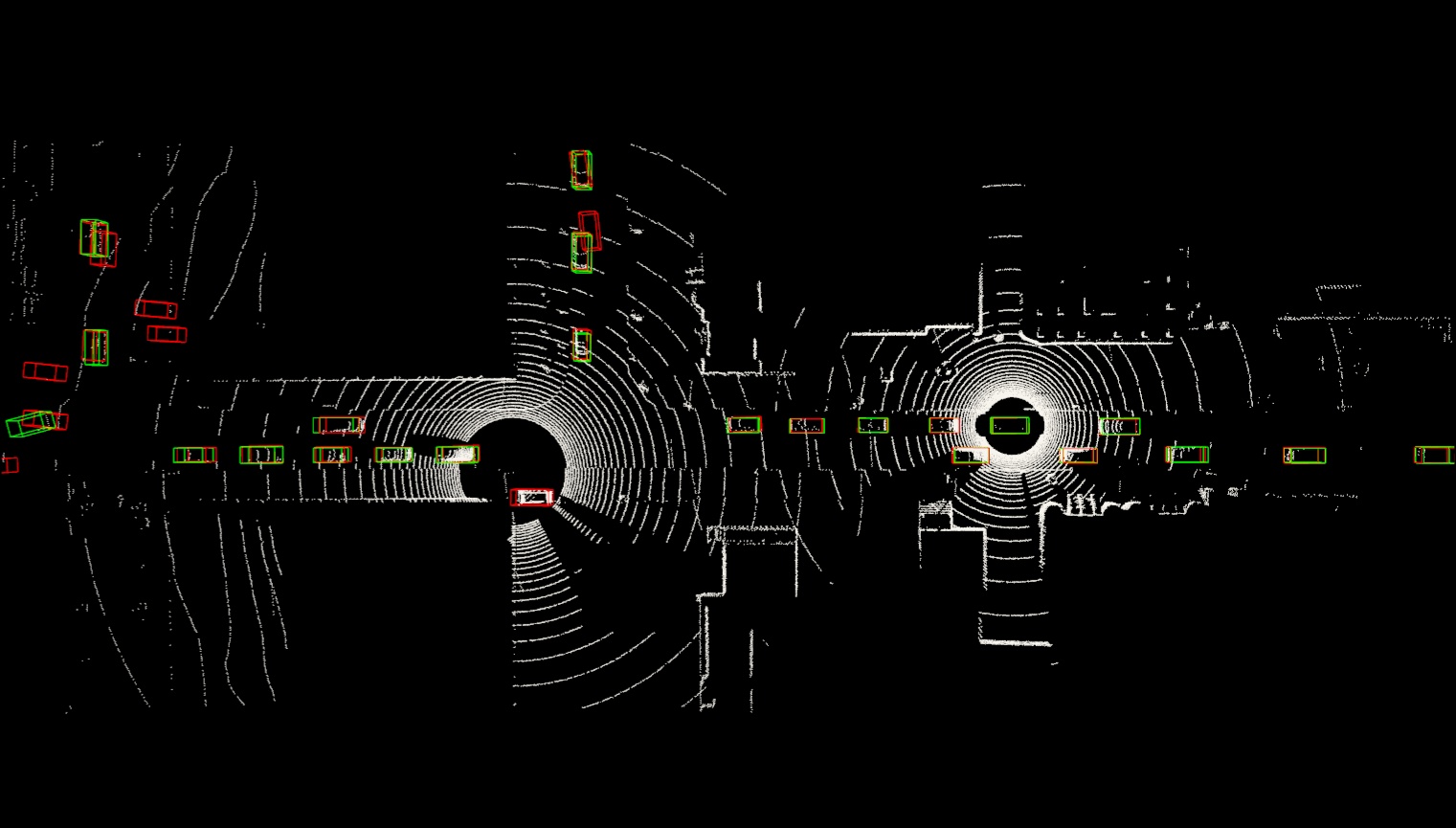}\\
\multirow[t]{1}{*}[\im_shift]{\begin{sideways}  DiscoNet~\cite{li2021learning}  \end{sideways}}  &
 \includegraphics[ width=\xwidth\linewidth]{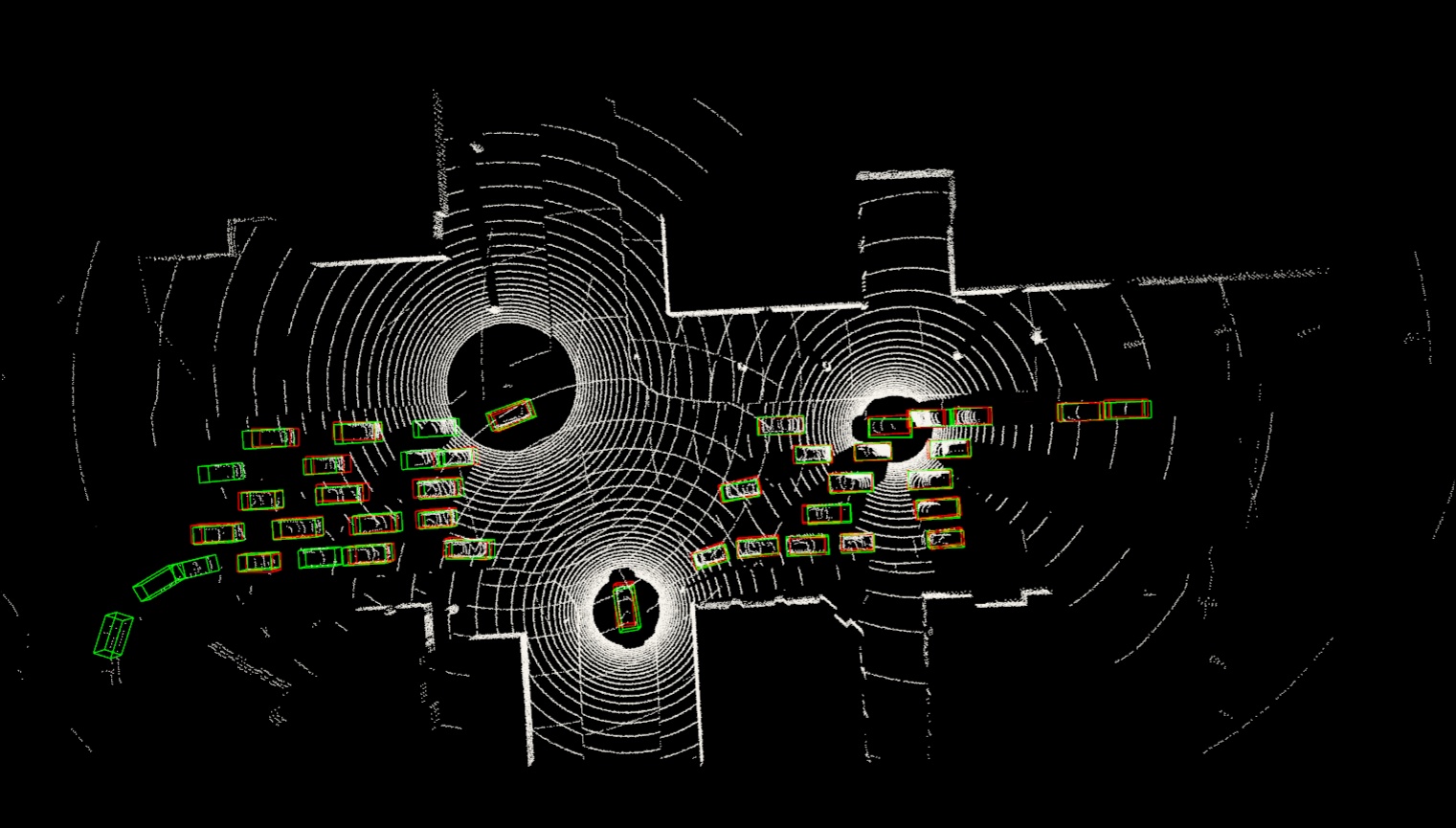}
& \includegraphics[ width=\xwidth\linewidth]{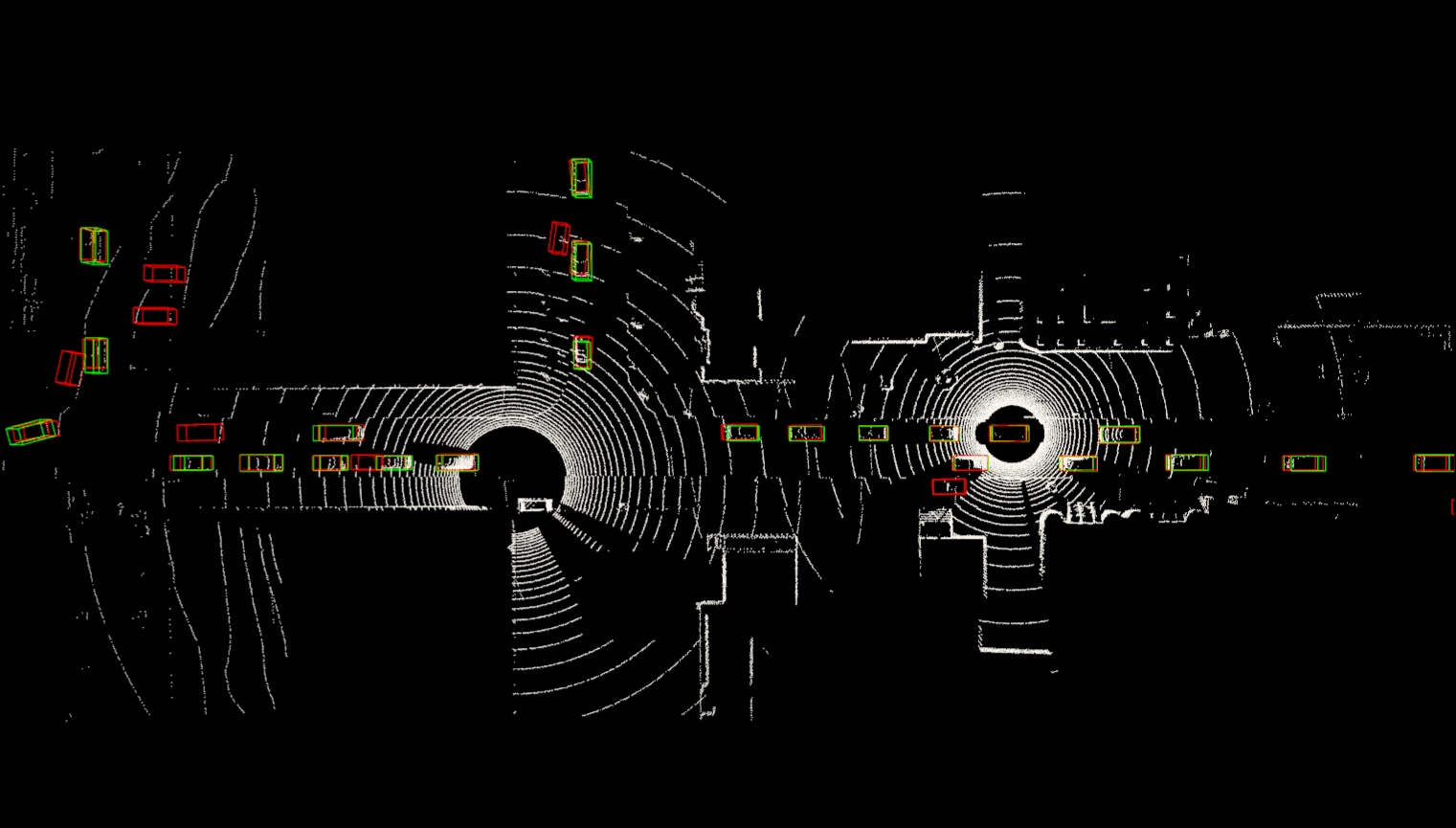}\\
\multirow[t]{1}{*}[0.03\textwidth]{\begin{sideways}  V2X-ViT (ours) \end{sideways}} &
 \includegraphics[ width=\xwidth\linewidth]{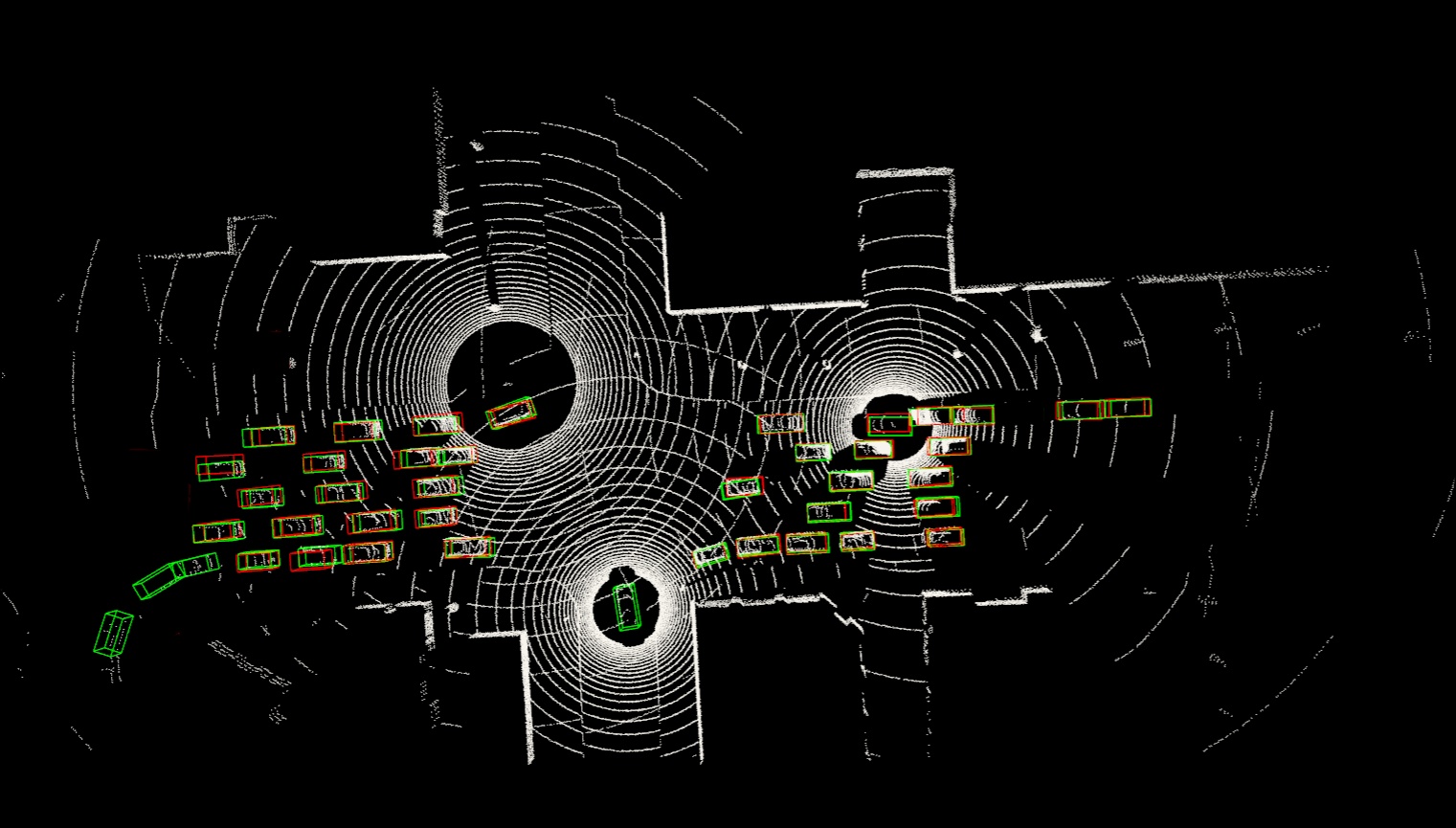}
& \includegraphics[ width=\xwidth\linewidth]{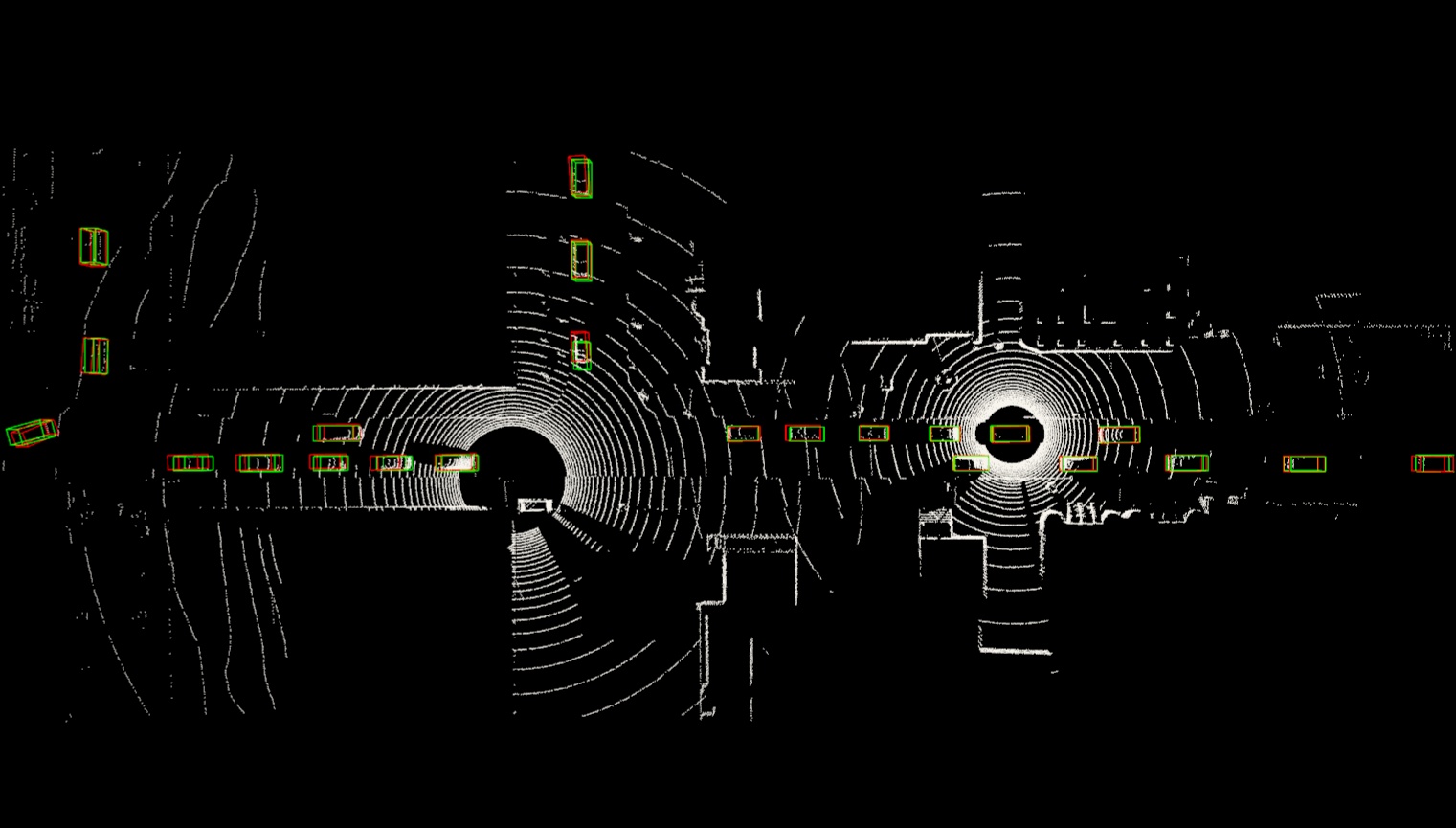}\\
\end{tabular}
\caption{\textbf{Qualitative comparison on scenarios 7-8.} \textcolor{green}{Green} and \textcolor{red}{red} 3D bounding boxes represent the groun truth and prediction respectively.}
\label{fig:qualitive3}
\end{figure*}

\sloppy
\cref{fig:qualitive1,fig:qualitive2,fig:qualitive3} demonstrate more detection visualizations of V2VNet\cite{wang2020v2vnet}, OPV2V~\cite{xu2021opv2v}, F-Cooper~\cite{chen2019f}, DiscoNet~\cite{li2021learning}, and our  V2X-ViT in different scenarios under \textit{Noisy Setting}.  V2X-ViT yields more robust performance in general with fewer regression displacements and fewer undetected objects. When the scenario is challenging with high-density traffic flow and more occlusions~(\eg, Scene 7 in~\cref{fig:qualitive3} ), our model can still identify most of the objects accurately.

\subsection{Attention visualization}
\begin{figure}[!tb]
     \centering
     \def\xwidth{0.32}
     \def\xxwidth{0.4}
     \def\xem{-3pt}
     \begin{subfigure}[b]{\xxwidth\textwidth}
         \centering
         \includegraphics[width=\textwidth]{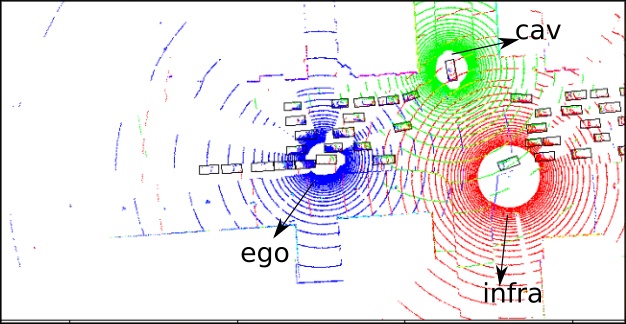}
         \label{fig:att_lidar1}
     \end{subfigure}
     \\ [\xem]
     \begin{subfigure}[b]{\xwidth\textwidth}
         \centering
         \includegraphics[width=\textwidth]{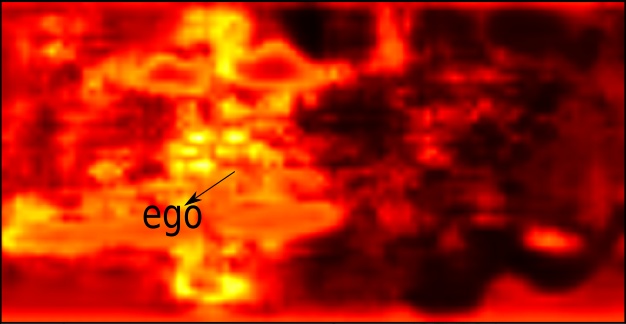}
         \label{fig:att_ego1}
     \end{subfigure}
     \begin{subfigure}[b]{\xwidth\textwidth}
         \centering
         \includegraphics[width=\textwidth]{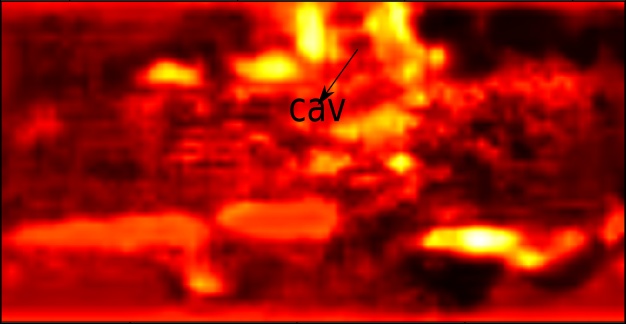}
         \label{fig:att_cav1}
     \end{subfigure}
     \begin{subfigure}[b]{\xwidth\textwidth}
         \centering
         \includegraphics[width=\textwidth]{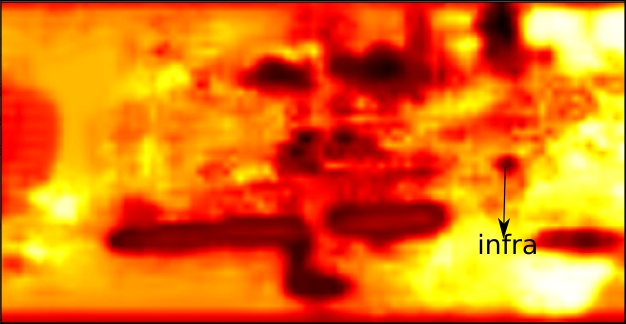}
         \label{fig:att_infra1}
     \end{subfigure}
          \begin{subfigure}[b]{\xxwidth\textwidth}
         \centering
         \includegraphics[width=\textwidth]{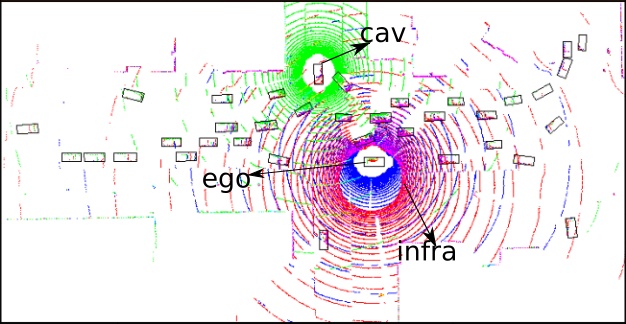}
         \label{fig:att_lidar2}
     \end{subfigure}
     \\[\xem]
     \begin{subfigure}[b]{\xwidth\textwidth}
         \centering
         \includegraphics[width=\textwidth]{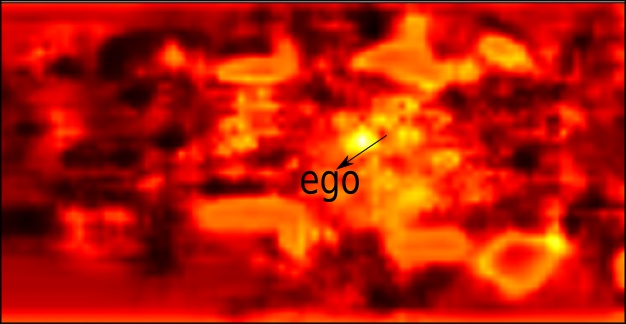}
         \label{fig:att_ego2}
     \end{subfigure}
     \begin{subfigure}[b]{\xwidth\textwidth}
         \centering
         \includegraphics[width=\textwidth]{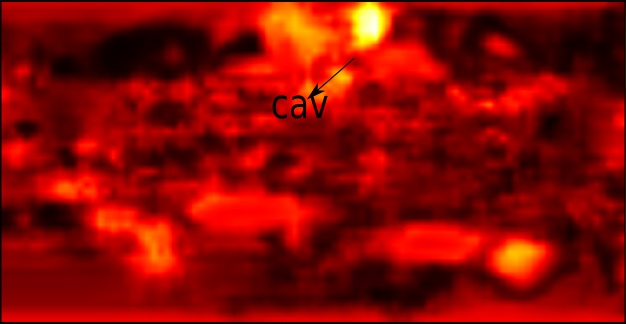}
         \label{fig:att_cav2}
     \end{subfigure}
     \begin{subfigure}[b]{\xwidth\textwidth}
         \centering
         \includegraphics[width=\textwidth]{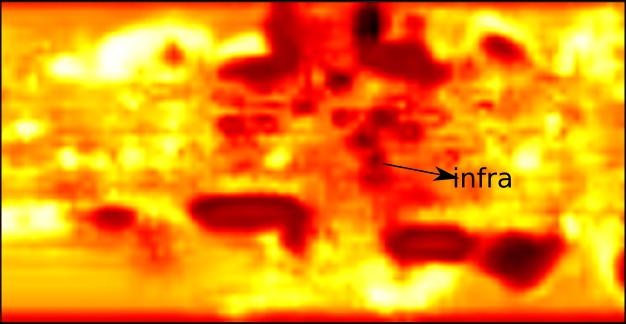}
         \label{fig:att_infra2}
     \end{subfigure}
     \\
          \begin{subfigure}[b]{\xxwidth\textwidth}
         \centering
         \includegraphics[width=\textwidth]{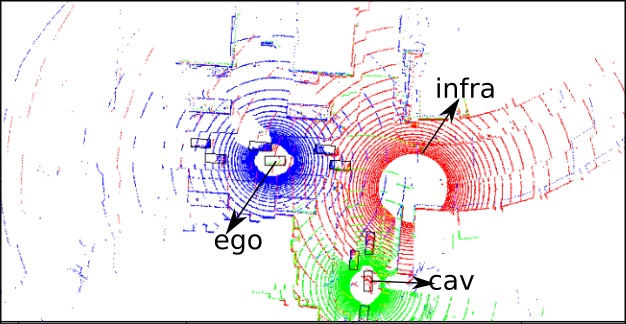}
         \caption{LiDAR point clouds}
         \label{fig:att_lidar3}
     \end{subfigure}
     \\
     \begin{subfigure}[b]{\xwidth\textwidth}
         \centering
         \includegraphics[width=\textwidth]{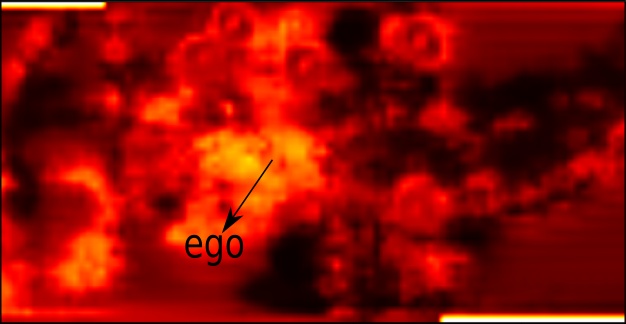}
         \caption{Attention ego$\rightarrow$ego}
         \label{fig:att_ego3}
     \end{subfigure}
     \begin{subfigure}[b]{\xwidth\textwidth}
         \centering
         \includegraphics[width=\textwidth]{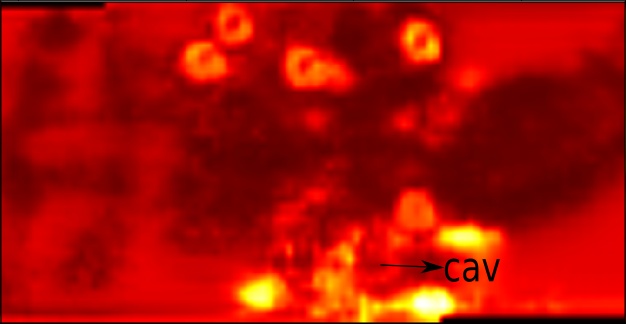}
         \caption{Attention ego$\rightarrow$cav}
         \label{fig:att_cav3}
     \end{subfigure}
     \begin{subfigure}[b]{\xwidth\textwidth}
         \centering
         \includegraphics[width=\textwidth]{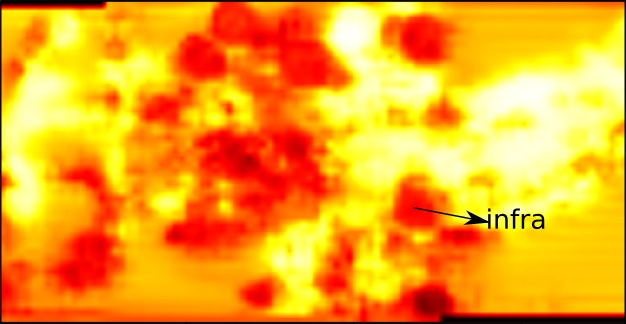}
         \caption{Attention ego$\rightarrow$infra}
         \label{fig:att_infra3}
     \end{subfigure}
     \caption{\textbf{Additional attention map visualizations on 3 different scenes.} V2X-ViT learned to pay more attention to infra features on occluded areas from AV's perspectives, thus yielding more robust detection under occlusions.}
     \label{fig:sup-infra}
\end{figure}

\cref{fig:sup-infra} shows more attention map visualizations of V2X-ViT under \textit{noisy setting}. The LiDAR points of ego vehicle, the other connected autonomous vehicle~(cav), and infrastructure are plotted in \textcolor{blue}{blue}, \textcolor{green}{green}, and \textcolor{red}{red} respectively. The brighter color in the attention map means more attention ego vehicle pays. Generally, the color of infrastructure attention maps is brighter than others, especially for the occluded regions of other agents, indicating the more importance ego vehicle assigns to the infrastructure. This observation agrees with our intuition that the sensor observation of infrastructure has fewer occlusions, which leads to better feature representations.

\subsection{Explanation on effects of transmission size}
 Here we provide more explanations of the data transmission size experiment in our paper. Different fusion strategies usually have distinct bandwidth requirements \eg, early fusion requires large bandwidth to transmit raw data, whereas late fusion only delivers minimal size of data. This communication volume will significantly influence the time delay, thus we need to simulate a more realistic time delay setting to study the effects of transmission size.

Following~\cite{rauch2011analysis}, we decompose the total time delay into two parts: i) the data transmission time $t_c$ during broadcasting, ii) the idle time $t_{i}$ caused by the lack of synchronization between the perception system and communication system. The total time delay is calculated as 
\begin{equation}
    t_{total} = t_c + t_i
\end{equation}
As mentioned in the paper, the data transmission time has
\begin{equation}
    t_c=f_s/v
    \label{eq:t_c}
\end{equation}
where $f_s$ is the data size and $v$ is the transmission rate. Idle time $t_i$ can be further decoupled into the idle time on the sender side and the time on the receiver side \ie, $t_i=t_{i,1}+t_{i,2}$.
 For $t_{i, 1}$,  the worst case in terms of delay happens when the communication system just misses a perception cycle and needs to wait for the next round. Similarly,  for  $t_{i, 2}$,  the worst case occurs when new data is received just after a
new cycle of the perception system has started. Assume both perception system and communication system have the same rate of 10$Hz$, then $0~ms< t_i < 200~ms$. We employ a uniform distribution $\mathcal{U}\left(0,200\right)$ to model this uncertainty. In summary, we use the following equation to mimic the real-world time delay. 
\begin{equation}
    t_c=f_s/v+\mathcal{U}\left(0,200\right)
\end{equation}
which captures the influence of transmission size and asynchrony-caused uncertainty. In practice, we sample the time delay according to Eq.~\ref{eq:t_c} and discretize it to the observed timestamps, which are discrete in a 10Hz update system.

\clearpage
%
%
\bibliographystyle{splncs04}
\bibliography{egbib}
\end{document}